\documentclass[acmsmall]{acmart}

\AtBeginDocument{%
  }

\usepackage{float}
\usepackage{multirow}
\usepackage{mathpartir}
\usepackage{fancyvrb}
\usepackage{xcolor}
\usepackage{subcaption}
\usepackage{amsmath}
\usepackage{caption}
\usepackage{listings}
\usepackage{tikz}
\usepackage{syntax}
\usepackage{matlab-prettifier}
\usepackage{adjustbox}
\usepackage{xspace}
\usepackage{comment}
\usepackage{amsmath}
\usepackage{wrapfig}
\usetikzlibrary{positioning}
\usetikzlibrary{calc}
\usetikzlibrary{fit}
\usetikzlibrary{trees}
\usetikzlibrary{shapes.geometric}
\usetikzlibrary{tikzmark}
\usetikzlibrary{decorations.pathreplacing}
\usepackage{relsize}
\usepackage{xfrac}
\usepackage{mathtools}
\usepackage[inline]{enumitem}
\usepackage{booktabs}
\usepackage{supertabular}
\usepackage{array}
\usepackage[inkscapeformat=png]{svg}
\usepackage{algorithm}
\usepackage[noend]{algpseudocode}
\usepackage{xparse}
\usepackage{etex}
\usepackage[cal=cm,bb=ams]{mathalfa}
\usepackage{pifont}

\usepackage[table]{xcolor}
\microtypesetup{expansion=false}

\algblock{Switch}{EndSwitch}
\algblockdefx{Case}{EndCase}[1]{\textbf{case} #1:}{}

\newcommand{\xmark}{\textcolor{red!80!black}{\ding{55}}}
\newcommand{\cmark}{\textcolor{green!60!black}{\ding{51}}}

\newtheorem{definition}{Definition}[section]  
\newtheorem{lemma}{Lemma}[section]            

\definecolor{WowColor}{rgb}{.75,0,.75}
\definecolor{SubtleColor}{rgb}{0,0,.50}

\definecolor{cAuto}{HTML}{0072B2}
\definecolor{cIvan}{HTML}{D55E00}
\definecolor{cEran}{HTML}{009E73}
\newcommand{\mkAuto}{\tikz[baseline=-0.55ex]{\fill[cAuto] (0,0) circle [radius=0.6ex];}}
\newcommand{\mkIvan}{\tikz[baseline=-0.55ex]{\fill[cIvan] (-0.6ex,-0.35ex) -- (0.6ex,-0.35ex) -- (0,0.65ex) -- cycle;}}
\newcommand{\mkEran}{\tikz[baseline=-0.55ex]{\fill[cEran] (-0.5ex,-0.4ex) rectangle (0.5ex,0.6ex);}}

\definecolor{cNS}{HTML}{7D4BB3}
\definecolor{cNR}{HTML}{118D8D}
\definecolor{cNRS}{HTML}{E0A100}
\newcommand{\mkNS}{\tikz[baseline=-0.55ex]{\fill[cNS] (0,0) circle [radius=0.6ex];}}
\newcommand{\mkNR}{\tikz[baseline=-0.55ex]{\fill[cNR] (-0.6ex,-0.35ex) -- (0.6ex,-0.35ex) -- (0,0.65ex) -- cycle;}}
\newcommand{\mkNRS}{\tikz[baseline=-0.55ex]{\fill[cNRS] (-0.5ex,-0.4ex) rectangle (0.5ex,0.6ex);}}
\newcounter{margincounter}

\newcommand{\cf}{\textsc{ConstraintFlow}\xspace}
\newcommand{\compiler}{\textsc{Neuton}\xspace}

\newcommand{\typepoly}{\cftypewords{\texttt{PolyExp}}\xspace}
\newcommand{\typezono}{\cftypewords{\texttt{SymExp}}\xspace}

\newcommand{\typereal}{\cftypewords{\texttt{Real}}\xspace}

\newcommand{\sparse}{g-BCSR\xspace}

\newcommand{\constant}{c}
\newcommand{\expr}{e}

\newcommand{\curr}{\cfkeywords{\texttt{curr}}\xspace}
\newcommand{\prev}{\cfkeywords{\texttt{prev}}\xspace}

\newcommand{\map}{\cfkeywords{\texttt{map}}\xspace}

\newcommand{\dott}{\cfkeywords{\texttt{dot}}\xspace}

\newcommand{\traverse}{\cfkeywords{\texttt{traverse}}\xspace}

\newcommand{\types}{t}

\newcommand{\fstore}{F}

\newcommand{\sstore}{\sigma}

\newcommand{\cfkeywords}[1]{\texttt{\footnotesize\bfseries\textcolor{keywords}{#1}}}
\newcommand{\cftypewords}[1]{\texttt{\footnotesize\bfseries\textcolor{typewords}{#1}}}

\newcommand{\func}{\cfkeywords{\texttt{Func}}\xspace}

\newcommand{\polyexp}{\cftypewords{\texttt{PolyExp}}\xspace}

\newcommand{\symexp}{\cftypewords{\texttt{SymExp}}\xspace}

\newcommand{\float}{\cftypewords{\texttt{Real}}\xspace}

\newcommand{\bool}{\cftypewords{\texttt{Bool}}\xspace}
\newcommand{\typeneuron}{\cftypewords{\texttt{Neuron}}\xspace}
\newcommand{\typesym}{\cftypewords{\texttt{Sym}}\xspace}

\newcommand{\true}{\mathsf{true}}
\newcommand{\false}{\mathsf{false}}

\usepackage{amsmath}
\usepackage{mathpartir}  

\usepackage{graphicx}
\usepackage{caption}
\usepackage[skip=0.5ex]{subcaption}

\makeatletter
\newsavebox{\@brx}
\newcommand{\llangle}[1][]{\savebox{\@brx}{\(\m@th{#1\langle}\)}%
  \mathopen{\copy\@brx\kern-0.5\wd\@brx\usebox{\@brx}}}
\newcommand{\rrangle}[1][]{\savebox{\@brx}{\(\m@th{#1\rangle}\)}%
  \mathclose{\copy\@brx\kern-0.5\wd\@brx\usebox{\@brx}}}
\makeatother

\newcounter{number}

\definecolor{diagramcolor}{rgb}{0.70,0.0,0.56}
\definecolor{keywords}{rgb}{0.05,0.05,0.9}
\definecolor{typewords}{rgb}{0,0.5,0}
\definecolor{greencomments}{rgb}{0,0.5,0}
\definecolor{turqusnumbers}{rgb}{0.17,0.57,0.69}
\definecolor{redstrings}{rgb}{0.5,0,0}

\definecolor{codegreen}{rgb}{0,0.6,0}
\definecolor{codegray}{rgb}{0.5,0.5,0.5}
\definecolor{codepurple}{rgb}{0.58,0,0.82}
\definecolor{backcolour}{RGB}{250, 250, 250}

\lstdefinelanguage{ConstraintFlow}
    {morekeywords={def, shape, as, curr, prev, prev0, prev1, mapList, transformer, ReLU, Affine, HardSwish, Maxpool, DotProduct, rev_ReLU, rev_Affine, rev_Maxpool, rev_Max, rev_Min, rev_Add, rev_Mult, func, map, true, false, traverse, dot, flow, forward, backward, sum, layer, sym, compare, avg, len, max, min, and, in, solver, currList, equations, minimize, maximize, mult, add, sigmoid, tanh},
    morekeywords = [4]{Bool, Int, Real, PolyExp, SymExp, Neuron, Noise, Ct},
    keywordstyle = \bfseries\color{keywords},
    keywordstyle = [4]{\bfseries\color{typewords}},
    sensitive=false, 
    morecomment=[l][\color{greencomments}]{///},
    morecomment=[l][\color{greencomments}]{//},
    morecomment=[s][\color{greencomments}]{{(*}{*)}},
    morestring=[b]",
    stringstyle=\color{redstrings}
    }

\definecolor{gbcsrcall}{HTML}{CC5500}
\definecolor{libcall}{HTML}{7A1FA2}
\lstdefinelanguage{NeutonPy}[]{Python}{
    morekeywords=[2]{repeat, binary, sum, inner_prod, matmul, clamp,
                     convert_to_float, unsqueeze},
    morekeywords=[3]{dot, get_metadata, get_poly_size, get_mat, get_const,
                     get_elem, PolyExpSparse, SparseTensor},
    keywordstyle=\bfseries\color{keywords},
    keywordstyle=[2]\bfseries\color{gbcsrcall},
    keywordstyle=[3]\bfseries\color{libcall},
    commentstyle=\rmfamily\itshape\color{greencomments},
    stringstyle=\color{redstrings},
    sensitive=true
}
\lstnewenvironment{pylisting}
    {\lstset{language=NeutonPy,
                basicstyle=\ttfamily\footnotesize,
                columns=fullflexible,
                keepspaces=true,
                showstringspaces=false,
                numbers=left,
                numbersep=5pt,
                numberstyle=\tiny\color{black},
                backgroundcolor=\color{backcolour},
                xleftmargin=13pt,
                breaklines=true,
                escapeinside={|}{|}
    }}
    {}

\lstnewenvironment{cflisting}
    {\lstset{language=ConstraintFlow,
                basicstyle=\ttfamily,
                breaklines=true,
                columns=fullflexible
    }}
    {}

\makeatletter
\lst@AddToHook{OnEmptyLine}{\addtocounter{lstnumber}{-1}}

\makeatother

\newcommand{\kblock}{\textsf{StencilBlock}\xspace}
\newcommand{\pblock}{\textsf{WindowBlock}\xspace}
\newcommand{\rblock}{\textsf{BroadcastBlock}\xspace}
\newcommand{\cblock}{\textsf{UniformBlock}\xspace}
\newcommand{\dblock}{\textsf{DenseBlock}\xspace}
\newcommand{\diagblock}{\textsf{DiagonalBlock}\xspace}

\newcommand{\isconst}{\kappa}

\newcommand{\irshape}{s}
\newcommand{\irbroadcast}{\beta}
\newcommand{\currs}{\textsf{currSize}\xspace}
\newcommand{\prevs}{\textsf{prevSize}\xspace}
\newcommand{\polys}{\textsf{polySize}\xspace}
\newcommand{\syms}{\textsf{symSize}\xspace}
\newcommand{\batchs}{\textsf{batchSize}\xspace}

\newcommand{\irme}[1][]{\ifthenelse{\equal{#1}{}}{m_e}{m_{e_{#1}}}}
\newcommand{\irm}{m}
\newcommand{\height}{\textsf{Ht}\xspace}

\newcommand{\isexpanded}{\textsf{IsExpanded}\xspace}
\newcommand{\lcm}{\textsf{LCM}\xspace}
\newcommand{\matchdims}{\textsf{alignShapes}\xspace}

\newcommand{\irexpression}[1][]{\ifthenelse{\equal{#1}{}}{i_e}{i_{e_{#1}}}}
\newcommand{\irconst}{\textsf{IrConst}\xspace}
\newcommand{\irvar}{\textsf{IrVar}\xspace}
\newcommand{\irnoise}{\textsf{IrSym}\xspace}

\newcommand{\irrepeat}{\textsf{IrRepeat}\xspace}
\newcommand{\irremovedim}{\textsf{IrRemoveDimension}\xspace}
\newcommand{\iradddim}{\textsf{IrAddDimension}\xspace}
\newcommand{\iradddimconst}{\textsf{IrAddDimensionConst}\xspace}
\newcommand{\irternary}{\textsf{IrTernary}\xspace}
\newcommand{\irbinary}{\textsf{IrBinary}\xspace}
\newcommand{\irclamp}{\textsf{IrClamp}\xspace}
\newcommand{\irunary}{\textsf{IrUnary}\xspace}
\newcommand{\irmult}{\textsf{IrMult}\xspace}
\newcommand{\irinner}{\textsf{IrInnerProduct}\xspace}
\newcommand{\irdot}{\textsf{IrDot}\xspace}
\newcommand{\ircombinepoly}{\textsf{IrCombineToPoly}\xspace}
\newcommand{\ircombinesym}{\textsf{IrCombineToSym}\xspace}
\newcommand{\irextractpolycoeff}{\textsf{IrExtractPolyCoeff}\xspace}
\newcommand{\irextractpolyconst}{\textsf{IrExtractPolyConst}\xspace}
\newcommand{\irextractsymcoeff}{\textsf{IrExtractSymCoeff}\xspace}
\newcommand{\irextractsymconst}{\textsf{IrExtractSymConst}\xspace}
\newcommand{\irneuronpoly}{\textsf{IrNeuronToPoly}\xspace}
\newcommand{\irconstpoly}{\textsf{IrConstToPoly}\xspace}
\newcommand{\irnoisesym}{\textsf{IrNoiseToSym}\xspace}
\newcommand{\irconstsym}{\textsf{IrConstToSym}\xspace}
\newcommand{\irreduce}{\textsf{IrReduce}\xspace}
\newcommand{\iraccess}{\textsf{IrAccess}\xspace}
\newcommand{\irmapcoeff}{\textsf{IrMapCoeff}\xspace}
\newcommand{\irmapneuron}{\textsf{IrMapNeuron}\xspace}
\newcommand{\irmapnoise}{\textsf{IrMapNoise}\xspace}

\newcommand{\irstatement}[1][]{\ifthenelse{\equal{#1}{}}{i_s}{i_{s_{#1}}}}
\newcommand{\irskip}{\textsf{IrSkip}\xspace}
\newcommand{\irassignment}{\textsf{IrAssignment}\xspace}
\newcommand{\irtransretbasic}{\textsf{IrReturn}\xspace}
\newcommand{\irwhile}{\textsf{IrWhile}\xspace}
\newcommand{\irite}{\textsf{IrITE}\xspace}

\newcommand{\irseq}{\textsf{IrSeq}\xspace}

\newcommand{\neuron}{\eta}

\setcopyright{none}   
\begin{document}

\title{Automatic Tensor-level Execution of Abstract Interpreters for Neural Networks}

\author{Avaljot Singh}
\email{avaljot2@illinois.edu}
\orcid{0009-0006-4167-8709}
\affiliation{%
  \institution{University of Illinois Urbana-Champaign}
  \country{USA}
}

\author{Yasmin Chandini Sarita}
\affiliation{%
  \institution{University of Illinois Urbana-Champaign}
  \country{USA}
}
\email{ysarita2@illinois.edu}

\author{Aditya Mishra}
\affiliation{%
  \institution{University of Illinois Urbana-Champaign}
  \country{USA}
}
\email{mishra27@illinois.edu}


\author{Gagandeep Singh}
\affiliation{%
  \institution{University of Illinois Urbana-Champaign}
  \country{USA}
}
\email{ggnds@illinois.edu}

\author{Charith Mendis}
\affiliation{%
  \institution{University of Illinois Urbana-Champaign}
  \country{USA}
}
\email{charithm@illinois.edu}


\begin{abstract}
Developing new abstract interpretation-based DNN certifiers is difficult
because their mathematical design is expressed at the neuron level, while
their implementations must operate at the tensor level. Existing libraries
such as auto\_LiRPA, IVAN, and ERAN support only a fixed set of certifiers;
extending them requires expertise in both formal methods and tensor
programming. Domain-specific languages such as \cf allow researchers to
specify certifiers at the neuron level, but cannot run them on real DNNs.

We present \compiler, a framework that automatically translates neuron-level certifier
specifications into tensor-level executables, making it practical to
design and run new DNN certifiers without manual tensor programming. \compiler
is enabled by three key contributions. First, a novel stack-based intermediate
representation (IR) paired with a shape analysis that infers the implicit
tensor operations: broadcasts, dimension expansions, and reductions, that are
absent from neuron-level specifications but required for correct tensor
execution. 
Second, the IR supports domain-specific rewrite rules that optimize common sequences of tensor operations arising during layer-level translation by rewriting them into more efficient equivalents.
Third, \sparse is a novel sparse tensor format of
variable-sized, layer-aligned blocks over a non-zero default value, tailored to
the sparsity of DNN certifiers. We prove that the translation
preserves the neuron-level semantics.

We evaluate \compiler on 8 certifiers and 32 DNN benchmarks. Beyond
compiling four widely-used existing certifiers, \compiler produces the first
executable implementations of four more: two previously specified but never
run, and two new designs. These expose precision--runtime trade-offs that no
existing library offers; for instance, the new SkipPoly certifier matches
DeepPoly's precision on all 32 DNNs while being as fast or faster on half of
them. On existing certifiers, \compiler's runtime is competitive with
hand-optimized libraries.

Our code is available at \url{https://github.com/uiuc-focal-lab/constraintflow.git}

\end{abstract}






\maketitle

\section{Introduction}
Deep Neural Networks (DNNs) are increasingly deployed in diverse applications, including safety-critical domains such as healthcare~\cite{AMATO201347} and autonomous driving~\cite{driving}. Therefore, ensuring these models behave reliably under adversarial or unexpected inputs has become a key challenge.
Abstract interpretation~\cite{absi} has emerged as one of the most practical and scalable
approaches to DNN formal certification~\cite{deeppoly,deepz,zhang2018crown,alphacrown,wang2021beta,ftpl}.
Rather than enumerating inputs, an abstract interpretation-based certifier
propagates overapproximations of sets of values through the DNN layer
by layer. If the resulting overapproximation at the output satisfies the
property, certification succeeds with a guarantee that covers every input in
the region simultaneously. This approach has been used to certify several properties
of DNNs, providing guarantees that testing cannot.

Building a new abstract interpretation-based DNN certifier, however, requires
simultaneous expertise in distinct areas: \emph{formal methods}
(designing abstract domains and proving soundness of abstract transformers);
\emph{DNN architecture knowledge} (understanding how operations compose and
how weight structure interacts with abstract values), and
\emph{tensor programming} (implementing the certifier as
efficient tensor operations). Each area alone is a substantial
research field; combining them is non-trivial. As a result, machine learning practitioners
who need certified models rely primarily on testing, and formal
certification remains accessible only to the few specialists. 
In this work, we aim to automate the development of abstract
interpretation-based DNN certifiers, making them accessible to practitioners
beyond this small group of specialists. 
Two classes of existing tools attempt to lower
this barrier, each with fundamental limitations.

\subsection{Existing Work}
\textbf{Libraries} such as auto\_LiRPA~\cite{lirpagithub}, IVAN~\cite{ivangithub}, and ERAN~\cite{eran} implement DNN certifiers for a fixed set of abstract domains and abstract transformers. This is limiting because different abstract domains trade off precision and scalability differently, and no single domain works best in every setting. At one extreme, an interval-based domain~\cite{interval} constrains each neuron by a single scalar lower and upper bound; it scales to large DNNs but produces loose over-approximations. At the other extreme, polyhedral domains~\cite{deeppoly,zhang2018crown} associate each neuron with bounds expressed as affine combinations of all other neurons in the DNN, producing much tighter over-approximations at significantly greater computational cost. Between these extremes lie zonotope-based~\cite{deepz} and hybrid domains~\cite{crownibp}, each with its own precision--scalability profile. The most suitable domain depends on the application, the DNN architecture, and the available compute budget. In principle, practitioners should be able to choose, or \textit{design} abstract domains that best match their requirements. In contrast, libraries expose only a fixed, predetermined set of domains and certifier algorithms. Adding a new domain, modifying an abstract transformer, or combining existing certifiers typically requires implementing the functionality from scratch or modifying library internals. Libraries thus offer efficient execution, but only for the certifiers they implement.

\textbf{Domain-specific languages} (DSLs) provide a promising alternative. They allow users to specify abstract domains and transformers without writing low-level tensor code, offering first-class support for the data types and operations that commonly arise in abstract interpretation for DNNs. As a result, certifiers can be expressed in a style that closely matches their presentation in the literature: as operations over individual neurons and their abstract values, processed one neuron at a time. ConstraintFlow~\cite{constraintflowsas} is one such DSL, enabling users to specify DNN certifiers declaratively while abstracting away implementation details, and verify their soundness~\cite{constraintflow}. However, \cf does not provide a way to execute a specified certifier on a real DNN.
To run a certifier, users must still manually derive and implement the tensor-level program.

Table~\ref{tab:landscape} summarizes this gap. Libraries can run certifiers on real DNNs but cannot be extended without tensor-programming expertise; DSLs make certifiers easy to specify and extend but cannot run them. To bridge this gap, we propose \compiler, a framework that automatically translates neuron-level specifications into tensor-level code. However, performing this translation is non-trivial and requires addressing the following key challenges.

\subsection{Key Challenges}
Most abstract interpretation-based DNN certifiers are specified at the level of
neurons: given the abstract value associated with a neuron's inputs,
the certifier defines the abstract value of its output
~\cite{deeppoly,deepz}. This level of granularity provides a natural interface
for specifying, reasoning about, and proving the soundness of certifiers
~\cite{constraintflow}. However, efficient execution must operate over entire
layers as tensor computations~\cite{lirpagithub,ivangithub,eran}. This leads to a \textit{neuron-to-tensor semantic gap}.
 
Unlike traditional DNN inference, which propagates concrete data points through
the DNN, certifiers symbolically propagate abstract values representing constraints over the
input neurons. These values range from simple interval bounds to
high-dimensional polyhedra and zonotopes, with each domain defining abstract
transformers that describe their propagation through the DNN. Automatically
mapping these computations to tensor programs is non-trivial: the translation
must reason not only across the neuron dimension but also over the internal
structure of the diverse abstract domains, giving rise to the two technical challenges
described next.

\textbf{Implicit Tensor Operations.}
The translation of neuron-level specifications to corresponding tensor programs introduces additional tensor operations that are implicit in the original specification. These include
introducing new dimensions, broadcasting values across those dimensions,
repeating tensors, and reducing intermediate results. 
Since such operations are not explicit in the neuron-level specification, they must be inferred during translation.
For instance, consider a simple example of multiplying
a tensor of shape $[n]$ by one of shape $[n,n]$, an operation that arises
frequently in DNN certification. Standard broadcasting can align the first tensor
with the second either row-wise (treating it as $[1,n]$) or column-wise (treating
it as $[n,1]$). Both run without error, yet
they compute different results, and only one corresponds to the neuron-level
meaning. 
Existing tensor frameworks such as NumPy~\cite{numpy} and PyTorch~\cite{pytorch} resolve implicit broadcasting using a default rule regardless of the specification, which might silently produce incorrect results. However, in the case of DNN certifiers, where the user's specification is at the neuron level, the corresponding tensor-level semantics must be inferred for each expression in the specification. Because the correct choice depends on the role each dimension plays in the specification, no fixed default rule suffices.
We show a detailed example in \S~\ref{sec:lifting-ex}.

\textbf{Sparsity.}
Abstract interpretation-based DNN certifiers naturally produce sparse tensors, so a naive dense implementation is inefficient, and existing sparse tensor formats are a poor fit. For instance, in several certifiers, abstract values are defined in terms of
subsets of neurons (or auxiliary variables) from the DNN. When these
relationships are represented as tensors, values associated with the same layer
often participate together, while values from other layers are absent. This
creates block-structured sparsity patterns aligned with DNN layer boundaries,
with block sizes varying across layers.
This structure is not captured by standard sparse formats. COO~\cite{coo} and
CSR~\cite{csr} represent sparsity at the individual-entry level; BCSR~\cite{bcsr}
requires uniform block sizes. Moreover, these formats assume zero as the default
value for absent entries, whereas certifier tensors may use values such as
$-\infty$ or $+\infty$ as defaults. Exploiting this structure efficiently
therefore requires a representation tailored to DNN certification, which we
describe in \S\ref{sec:sparsity-ov}.

\begin{table}[t]
\centering
\small
\caption{Comparison of existing tool classes and \compiler across key
         capabilities. }
\label{tab:landscape}
\resizebox{\columnwidth}{!}{%
\setlength{\tabcolsep}{10pt}
\begin{tabular}{@{}lccc@{}}
\toprule
\textbf{Required Features} & \textbf{Libraries} & \textbf{Specification DSLs} & \textbf{Neuton} \\
 & {\small(e.g., auto\_LiRPA, ERAN, IVAN)} & {\small(e.g., \cf)} & {\small(This Work)} \\
\midrule
Execution on real DNNs                          & \cmark & \xmark & \cmark \\[3pt]
Supports user-defined certifiers                & \xmark & \cmark & \cmark \\[3pt]
Certifiers specified without tensor programming & \xmark & \cmark & \cmark \\[3pt]
Automatically derived tensor implementation    & \xmark & \xmark & \cmark \\
\bottomrule
\end{tabular}%
}
\end{table}

\subsection{This Work}
We present \compiler, the first system that automatically translates neuron-level specifications written in \cf into executable tensor-level code, closing the neuron-to-tensor semantic gap.
To infer implicit tensor operations, \compiler introduces a novel
\emph{stack-based intermediate representation (IR)} where each node carries
explicit \emph{metadata}: an intrinsic shape (the minimal shape of the
tensor at the single-neuron level) and a broadcast factor per dimension
(how much the tensor must be replicated when lifted to the full layer).
A \emph{shape analysis} uses this metadata to automatically insert
all required dimension expansions, broadcasts, and reductions, resolving
the ambiguities that the neuron-level specification leaves implicit.
The IR also supports \emph{domain-specific rewrite rules} that recognize common certifier computation patterns and replace them with more efficient tensor kernels, eliminating unnecessary intermediate tensor materializations.

To handle the block-structured sparsity of certifier tensors, \compiler
introduces \sparse (\emph{Generalized Block Compressed Sparse Representation}), a novel
sparse tensor format. \sparse organizes tensors into variable-sized blocks 
aligned to DNN layers, stores each block in a compact form matching its internal structure, and supports
an explicit non-zero default value. All tensor operations required during
certification are implemented directly on the \sparse representation.
Fig.~\ref{fig:pipeline} gives an end-to-end view of the \compiler pipeline.

We evaluate \compiler on eight certifiers of varying complexity and 32 DNN
benchmarks. Four are widely-used existing certifiers (IBP, DeepZ, DeepPoly,
and CROWN-IBP), for which \compiler faithfully reproduces the established
certification behavior. Of the remaining four, two were proposed in prior work~\cite{constraintflow} but had never
been executed on real DNNs, and two are new certifiers we design.
Together, these certifiers span a broad range of the precision--runtime
trade-off space. For certifiers with hand-optimized baselines,
\compiler-generated executables attain competitive performance. For the
remaining certifiers, no baseline exists, and \compiler provides the first
means of executing them on real DNNs. For example, our new SkipPoly certifier matches DeepPoly's precision on all 32 DNNs while being as fast or faster on half of them. Thus, \compiler enables exploring the runtime--precision trade-offs of new certifiers by editing a specification rather than an implementation.
Further, an ablation shows that \compiler's optimizations are essential: without \sparse and the domain-specific rewrites, a direct tensorization of the specification is, on average, $30\times$ slower and crashes or times out on many configurations that \compiler handles (\S\ref{sec:ablation}). Following are the key contributions:

\begin{itemize}

    \item We present \compiler, a framework that
    automatically translates neuron-level specifications into 
    tensor-level executables, making it practical to design and run new DNN
    certifiers without manual tensor programming. This is achieved via a stack-based IR, a novel shape analysis, and domain-specific rewrites. We prove that the translated code is semantically equivalent to the neuron-level specification (\S\ref{sec:compiler}).

    \item We develop \sparse, a novel sparse tensor format with variable-sized blocks,
    per-block structural compression, and non-zero defaults, designed for the
    structured sparsity patterns that arise in abstract interpretation-based DNN certifiers
    (\S\ref{sec:sparserepresentation}).

    \item We evaluate \compiler on eight certifiers and 32 DNN benchmarks, demonstrating
    that it enables new certifier designs previously impractical to
    implement, while achieving competitive performance with hand-optimized
    libraries on existing certifiers (\S\ref{sec:eval}).

\end{itemize}

\section{Background}
\label{sec:background}

We provide background on abstract interpretation-based DNN
certifiers and their abstract domains (\S\ref{sec:background-cert}), DeepPoly~\cite{deeppoly} 
(\S\ref{sec:background-deeppoly}), and \cf~\cite{constraintflowsas,constraintflow}, used for specifying certifiers (\S\ref{sec:cf-background}).

\subsection{Abstract Interpretation-Based DNN Certifiers}
\label{sec:background-cert}

A DNN is modeled as a Directed Acyclic Graph (DAG) whose nodes are
\emph{neurons} and whose edges represent weights. A neuron $\neuron$'s value is determined by a
\emph{DNN operation} $f$ applied to a set of \emph{previous} neurons $p$.
Primitive operations include addition, multiplication, and non-linear
activations (ReLU, Sigmoid, etc.); composite operations include affine
transformations (fully connected and convolutional layers) and
their compositions.
DNN certifiers check properties of the form $(\varphi, \psi)$: for all
inputs in the set $\varphi$, the network outputs must lie in $\psi$.
A common example property is \emph{local robustness}: all inputs
within a bounded perturbation of a given input yield the same classification.

Abstract interpretation-based certifiers associate each neuron $\neuron$ with an \emph{abstract shape} $s$ that over-approximates its value as a conjunction of constraints $\mathcal{P}(s, \neuron)$. These shapes are propagated through the DNN by applying \emph{abstract transformers} for each operation, producing updated shapes for the output neurons. The property is certified if the resulting shapes at the output layer lie entirely within $\psi$.
Contemporary certifiers commonly use three types of constraints.
\begin{enumerate}
    \item 
\emph{Concrete (interval) bounds} constrain each neuron $\neuron$ by
$l \leq \neuron \leq u$ with $l, u \in \mathbb{R} \cup \{-\infty, \infty\}$.
They are cheap to propagate and form the basis of IBP~\cite{interval} and
CROWN-IBP~\cite{crownibp}.

\item
\emph{Polyhedral expressions} associate each neuron $\neuron$ with affine
lower and upper bounds $L \leq \neuron \leq U$, where $L$ and $U$ are
affine combinations of other neurons in the network:
$e = c_0 + \sum_i c_i \neuron_i$.
These expressions are more precise than intervals because they track
relationships between neurons. DeepPoly~\cite{deeppoly} and
CROWN~\cite{zhang2018crown} are prominent examples.

\item
\emph{Symbolic (zonotope) expressions} represent neuron values
as $e = d_0 + \sum_i d_i \varepsilon_i$, where each $\varepsilon_i \in [-1, 1]$
is a \textit{symbolic noise variable}. Because the noise
variables are shared, these expressions capture correlations between
neurons. They are used in certifiers such as DeepZ~\cite{deepz}.
\end{enumerate}

A core computation in DNN certifiers is evaluating the lower (or upper) bound
of an expression $e = c_0 + \sum_i c_i \neuron_i$. Each coefficient $c_i$ determines which bound of $\neuron_i$ to use:
for a term $c_i \cdot \neuron_i$, the contribution
to the lower bound is $c_i \cdot l_{\neuron_i}$ if $c_i > 0$
and $c_i \cdot u_{\neuron_i}$ if $c_i < 0$, where $l_{\neuron_i}$ and
$u_{\neuron_i}$ are the concrete lower and upper bounds of $\neuron_i$.
The same sign-based selection applies in zonotope domains: the lower bound
of $e = d_0 + \sum_i d_i \varepsilon_i$ is obtained by substituting
$\varepsilon_i = -1$ when $d_i > 0$ and $\varepsilon_i = 1$ when $d_i < 0$.
This pattern of iterating over (term, coefficient) pairs and selecting a
tighter bound appears repeatedly across certifier types and domains.
%
%
Certifiers such as DeepPoly and
several of its variants~\cite{crownibp,zhang2018crown} further refine these bounds by traversing the DNN graph recursively (\S\ref{sec:background-deeppoly}).

\subsection{Running Example: DeepPoly}
\label{sec:background-deeppoly}

DeepPoly~\cite{deeppoly} is one of the most widely used abstract
interpretation-based certifiers. We use it as the running example throughout
this paper because it exercises both polyhedral bounds and graph traversal,
illustrating the key challenges that \compiler must address.
DeepPoly associates each neuron $\neuron$ with the abstract shape
$s_\neuron = \langle l_\neuron,\, u_\neuron,\, L_\neuron,\, U_\neuron \rangle$,
where $l_\neuron, u_\neuron$ are concrete bounds and
$L_\neuron$, $U_\neuron$ are polyhedral  bounds. The
constraints on $\neuron$ are:
$\mathcal{P}(s_\neuron,\, \neuron)
  \;\triangleq\;
  (l_\neuron \leq \neuron \leq u_\neuron)
  \;\wedge\;
  (L_\neuron \leq \neuron \leq U_\neuron).$

\paragraph{Affine transformer.}
For a neuron $\neuron_0$ computed by an affine operation with previous neurons
$\neuron_1, \ldots, \neuron_m$, weights $w_i$, and bias $b$, the polyhedral
bounds are set to:
  $L_{\neuron_0}' \;=\; U_{\neuron_0}' \;=\; b + \textstyle\sum_i w_i \neuron_i.$
The concrete bound $l_{\neuron_0}'$ is then obtained by
\emph{backsubstitution}: starting from $L_{\neuron_0}'$, each constituent
neuron $\neuron_i$ is recursively replaced by its own polyhedral lower bound
$L_{\neuron_i}$ (if $w_i \geq 0$) or upper bound $U_{\neuron_i}$ (if
$w_i < 0$). This substitution is applied layer by layer, traversing the DNN
in reverse, until all remaining neurons are input neurons whose concrete
bounds are then substituted in.

\paragraph{ReLU transformer.}
For $\neuron_0 = \max(0, \neuron_1)$, the transformer branches
on the concrete bounds of $\neuron_1$. If $l_{\neuron_1} \geq 0$, the neuron
is always active and $L_{\neuron_0}' = U_{\neuron_0}' = \neuron_1$. If
$u_{\neuron_1} \leq 0$, it is always inactive and
$L_{\neuron_0}' = U_{\neuron_0}' = 0$. Otherwise, it
introduces a linear relaxation parameterized by $l_{\neuron_1}$ and
$u_{\neuron_1}$.

\subsection{\cf: Neuron-Level Certifier Specification}
\label{sec:cf-background}

Since we use \cf~\cite{constraintflowsas,constraintflow} as
the specification language, we describe the constructs relevant to
our compilation. These constructs reflect operations that are common across
abstract interpretation-based DNN certifiers; \cf exposes them as first-class
language features. Certifiers in \cf are written at the level of a
representative neuron \curr.

\paragraph{Types.}
DNN certifiers use 3 categories of abstract values, described in
\S\ref{sec:background-cert}: scalars, polyhedral expressions over neurons,
and zonotope expressions over symbolic noise variables. \cf makes these
first-class types: \typereal for concrete scalars, \typepoly for polyhedral
expressions ($e = c_0 + \sum_i c_i \neuron_i$), and \typezono for zonotope
expressions ($e = d_0 + \sum_i d_i \varepsilon_i$). The abstract shape for
a neuron is a record of typed fields; for DeepPoly, this is
\texttt{L}, \texttt{U} of type \typepoly and \texttt{l}, \texttt{u} of
type \typereal.

\paragraph{Constructs.}
\cf provides \curr for the neuron being transformed in the current layer and \prev
for neurons in the preceding layer; field \texttt{F} of a neuron $\neuron$ is accessed as $\neuron$\texttt{[F]}.
Affine combinations of predecessors, the starting point of most
certifier computations, arise everywhere in DNN certification.
\cf exposes this via \prev.\dott(\curr\texttt{[w]}), which builds the \typepoly
expression $\sum_i w_i \neuron_i$ that assigns weight $w_i$ to each previous
neuron $\neuron_i$.
%
\cf also provides the \map construct to specify the bound-computation pattern from \S\ref{sec:background-cert}.
\texttt{e.\map(f)} applies
\texttt{\func f(\typeneuron n, \typereal c)} to each pair $(\neuron_i, c_i)$ in
expression $e$ and sums the results.
For example, \texttt{f(n, c) = (c > 0) ? c * n[l] : c * n[u]} implements the
sign-based lower-bound computation of \S\ref{sec:background-cert}, selecting
each neuron's concrete lower or upper bound according to the sign of its
coefficient.
%
Graph traversal for backsubstitution, described in \S\ref{sec:background-deeppoly},
is another operation common across certifiers.
\cf exposes it via \traverse: \texttt{e.\traverse(dir, stop, priority, f)}
walks the DNN in direction \texttt{dir}, applying \texttt{f} at each
step in the order of \texttt{priority} until \texttt{stop} condition.
In DeepPoly's affine transformer, \traverse drives backsubstitution by
repeatedly replacing each constituent neuron with a function of its own
polyhedral bound.

All the constructs above are specified one neuron at a time. However, a DNN layer can contain thousands of neurons, all subject to the same transformer. Although \cf provides operational semantics for certifiers by formally defining the abstract value of each neuron, applying these semantics directly via a neuron-by-neuron execution is computationally prohibitive. Efficient implementations must instead operate on entire layers at once using tensor operations. 

\section{Overview}
\label{sec:overview}

\compiler automatically translates neuron-level certifier specifications into
efficient tensor-level executables. 
Following, we present a \cf snippet specifying part of the DeepPoly certifier, which serves as our running example. 
The full DeepPoly certifier is specified in about 20 lines of \cf.
It begins with a record defining the abstract shape (the tuple
$\langle l, u, L, U \rangle$ per neuron), followed by user-defined functions
for bound computation, and concludes with the abstract transformers for each
DNN operation type. The complete specification appears in prior
work~\cite{constraintflow}. For this overview, we focus on a small but
central code snippet, shown in Fig.~\ref{fig:cf-code}, extracted from the affine transformer.
Lines~1-3 show a part of the affine transformer: \texttt{f1}
selects the lower or upper polyhedral bound of each neuron depending on
the sign of the coefficient; \texttt{g1} drives backsubstitution by
calling \traverse on \prev.\dott(\curr\texttt{[w]}~+~\curr\texttt{[b]}), walking
backwards through the DNN until only input neurons remain.
Lines~5-7 show a simplified variant, used as our running example, which
replaces \traverse with \map, always uses the lower bound, and omits the bias
term, to avoid complexity. 
The operational semantics from prior work~\cite{constraintflow} process one neuron at a
time. However, directly running it on a real DNN is impractical.
This motivates the need for a \emph{compiler} that compiles the neuron-level specification into efficient tensor operations.
\begin{figure}[t]
\begin{lstlisting}[numbers=left]
Func f1(Neuron n, Real c) = c >= 0 ? (c * n[L]) : (c * n[U]);
Func g1(PolyExp e) = e.traverse(backward, ..., f1);
  ... g1(prev.dot(curr[w]) + curr[b]) ...
// Simplified variant (non-negative weights; used as running example)
Func f(Neuron n, Real c) = (c * n[L]);
Func g(PolyExp e) = e.map(f);
  ... g(prev.dot(curr[w])) ...
\end{lstlisting}
\caption{Simplified version of neuron-level expression from DeepPoly}
\label{fig:cf-code}
\end{figure}

This section
illustrates the challenge of
translating a neuron-level certifier expression into tensor operations
(\S\ref{sec:lifting-ex}),
describes the \compiler compilation pipeline (\S\ref{sec:pipeline-ov}),
motivates the \sparse sparse format (\S\ref{sec:sparsity-ov}), and states the
correctness of the compilation and the generality it enables (\S\ref{sec:new-certs-ov}). 


\subsection{The Neuron-to-Tensor Lifting Challenge}
\label{sec:lifting-ex}

The neuron-level \cf expressions in Fig.~\ref{fig:cf-code} are, in
principle, tensorizable. Polyhedral expressions are coefficient vectors over
neurons and can be stored as tensors. The \dott construct assembles affine
combinations, and \map iterates over (neuron, coefficient) pairs to produce a
scalar: both operations have natural tensor analogues. Moreover, since a \cf
specification is written for one representative neuron \curr, the same
computation applies uniformly to every neuron in a layer, creating a
\emph{tensorization opportunity}: process all $m$ neurons in the current layer
together, rather than one at a time.
However, realizing this opportunity is far from straightforward.

We trace the single expression \texttt{g(}\prev.\dott(\curr\texttt{[w]}))
through two cases: first for a
single neuron, then for the entire layer. The computation exposes the
decisions that \compiler must make. Fix a DNN with $n$ total neurons and a
fully-connected layer with $m$ neurons
in the current layer and $p$ neurons in the previous layer. We use $x$-axis to denote the dimension for current-layer
neurons, $y$-axis for the DNN neuron index for \map, and $z$-axis
for the polyhedral coefficient index.

\begin{figure}[t]
\centering
\begin{subfigure}[t]{0.40\columnwidth}
  \centering
  \includegraphics[width=\linewidth,trim={0 1cm 0 1.3cm},clip]{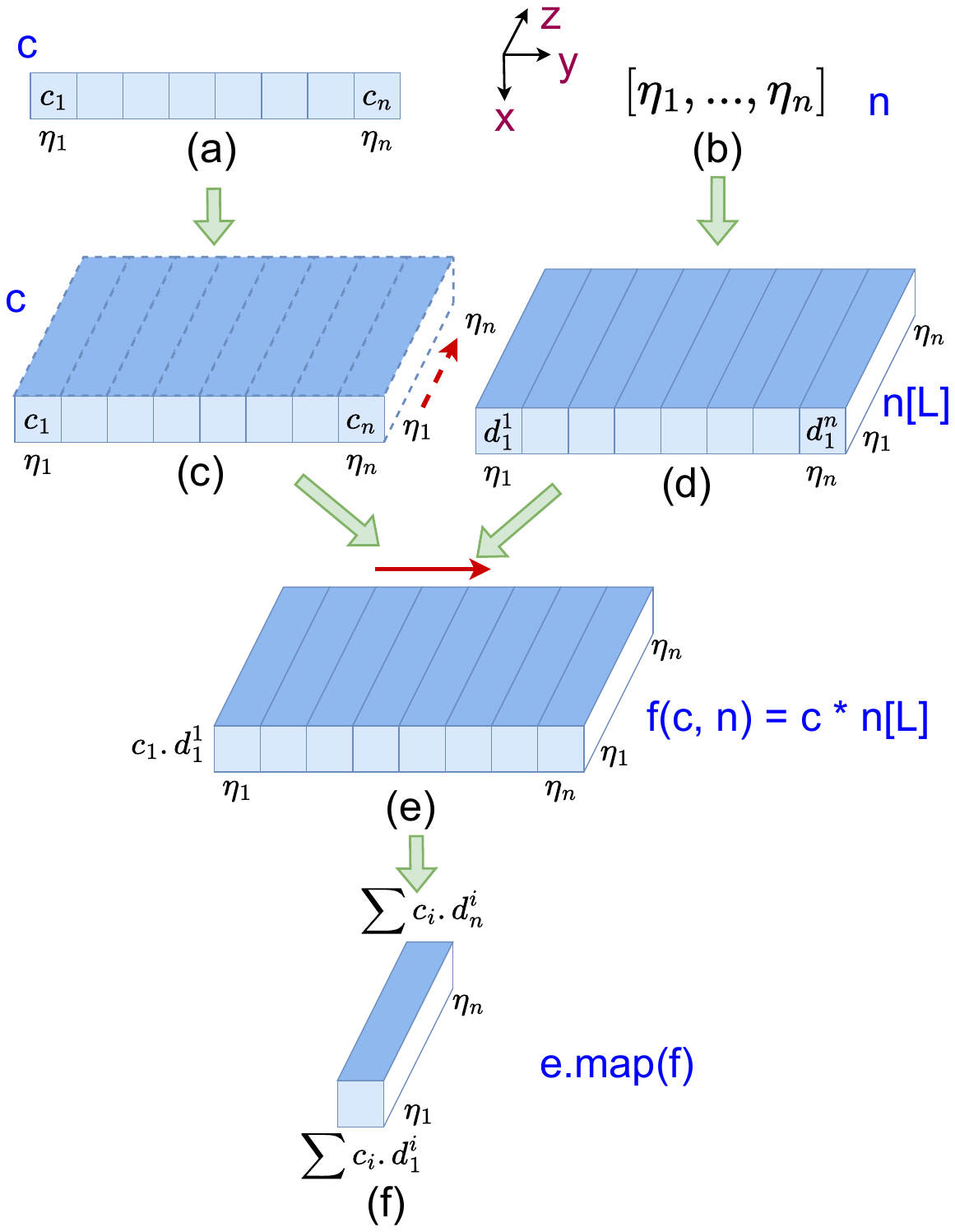}
  \caption{Single neuron. (a)~$v\;[n]$. (b)~Neurons. (c)~$v$ repeated $[n,n]$.
           (d)~$\mathbf{L}\;[n,n]$. (e)~Product $[n,n]$. (f)~Result $[n]$.}
  \label{fig:tensor-single}
\end{subfigure}
\hfill
\begin{subfigure}[t]{0.56\columnwidth}
  \centering
  \includegraphics[width=\linewidth,trim={0cm 5.5cm 0cm 6.0cm},clip]{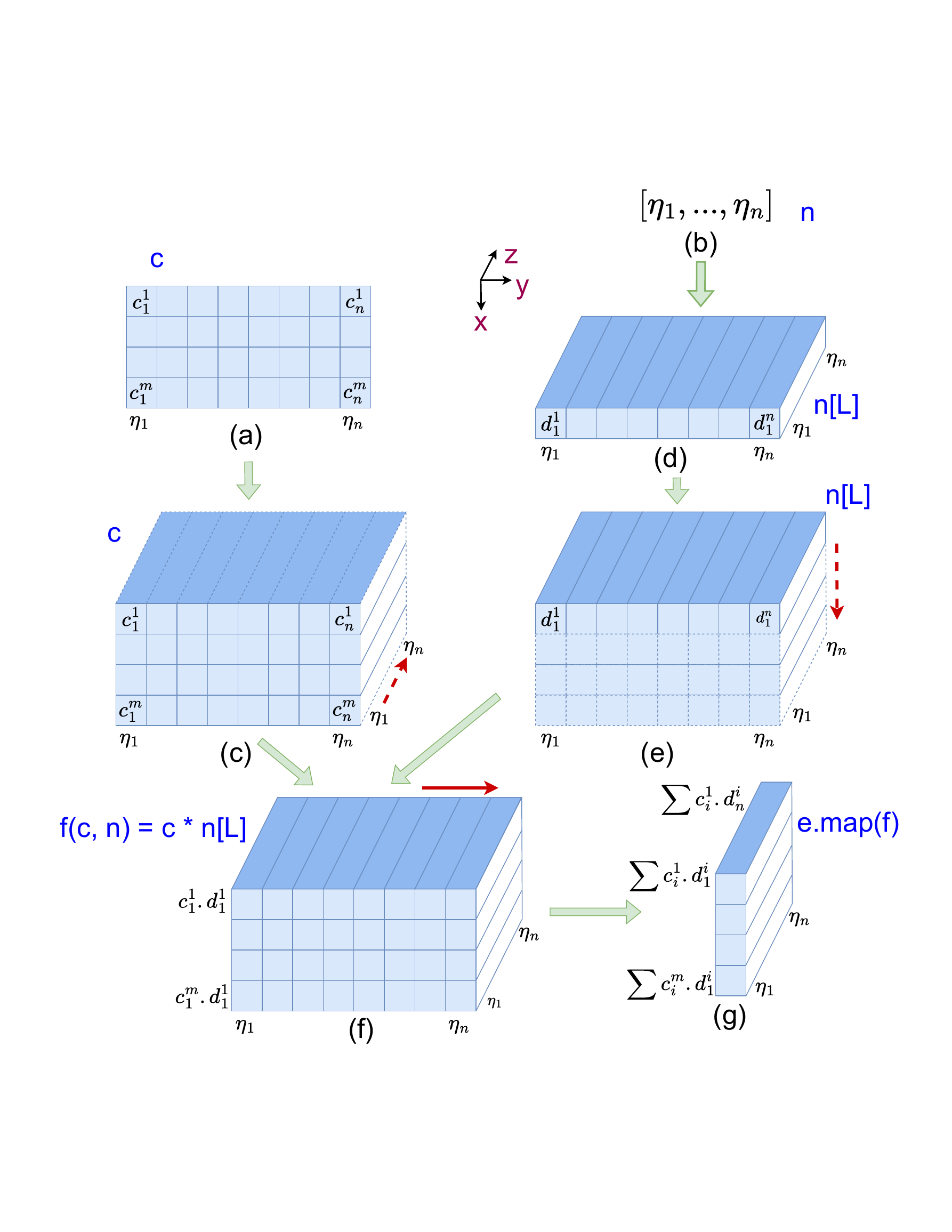}
  \caption{All $m$ neurons. (a)~$V\;[m,n]$. (b)~Neurons. (c)~$V$ repeated $[m,n,n]$.
           (d)~$\mathbf{L}\;[n,n]$. (e)~$\mathbf{L}$ repeated $[m,n,n]$.
           (f)~Product $[m,n,n]$. (g)~Result $[m,n]$.}
  \label{fig:tensor-layer}
\end{subfigure}
\caption{Tensorized computation of \texttt{g(}\prev.\dott(\curr\texttt{[w]})).}
\label{fig:tensor-both}
\end{figure}

\paragraph{Single-neuron case.}
Consider one current neuron $\neuron_0$.
Its weight vector \curr\texttt{[w]} is a 1D tensor of shape $[p]$. \dott assembles a polyhedral expression
$v = \sum_{i=1}^{p} w_i \cdot \neuron_i$,
assigning coefficient $w_i$ to previous neuron $\neuron_i$.
This polyhedral value $v$ is exactly the argument $e$ that gets passed to $g$ at
line~6 of Fig.~\ref{fig:cf-code}; we now trace how $g(e) = e.\map(f)$ must
execute as tensor operations.
Although only $p$ weights are non-zero, $v$ must in principle track a
coefficient for every one of the $n$ neurons in the DNN, because a polyhedral
expression can involve any neuron. Stored as a tensor, $v$'s coefficients therefore form a 1D array of
shape $[n]$, with the $n - p$ entries for neurons outside the previous layer
set to zero (Fig.~\ref{fig:tensor-single}(a)).
In practice this yields highly sparse tensors; we return to this in
\S\ref{sec:sparsity-ov}.

Applying \texttt{g(v)~=~v.}\map\texttt{(f)} evaluates
$f(\neuron, c) = c \cdot \neuron[\texttt{L}]$ for each (neuron, coefficient)
pair in $v$ and sums the results:
  $g(v) \;=\; \textstyle\sum_{i=1}^{p} w_i \cdot L_{\neuron_i}$
Here $L_{\neuron_i}$ is the polyhedral lower bound of $\neuron_i$, itself a
polyhedral expression: a 1D array of coefficients of shape $[n]$.
Fig.~\ref{fig:tensor-single}(b) shows the neurons, each carrying its own
polyhedral lower bound as an $[n]$-vector; stacking all $n$ of these vectors
forms $\mathbf{L}$ (Fig.~\ref{fig:tensor-single}(d)), an $n \times n$ tensor.
To evaluate the sum:

\begin{enumerate}[topsep=2pt,itemsep=0pt,label=(\arabic*)]
\item \textbf{Repeat} $v$ from Fig.~\ref{fig:tensor-single}(a)
      $n$ times along the $z$-axis to obtain shape $[n, n]$
      (Fig.~\ref{fig:tensor-single}(c)).
\item \textbf{Multiply} element-wise with $\mathbf{L}$ (shape $[n,n]$,
      indexed by $y$ and $z$): shape $[n, n]$
      (Fig.~\ref{fig:tensor-single}(e)).
\item \textbf{Reduce} (sum) along the $y$-axis: shape $[n]$
      (Fig.~\ref{fig:tensor-single}(f)).
\end{enumerate}

Although none of these tensor operations appear anywhere in a neuron-level
certifier description, steps (1)--(3) effectively compute
$g(v) = \mathbf{L}^\top w$: a matrix-vector product, emerging entirely from
the compilation of a scalar specification.
The repeat--multiply--reduce sequence that collapses into a matrix product is
one of several domain-specific patterns that frequently arise when converting
abstract interpretation-based DNN certifier specifications into tensor
operations. \compiler recognizes such domain-specific patterns and exploits them automatically.

\paragraph{Whole-layer case.}
A DNN layer contains not one but $m$ neurons, all subject to the same
transformer. Processing them one at a time is impractical; an efficient
implementation must apply the transformer to all $m$ neurons simultaneously.
When this is done,
\curr\texttt{[w]} becomes a 2D weight matrix of shape $[m, p]$, and
\prev.\dott(\curr\texttt{[w]}) produces $m$ polyhedral values
$v_1, \ldots, v_m$.
As in the single-neuron case, each polyhedral value must carry a coefficient
for every one of the $n$ DNN neurons, yielding a 2D tensor $V$ of shape
$[m, n]$ where only the $p$ columns corresponding to the previous layer are
non-zero (Fig.~\ref{fig:tensor-layer}(a)).
The $x$-axis indexes the $m$ current neurons and the $y$-axis indexes the
$n$ DNN neurons.
The lower-bound matrix $\mathbf{L}$ is unchanged: the
polyhedral bounds of the previous neurons are the same for all $m$ current
neurons.
Lifting to the entire layer is not achieved by simply adding another dimension to every tensor. Instead, the choice of which axis to expand for each tensor, and the choice of which axis to reduce, changes in a nontrivial manner.

\begin{enumerate}[topsep=2pt,itemsep=0pt,label=(\arabic*)]
\item \textbf{Repeat} $V$ (shape $[m,n]$) $n$ times along the $z$-axis to
      obtain $[m,n,n]$ (Fig.~\ref{fig:tensor-layer}(c)).
\item \textbf{Repeat} $\mathbf{L}$ (shape $[n,n]$, with $y$-axis for DNN
      neurons and $z$-axis for coefficients) $m$ times along the $x$-axis to
      obtain $[m,n,n]$ (Fig.~\ref{fig:tensor-layer}(e)).
      \emph{This step has no counterpart in the single-neuron case: $\mathbf{L}$
      was a fixed $[n,n]$ matrix there, but here it must be tiled $m$ times
      along the $x$-axis to align with the $m$ current neurons in $V$.}
\item \textbf{Multiply} element-wise: shape $[m,n,n]$
      (Fig.~\ref{fig:tensor-layer}(f)).
\item \textbf{Reduce} (sum) along the $y$-axis: shape $[m,n]$
      (Fig.~\ref{fig:tensor-layer}(g)).
\end{enumerate}

Although again absent from any neuron-level certifier description, steps
(1)--(4) effectively implement $G(V) = V \cdot \mathbf{L}^\top$: a
matrix product, inferred from compilation of the specification.
%
Further, certifiers are typically applied to a batch of $b$ inputs simultaneously,
introducing yet another dimension. Again, the axis choices are
non-trivial: tensors whose values differ per input (such as $V$) gain a
leading $b$-dimension, while tensors that depend only on the DNN structure
(such as $\mathbf{L}$) do not, and all previously chosen axes shift
accordingly. We omit the derivation, but the same challenges of dimension insertion and alignment recur at
every scale.

\paragraph{The compilation problem.}
The neuron-level expression \texttt{g(}\prev.\dott(\curr\texttt{[w]})) contains
none of the tensor operations derived above. \compiler must determine all of
them from the neuron-level specification alone. Three challenges arise:

\begin{itemize}[topsep=2pt,itemsep=0pt]
\item \emph{Tensor axes.}
  Which axis to insert when expanding each tensor, which axis to reduce, and
  how these choices shift when lifting from a single neuron to a full layer to
  a batch: none of this is specified by the neuron-level description.
  The correct choices depend on the semantic role of each dimension, and an incorrect choice
  silently produces unsound results.
\item \emph{Rewrite opportunities.}
  Executing the inferred sequence directly materializes large intermediate
  tensors (e.g., the $[m,n,n]$ product above). Many such sequences can instead
  be collapsed into a single efficient primitive; the repeat--multiply--reduce
  sequence above, for instance, is exactly a \texttt{matmul}. \compiler applies
  such rewrites using the shape and broadcast fields inferred during shape
  analysis, and we prove them correct in \S\ref{sec:rewrites}.
\item \emph{Sparsity.}
  The coefficient tensors (such as $\mathbf{L}$) are highly sparse: the bound
  of a neuron at any intermediate step involves only neurons in the layers
  traversed so far, leaving most of the $n \times n$ tensor at its default
  value. We design a
  novel sparse tensor format that captures the layer-aligned block sparsity of
  certifier tensors and supports a non-zero default value.
\end{itemize}

These challenges are not addressed by any existing framework.
Tools such as PyTorch~\cite{pytorch}, NumPy~\cite{numpy},
Halide~\cite{halide}, XLA~\cite{xla} are good at optimizing tensor computations, but they take tensor
programs as input: the programmer must already know which tensor operations
to perform and how to shape them. None of these frameworks attempts to
automatically translate a neuron-level description into the tensor
operations it implies. \compiler attempts to solve this problem.

\begin{figure}[t]
\centering
\vspace{-5mm}
\includegraphics[width=0.8\columnwidth,trim={0.5in 2.4in 3.0in 1.2in},clip]{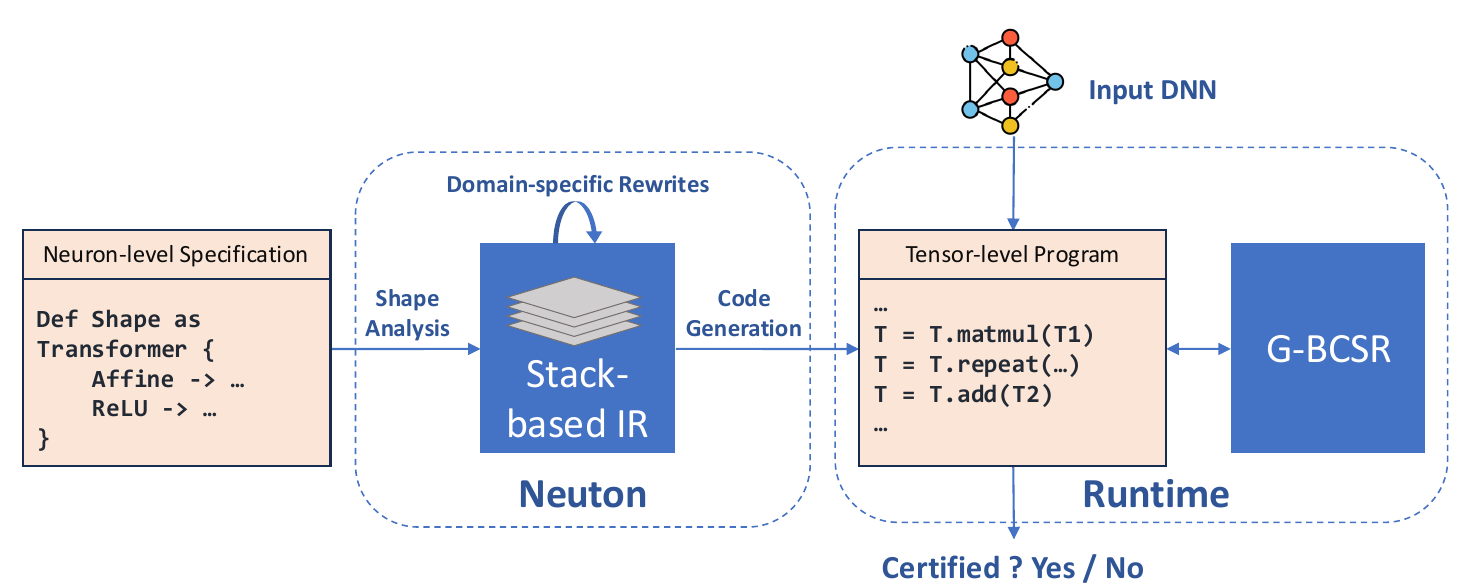}
\caption{Overview of the compilation pipeline.}
\vspace{-5mm}
\label{fig:pipeline}
\end{figure}

\subsection{The \compiler Compilation Pipeline}
\label{sec:pipeline-ov}

\compiler addresses the three challenges identified in \S\ref{sec:lifting-ex}. Fig.~\ref{fig:pipeline} shows the compilation pipeline. A neuron-level \cf specification enters on the left; the \compiler front end parses it and, using \emph{shape analysis}, lowers it to a \emph{stack-based tensor IR} in which each node carries metadata describing its tensor shape and layout, with all implicit operations made explicit. A \emph{rewrite-rule optimizer} then fuses redundant sequences into efficient primitives. Finally, \emph{code generation} emits Python code that, at runtime, operates over a novel \sparse sparse tensor backend rather than plain dense PyTorch tensors; this backend is the subject of \S\ref{sec:sparsity-ov}.


\compiler parses the \cf specification and builds a \emph{stack-based IR}.
Each IR node is annotated with \emph{metadata} that tracks the node's type (e.g., $\typepoly$, $\typereal$, $\typeneuron$),
its \emph{intrinsic shape} (the minimal tensor shape needed for execution) 
and its \emph{broadcast factor} 
(how many times each dimension must logically be replicated when lifted to the
full layer or batch). The formal definition is in \S\ref{sec:compiler}.
For example, the neuron list in the whole-layer case has intrinsic shape
$[1, n]$ and broadcast factor $[m, 1]$, while the weight tensor has intrinsic
shape $[m, n]$ and broadcast factor $[1, 1]$.

\emph{Shape analysis} resolves the axis-determination problem from
\S\ref{sec:lifting-ex}.
When two IR nodes are combined, shape analysis aligns their broadcast
factors: a dimension with factor $m$ in one operand and $1$ in the other
is explicitly repeated $m$ times, and the correct axis is determined by the
\emph{position} of that factor in the metadata vector, not guessed from raw
tensor shape.
This is what decides that $\mathbf{L}$ expands along the $x$-axis (not $z$)
in the whole-layer case: its broadcast factor has $m$ at position 0.

\begin{figure}[t]
\centering
\includegraphics[width=\columnwidth,trim={1cm 38cm 2cm 3cm},clip]{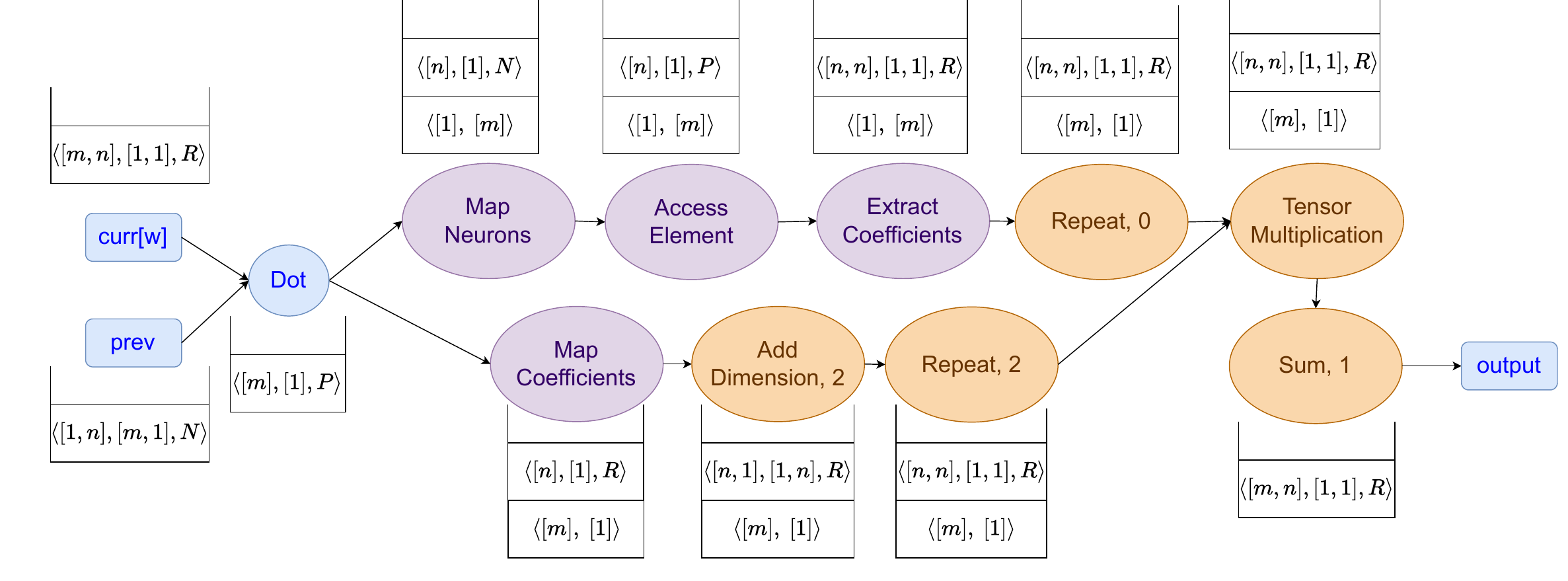}
\caption{\compiler IR for \texttt{g(}\prev.\dott(\curr\texttt{[w]})) applied to $m$ neurons.
Each node carries stack-based metadata. Each element of the stack is $\langle \text{intrinsic shape}, \text{broadcast factor} \rangle$. The top element also records the type of the expression.
Purple nodes arise from the \map construct; orange nodes are tensor operations
inserted automatically by shape analysis.
$P \equiv \typepoly$, $R \equiv \typereal$, $N \equiv \typeneuron$.}
\label{fig:irfinal}
\end{figure}

The stack structure becomes essential for the \map construct.
\map applies a user-defined function $f$ to each
(neuron, coefficient) pair in a polyhedral expression and sums the results,
introducing an intermediate tensor dimension that must ultimately be
reduced away. The challenge is that $f$ may itself introduce additional
dimensions, making it non-trivial to identify \emph{which} dimension to
reduce.
\compiler handles this by maintaining a stack of metadata elements: upon
entering a \map, the current metadata frame is pushed; upon exiting, it is
popped. The popped frame identifies exactly the dimension that the current
\map introduced, so the reduction is applied to the correct axis regardless
of what $f$ does internally or whether \map constructs are nested.
Fig.~\ref{fig:irfinal} shows the resulting IR for
\texttt{g(}\prev.\dott(\curr\texttt{[w]})) applied to $m$ neurons.

\begin{figure}[t]
\begin{pylisting}
    W     = curr.get_metadata('weight')   # curr[w]
    e     = prev.dot(W)                   # Dot
    C     = e.get_mat(abs_elem)           # Map Coefficients
    L     = abs_elem.get_elem('L')        # Map Neurons, Access Element
    Lm    = L.get_mat(abs_elem)           # Extract Coefficients
    # g(e) = e.map(f),  f(n, c) = c * n[L] :
    |\tikzmark{bt}|Cd    = C.unsqueeze(2)                # Add Dimension, 2
    Cr    = repeat(Cd, axis=2)            # Repeat, 2
    Lr    = repeat(Lm, axis=0)            # Repeat, 0
    prod  = binary(Cr, Lr, operator.mul)  # Tensor Multiplication|\tikzmark{re}|
    |\tikzmark{bb}|coeff = prod.sum(axis=1)              # Sum, 1
\end{pylisting}
\begin{tikzpicture}[remember picture, overlay]
  \coordinate (bx)  at ([xshift=1.5mm]pic cs:re);
  \coordinate (tbt) at (pic cs:bt);
  \coordinate (tbb) at (pic cs:bb);
  \draw[decorate, decoration={brace, amplitude=4pt}, thick, gbcsrcall]
    ([yshift=1.7ex]bx |- tbt) -- ([yshift=-0.4ex]bx |- tbb)
    node[midway, right=3pt, align=left, black]
      {\small$\Longrightarrow$ \texttt{\textcolor{gbcsrcall}{inner\_prod}(C, Lm)}\\
       {\scriptsize\itshape fused by rewrite R1 (\S~\ref{sec:rewrites})}};
\end{tikzpicture}
\caption{Code generated by \compiler corresponding to the IR in Fig.~\ref{fig:irfinal}. \textcolor{libcall}{Purple} calls are \cf-specific (operations on polyhedral expressions); \textcolor{gbcsrcall}{orange} calls are the \sparse tensor
API. 
}
\label{fig:gencode}
\end{figure}

\paragraph{Domain-specific rewrites.}
Before code generation, \compiler applies domain-specific rewrite rules to the
IR, in addition to standard passes such as common subexpression elimination and
loop-invariant code motion. The most important one applies whenever shape analysis produces a
repeat--multiply--reduce sequence that contracts over a shared axis: \compiler
collapses the whole sequence into a single \texttt{matmul} call, avoiding the
large intermediate tensors that the sequence would otherwise materialize. In the running example, this reduces the
three-step (or four-step) sequence from \S\ref{sec:lifting-ex} to one matrix
multiplication, which state-of-the-art tensor libraries execute highly
efficiently. 

\paragraph{Code generation.}
Code generation lowers the optimized IR to Python.
Because \compiler compiles a certifier specification independently of any
particular DNN, the generated 
executable runs on any DNN without recompilation.
The emitted code calls into a runtime backend built on \sparse, our novel
sparse tensor representation, rather than plain PyTorch tensors. At runtime, \sparse tensors are constructed from the DNN's
weight matrices and all certified operations proceed on the compressed
representation.
Fig.~\ref{fig:gencode} shows the code generated for the running example. Every IR
node is lowered to either calls to our DNN-certification
library or calls to the generic \sparse tensor API. Applying the rewrite rules
collapses lines 8-12 into a
single \texttt{inner\_prod} call. 

\subsection{\sparse: Generalized Block Compressed Sparse Representation}
\label{sec:sparsity-ov}

\begin{figure}[t]
\centering
\hspace{0.1\columnwidth}
\begin{minipage}[t]{0.20\columnwidth}
  \centering
  \includegraphics[width=\linewidth,trim={4cm 1cm 10cm 2cm},clip]{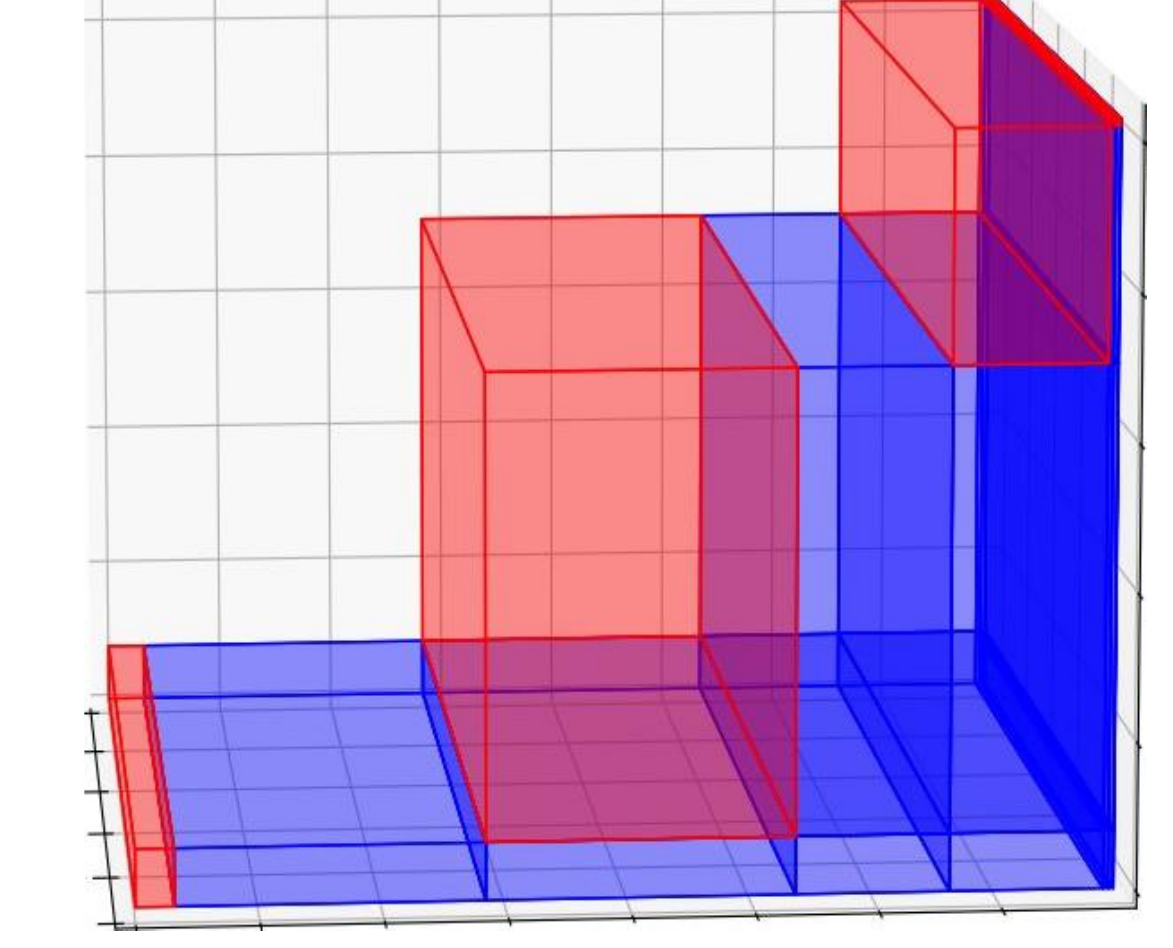}
\end{minipage}
\hfill
\begin{minipage}[t]{0.20\columnwidth}
  \centering
  \includegraphics[width=\linewidth,trim={6cm 1cm 8cm 2cm},clip]{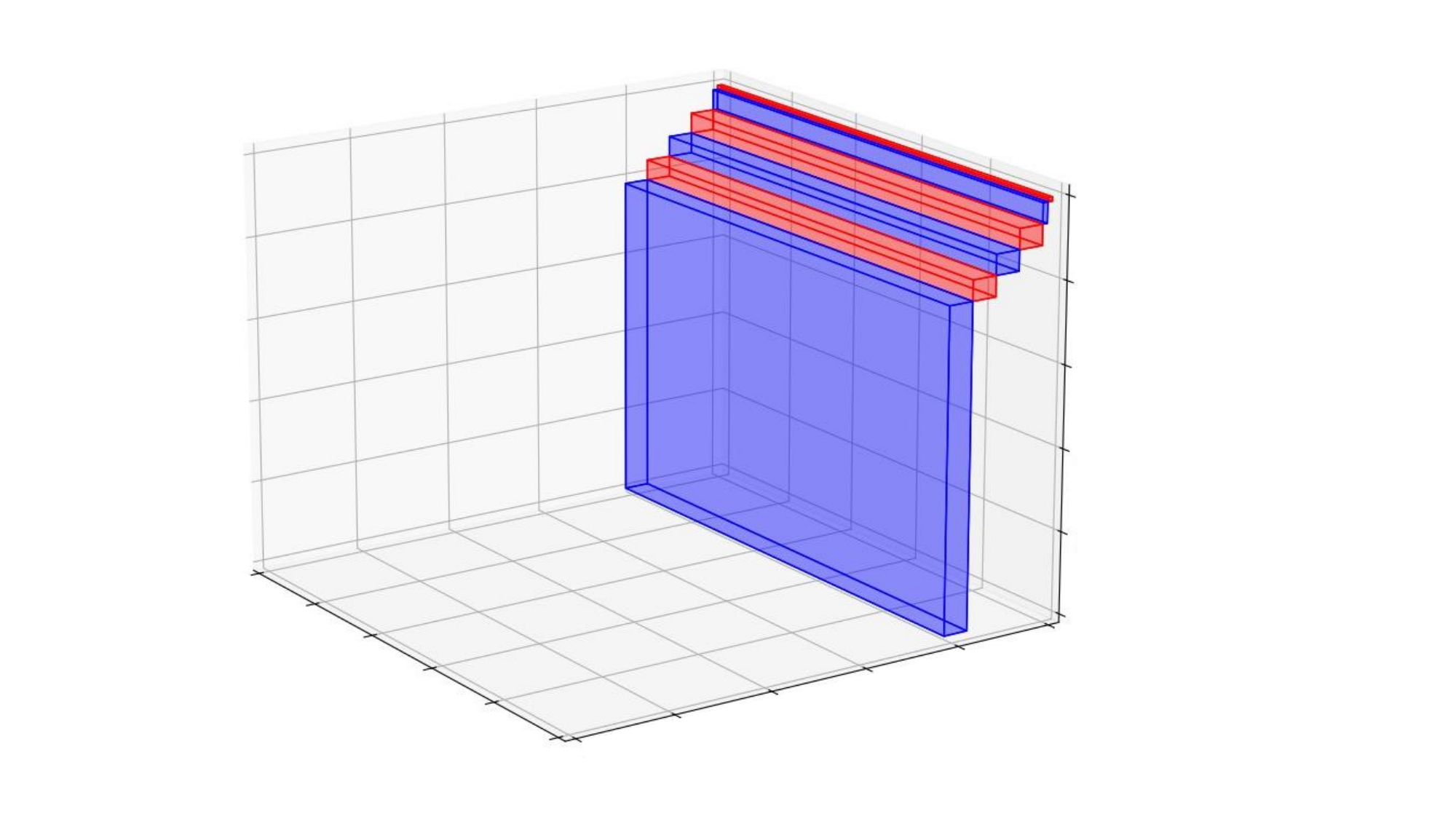}
\end{minipage}
\hfill
\begin{minipage}[t]{0.20\columnwidth}
  \centering
  \includegraphics[width=\linewidth,trim={7cm 2cm 6cm 2cm},clip]{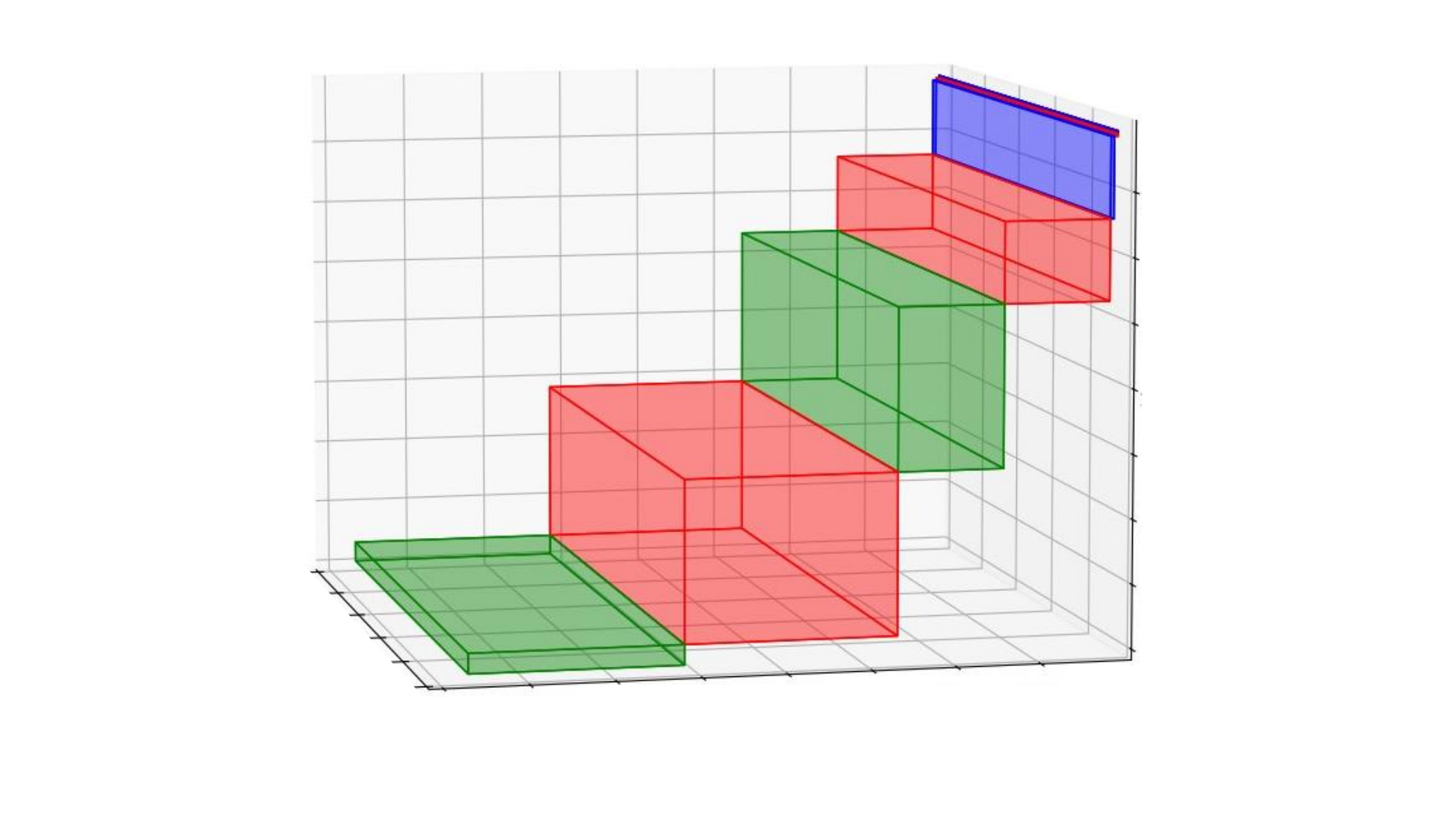}
\end{minipage}
\hspace{0.1\columnwidth}
\caption{Examples of tensor sparsity patterns observed across
different certifiers and DNN architectures.
Blue = dense block (fully connected layer); red = diagonal block (ReLU);
green = stencil block.}
\label{fig:sparsity}
\end{figure}

The lower-bound matrix $\mathbf{L}$ introduced in \S\ref{sec:lifting-ex}
is not dense in practice.
For instance, during DeepPoly's backsubstitution, the bound of a given neuron accumulates
coefficients only for neurons in the layers traversed so far; the rest are
zero. This creates a \emph{layer-aligned block structure}: the non-zero
values of $\mathbf{L}$ form contiguous blocks, one per DNN layer, with
widths equal to the layer widths.
Fig.~\ref{fig:sparsity} shows three examples of this structure observed
across different certifiers and DNN architectures. The exact pattern varies:
it depends on which certifier is used, which DNN layers are involved, and how
far backsubstitution has proceeded. What is common across all cases is that
the sparsity is \emph{block-structured with variable block sizes}, and the
blocks themselves have internal structure (dense for fully-connected layers,
diagonal for ReLU, kernel-shaped for convolutions).

No existing sparse format handles this combination.
COO~\cite{coo} and CSR~\cite{csr} treat sparsity as unstructured and
incur high overhead per non-zero entry.
BCSR~\cite{bcsr} requires uniform block sizes, which DNN layer widths
do not satisfy.
All standard formats assume zero as the default for absent entries, but
abstract interpretation uses $\pm\infty$ as initial bounds.
\sparse (\emph{Generalized Block Compressed Sparse Representation}) addresses
all of these requirements.
It organizes the coefficient matrix into variable-sized blocks
aligned to DNN layers, stores each block in the most compact
representation, and carries
an explicit non-zero default value. 

\subsection{Correctness and Generality}
\label{sec:new-certs-ov}

\cf specifications have a neuron-level operational
semantics~\cite{constraintflow}, whereas \compiler executes them at the tensor
level. We prove that this change of execution level preserves the semantics:
the generated code produces, for every neuron, the same abstract value as the
neuron-level semantics. \compiler reaches the tensor level in three steps:
shape analysis lowers the specification to a tensor IR, domain-specific rewrite
rules optimize the IR, and the IR runs over the \sparse sparse tensor
representation. Code generation assigns the IR a tensor-level semantics, and we
prove that all three steps preserve meaning. The soundness that \cf establishes
at the neuron level therefore carries over to the compiled tensor-level operations.
\S\ref{sec:correctness} develops this in detail. 

This makes a certifier written as a neuron-level specification directly executable
on DNNs, with no tensor code written by hand. \S\ref{sec:eval} demonstrates
this on 8 certifiers. These include 4 with existing hand-optimized libraries, 2 that
had \cf specifications~\cite{constraintflow} with no executables, and 2 that are entirely new. Additionally, we specify a family of
DeepPoly~\cite{deeppoly} variants that trade precision for cost, each obtained by a
small edit to a specification that would otherwise demand a separate tensor-level
reimplementation. Across these benchmarks, \compiler is competitive with the
hand-optimized baselines while spanning a design space that no single library
covers.

\section{The \compiler Compiler}
\label{sec:compiler}

%
%
The compilation pipeline, shown in Fig.~\ref{fig:pipeline}, has three 
stages addressing the three challenges identified in \S~\ref{sec:overview}. First, a neuron-level \cf specification is compiled into an Intermediate
Representation (IR) using a novel \emph{shape analysis} pass that traverses the 
specification and inserts the tensor operations implicit in the specification 
(\S\ref{sec:shapeanalysis}). Second, domain-specific rewrite rules then optimize 
the IR, and the code generation emits the final code (\S\ref{sec:rewrites}). Third, the generated code uses \sparse, a sparse tensor backend optimized for the sparsity observed in DNN certifiers (\S\ref{sec:sparserepresentation}).

\subsection{Intermediate Representation and Metadata}
\label{sec:irandmetadata}

\paragraph{IR nodes.}
The IR has two layers.
\emph{Expression nodes} represent computed values; \emph{statement nodes} handle
control flow.
Fig.~\ref{fig:irgrammar-main} shows the key IR nodes;
the complete grammar is in Appendix~\ref{app:irgrammar}.
The base IR nodes include constants ($\irconst$), variables ($\irvar$), and fresh symbolic variables ($\irnoise$).
Beyond these, there are two kinds of expressions: (i)~tensor expressions, and (ii)~domain-specific expressions.
Tensor expressions represent standard tensor operations supported by tensor libraries:
$\irbinary$, $\irinner$, and similar nodes handle element-wise, ternary, and reduction computations,
while $\iradddim$, $\irrepeat$, etc, enable tensor shape manipulations.
Domain-specific expressions handle operations over polyhedral and symbolic expressions,
both of which consist of two components: coefficients and a constant.
\compiler defines IR nodes to manipulate these abstractly:
$\irextractpolyconst$, $\irextractpolycoeff$, and their \textsf{Sym} counterparts extract
components from existing expressions;
$\ircombinepoly$ and $\ircombinesym$ reassemble them into new expressions;
and $\irconstpoly$, $\irneuronpoly$, and similar nodes construct an expression from a
single component, treating the other as zero.
Finally, at every invocation of \map, the components of the input expression are
extracted using $\irmapcoeff$, $\irmapneuron$, and $\irmapnoise$ IR nodes.

\begin{figure}[t]
\centering\small
\begin{grammar}
<IR Expression> $i_e$ ::= $\irconst \mid \irvar \mid \irnoise$
  \alt $\irbinary(i_{e_1}, i_{e_2})
       \mid \irrepeat(i_{e_1})
       \mid \irreduce(i_{e_1})
       \mid \iradddim(i_{e_1})$
  \alt $\ircombinepoly(i_{e_1}, i_{e_2})
       \mid \irextractpolycoeff(i_{e_1})
       \mid \irextractpolyconst(i_{e_1})$
  \alt $\irneuronpoly(i_{e_1})
       \mid \irconstpoly(i_{e_1})$
  \alt $\irmapneuron(i_{e_1})
       \mid \irmapcoeff(i_{e_1})
       \mid \irmapnoise(i_{e_1})$

<IR Statement> $i_s$ ::= $\irassignment(\irvar,\, i_e)
       \mid \irseq(i_{s_1},\, i_{s_2})
       \mid \irwhile(i_e,\, i_s)
       \mid \irite(i_e,\, i_{s_1},\, i_{s_2})$
\end{grammar}
\caption{Subset of IR nodes. Each composite node takes sub-expressions as arguments.}
\label{fig:irgrammar-main}
\end{figure}

\paragraph{Metadata.}
The central design challenge in shape analysis is that tensors in DNN
certification routinely arise with shapes that need not be fully materialized.
For example, a coefficient matrix of effective shape $[m,n]$ might be produced
by broadcasting a $[1,n]$ vector across $m$ current-layer neurons. However, materializing
the full $[m,n]$ array is unnecessary and expensive. So, shape analysis needs a way
to distinguish ``physically stored'' dimensions from ``logically repeated''.

\begin{definition}[IR Metadata Element]
\label{def:irme}
An \emph{IrMetadataElement} $\irme$ is a tuple
$\langle \types,\, \isconst,\, \irshape,\, \irbroadcast \rangle$ where:
\begin{enumerate}[topsep=2pt,itemsep=0pt]
\item $\types$: the type of the node (e.g., \typeneuron, \polyexp, \float);
\item $\isconst \in \{\true, \false\}$: whether the node is a 
  constant;
\item $\irshape = [\constant_1, \ldots, \constant_k]$: the
  \emph{intrinsic shape}: the physically materialized dimensions of the tensor;
\item $\irbroadcast = [\constant_1, \ldots, \constant_k]$: the
  \emph{broadcast factor}: a vector capturing the multiplicity with which each dimension may be logically repeated (broadcast) during execution.
\end{enumerate}
\end{definition}

The \emph{effective shape} of $\irme$ is $\irme(\irshape) \otimes \irme(\irbroadcast)$
(elementwise product). An element is \emph{expanded} if $\irme(\irbroadcast) = \mathbf{1}$.
Since exact layer sizes are unknown at compile time, the constants
$\constant_i$ are drawn from a set of symbolic constants $\{\currs, \prevs, \polys, \syms, \batchs\}$,
which stand for the current layer size, previous layer size, polyhedral
expression size, zonotope expression size, and batch size, respectively.
The $\isconst$ flag handles a subtlety: constants such as $0$ and $1$
are polymorphic, taking the shape of whatever operand they are combined with.
Tracking $\isconst$ explicitly lets shape analysis skip premature materialization
of constants and enables constant propagation downstream.

\paragraph{Metadata stack}
While most IR nodes can work with a single metadata element, the \map construct
requires a \emph{stack}. The reason, explained in \S\ref{sec:shapeanalysis-map},
is that \map introduces a tensor dimension that must eventually be reduced
away, but the user-defined function $f$ applied inside \map may itself
introduce additional dimensions. Without careful
bookkeeping, the final reduction could target the wrong axis. \compiler
maintains a metadata stack
$\irm = \irme[k] :: \cdots :: \irme[1]$ (top at left): entering a \map pushes a
new element representing the newly introduced dimension; exiting pops it. The
top of the stack always identifies the axis that the latest
active \map introduced.


\subsection{Shape Alignment}
\label{sec:alignshapes}

Before describing shape analysis (\S\ref{sec:shapeanalysis}), we first define a central helper it relies on: \emph{shape alignment}.
Whenever two IR expressions must be combined (e.g., in a binary operation), their metadata stacks may not match: one may have more dimensions than the other, or the same number but with different effective sizes along some axis.
Shape alignment minimally resolves these mismatches without prematurely materializing dimensions.
The alignment operation, $\matchdims(i_{e_1}, i_{e_2}, m_1, m_2)$, takes two IR expressions and their metadata stacks and returns aligned versions $i_{e_1}'$, $i_{e_2}'$ sharing a common metadata stack $m$ whose effective shape covers both operands.

Rather than fully expand both operands to their effective shapes (setting all
broadcast factors to $\mathbf{1}$) before combining them, \matchdims\ expands each
dimension only as far as the current operation requires. 
%
%
Two metadata elements may
differ in \emph{rank} (number of dimensions), so \textsc{AlignShapes} first pads the shorter
element with new dimensions of intrinsic shape $1$ and broadcast factor equal to
the effective size of the extra dimensions in the longer one.
Once both elements share a rank, \textsc{AlignShapes} merges them axis by axis: where the intrinsic shapes agree ($s_1[i]=s_2[i]$) it keeps
$s[i]$ and $\beta[i]$; where they disagree, necessarily with one side a
singleton ($s_k[i]=1$), it expands the singleton
to the effective size $s[i]=s_k[i]\times\beta_k[i]$ and resets $\beta[i]=1$.
The process is lifted this to whole stacks, one metadata element at a time. Finally,
\textsc{AlignShapes} realizes the two cases at the IR level: a rank increase becomes an $\iradddim$, and a dimension moved from the broadcast
factor into the intrinsic shape becomes an $\irrepeat$, yielding two IR expressions that share the common stack
$m$. \textsc{AlignShapes} is formalized in  Appendix~\ref{app:irgrammar}.

\subsection{Shape Analysis}
\label{sec:shapeanalysis}

Shape analysis is the core translation pass. It traverses the \cf AST and, for
each construct, emits IR node(s) together with their metadata. The pass takes
as input a \cf expression $\expr$ and a context $(\Gamma, \sstore, \fstore)$
mapping \cf variables to their types, IR nodes, and function definitions,
respectively. It returns a triple $(i_s, i_e, \irm)$: a statement node (for
control-flow constructs), an expression node, and the expression's metadata
stack.
%
We write
$\Gamma,\sigma,\fstore \vdash e \leadsto (i_s, i_e, m)$
for the \emph{translation judgement}.
There are two base cases: a constant $c$ produces
$(\irskip$, $\irconst(c)$, $\langle\mathrm{type}(c)$,$\true$,$[1]$,$[\currs]\rangle)$
and a variable $x$ produces $(\irskip,\, i_e,\, m)$ where $\sigma(x) = (i_e, m)$.
The complete inference rules are in
Appendix~\ref{app:formalism}. Below we present key cases.

\subsubsection{Binary Operations}

For scalar-typed binary expressions $\expr_1 \oplus \expr_2$, shape analysis
recursively processes both operands, aligns their shapes, and emits the binary
IR node.
The resulting metadata comes directly from $\matchdims$.
The operator $\oplus$ ranges over arithmetic operators ($+$, $-$, $\times$, $/$) and
comparisons ($<$, $\leq$, $=$).
For binary expressions of type \polyexp or \symexp, shape analysis applies the operation component-wise to the coefficients and the constants of the two operands; the \textsc{BinPoly} and \textsc{BinSym} rules are given in Appendix~\ref{app:formalism}.

\subsubsection{Shape Analysis of \map}
\label{sec:shapeanalysis-map}

Recall from \S\ref{sec:overview} that $\expr.\map(f)$ iterates $f$ over the neuron-coefficient pairs of the polyhedral expression $\expr$ and accumulates the results.
%
For $\expr.\map(f)$ where
$f : \typeneuron \times \float \to \polyexp$, shape analysis must:
(1)~extract the neurons $N$ and coefficients $C$ from the polyhedral expression
$\expr$, introducing a new axis of size $\polys$ for the neuron-coefficient
pairs;
(2)~apply $f$ pointwise along that axis; and
(3)~reduce over that axis to sum the results.
Shape analysis realizes these steps by threading a metadata stack that keeps track
of the $\polys$ axis across the whole body of $f$, so that step~3 can reliably
locate the axis it must reduce even after $f$ has introduced dimensions of its own.
The \textsc{Map} rule is formalized in Fig.~\ref{fig:map-rule}.

\emph{Step~1: Extract and push.}
The map metadata element $m_e^{\mathrm{map}}$ records the new $\polys$ axis.
The updated store $\sigma^+$ binds $f$'s arguments: the coefficient variable $c_f$ receives $\irmapcoeff(i_{e_1})$ with $\polys$ as the intrinsic shape. Each coefficient is distinct, so all $\polys$ of them must be stored.
The neuron $n_f$ gets $\irmapneuron(i_{e_1})$ with $\polys$ in the \emph{broadcast factor} ($m_1^{bc}$). The same neurons are shared across all coefficient slices, so a single copy is reused.
The constant term $i_{e_1}.k$ is extracted separately because it does not participate in the per-element application of $f$.

\emph{Step~2: Apply $f$.}
$f$'s body $e_f$ is compiled under $\sigma^+$.
The rule requires the returned metadata stack to decompose as $m_e^{f}{::}m_{\mathit{out}}$ for some top element $m_e^{f}$.

\emph{Step~3: Reduce and pop ($\irreduce$ and $\matchdims$ premises).}
$\irreduce$ uses $m_e^{f}$, the top element returned by the body, to identify and remove the map axis from the compiled coefficient and constant outputs, popping it from the stack.
The original constant $i_{e_1}.k$ is then aligned with the reduced constant and added back; $\ircombinepoly$ assembles the final polyhedral expression.

\begin{figure}[t]
\centering\small
\begin{mathpar}
\inferrule*[Left=\textsc{Map}]{
  \Gamma,\sigma,F \vdash e' \leadsto (i_{s_1},\, i_{e_1},\, m_1)
  \quad F(f) = (n_f,\, c_f,\, e_f) \\
  m_e^{\mathrm{map}} = \langle\typeneuron,\;\false,\;[\polys],\;[1]\rangle \\
  \sigma^{+} = \sigma\bigl[n_f \mapsto (\irmapneuron(i_{e_1}),\;
                                         m_e^{\mathrm{map}}{::}m_1^{bc}),\;
                            c_f \mapsto (\irmapcoeff(i_{e_1}),\;
                                         m_e^{\mathrm{map}}{::}m_1)\bigr] \\
  \Gamma,\;\sigma^{+},\;F \;\vdash\; e_f
    \;\leadsto\; (i_{s_f},\; i_{e_f},\; m_e^{f}{::}m_{\mathit{out}}) \\
  i_{e_{pc}} = \irreduce(i_{e_f}.C,\; m_e^{f}) \quad
  i_{e_{pz}} = \irreduce(i_{e_f}.k,\; m_e^{f}) \\
  \matchdims(i_{e_1}.k,\;
             m_1[\types{\mapsto}\float],\;
             i_{e_{pz}},\;
             m_{\mathit{out}})
  = (i_{e_k}',\; i_{e_{pz}}',\; m')
}{
  \Gamma,\;\sigma,\;F \;\vdash\; e'.\map(f) \;\leadsto\;
  \Bigl(\irseq(i_{s_1},\;i_{s_f}),\;
    \ircombinepoly\bigl(
      i_{e_{pc}},\;
      \irbinary(i_{e_k}',\;i_{e_{pz}}',\;{+})\bigr),\;
    m'\Bigr)
}
\end{mathpar}
\caption{Shape analysis of \map on a polyhedral expression.
  Notation: $i_e.C \triangleq \irextractpolycoeff(i_e)$;
  $i_e.k \triangleq \irextractpolyconst(i_e)$;
  $m^{bc}$: metadata stack with the $\polys$ dimension moved to the broadcast factor.
  }
\label{fig:map-rule}
\end{figure}


\subsubsection{Shape Analysis of \traverse}
\label{sec:shapeanalysis-traverse}

The \traverse construct is a general fixed-point iteration over DNN structure.
Starting from a polyhedral expression over some set of neurons, it repeatedly applies a user-defined replacement function that rewrites each referenced neuron in terms of other neurons, until a stopping criterion is met.
Different certifiers instantiate this differently: some substitute neurons in terms of neurons from a preceding layer (as in backward-propagation certifiers); others substitute in terms of neighbors within the same layer.
At each iteration, the set of neurons referenced by the expression may change, and the shape of the expression changes accordingly.
\traverse compiles to a \texttt{while} loop.
Compiling each iteration is similar to compiling a \map body: we extract neurons
and coefficients from the loop-carried polyhedral expression, apply the
replacement function with an $\irreduce$ step, and re-assemble.
Unlike \map, the reduced result is not the final output but instead replaces the
loop-carried polyhedral expression for the next iteration.

A challenge is that the expression's shape evolves across iterations:
a neuron from layer $\ell$ may expand into several from layer $\ell{-}1$,
which each expand further at the next step. Unlike \map, where the new axis
size ($\polys$) is fixed at entry, the loop-carried shape changes for each iteration. \compiler handles this by introducing fresh symbolic shape variable $\polys_\ell$ in the
loop-carried metadata, 
whose value is populated at runtime for each iteration.

\subsection{Domain-Specific Rewrites and Code Generation}
\label{sec:rewrites}

\compiler applies standard compiler passes (common
subexpression elimination, loop-invariant code motion) followed by a set of
domain-specific rewrite rules. Each rule has the form $i_{e_1} \Rightarrow_P i_{e_2}$:
when the precondition $P$ holds, the LHS IR expression $i_{e_1}$ is
replaced by $i_{e_2}$. Fig.~\ref{fig:rewrite-rules} presents four example domain-specific rewrite rules that can be applied to the IR of many DNN certifiers.

\textit{(R1) Contraction-to-MatMul.}
Whenever the IR contains a repeat-multiply-reduce sequence over a shared axis, Rule~R1
collapses it into a single $\irinner$ call (lowered to \texttt{matmul}).
The precondition $P_m$ requires that $\irexpression[1]$ is repeated along its
\emph{last} axis, $\irexpression[2]$ along its \emph{first} axis, and the
reduction targets the shared middle axis. In \compiler, we use a more general version of this rewrite rule that extends to tensors with more than 3 dimensions and differing numbers of dimensions.
Applying it replaces four tensor operations with a single \texttt{matmul} call. 

\textit{(R2) Ternary-to-Multiply.}
Rule~R1 cannot fire directly when the specification contains a conditional as in line 1 of Fig.~\ref{fig:cf-code}. 
Using Rule~R2, the ternary $(\irexpression[b]\ ?\ \irexpression[2]\
:\ \irexpression[3])$ becomes $\irexpression[b] \times \irexpression[2] + (1
- \irexpression[b]) \times \irexpression[3]$, as shown in the RHS tree in
Fig.~\ref{fig:rewrite-rules}. The precondition $P_t$ requires that neither
$\irexpression[2]$ nor $\irexpression[3]$ contains a division, to avoid
division-by-zero when $\irexpression[b] = 0$. With the ternary eliminated,
$\irexpression[b]$, $\irexpression[2]$, $\irexpression[3]$ participate in
ordinary multiplications, creating a possibility for the Rule~R1.

\textit{(R3) Clamp-Lower; (R4) Clamp-Upper.}
Rules R3 and R4 recognize products of a Boolean mask with a linear expression
and replace them with a clamping operation, executable as a single fused kernel.
R3 (\emph{Clamp-Lower}) fires when a $(\irexpression[1] \geq \irexpression[2])$ mask is multiplied
by $\irexpression[1]$ (same expression) and $\irexpression[2] = 0$; the result
is $\irclamp(\irexpression[1], \mathit{min}{=}0)$, which lower-bounds the value at zero.
R4 (\emph{Clamp-Upper}) is symmetric with $\leq$ and $\mathit{max}$, upper-bounding at zero.
These patterns arise in ReLU transformer computations
and avoid materializing the Boolean mask as a separate tensor.


\begin{figure}[t]
\centering
\tikzset{
  irn/.style={draw, rounded corners=3pt, fill=blue!12,
              font=\scriptsize\sffamily, inner sep=3pt,
              minimum height=1.6em},
  lf/.style={draw, circle, fill=orange!20,
             font=\small\itshape, inner sep=2pt,
             minimum size=1.6em},
  every child/.style={edge from parent/.style={draw,-}},
}
{%
\setlength{\tabcolsep}{5pt}%
\renewcommand{\arraystretch}{1.0}%
\newlength{\qw}%
\setlength{\qw}{\dimexpr0.5\linewidth-2\tabcolsep-1.5\arrayrulewidth\relax}%
\begin{tabular}{|>{\centering\arraybackslash}p{\qw}|>{\centering\arraybackslash}p{\qw}|}
\hline
{\small\textsf{(R1)} Contraction-to-MatMul}\par\smallskip
\scalebox{0.77}{%
\begin{tabular}{@{}c@{\;$\Rightarrow_{P_m}$\;}c@{}}
\begin{tikzpicture}[baseline=(current bounding box.center), grow=down,
    level distance=8mm, level 2/.style={sibling distance=15mm}]
  \node[irn] {IrReduce}
    child { node[irn] {IrMult}
      child { node[irn] {IrRepeat}
        child { node[lf] {$\irexpression[1]$} } }
      child { node[irn] {IrRepeat}
        child { node[lf] {$\irexpression[2]$} } }
    };
\end{tikzpicture}
&
\begin{tikzpicture}[baseline=(current bounding box.center), grow=down,
    level distance=8mm, level 1/.style={sibling distance=12mm}]
  \node[irn] {IrInnerProduct}
    child { node[lf] {$\irexpression[1]$} }
    child { node[lf] {$\irexpression[2]$} };
\end{tikzpicture}
\end{tabular}}\par\smallskip
{\footnotesize\itshape$P$: $e_1$ repeated along last axis;
  $e_2$ along first; reduce over shared middle.}
  
&
{\small\textsf{(R3)} Clamp-Lower ($\geq 0$)}\par\smallskip
\scalebox{0.77}{%
\begin{tabular}{@{}c@{\;$\Rightarrow_{P_c}$\;}c@{}}
\begin{tikzpicture}[baseline=(current bounding box.center), grow=down,
    level distance=8mm,
    level 1/.style={sibling distance=18mm},
    level 2/.style={sibling distance=9mm}]
  \node[irn] {IrBinary($\times$)}
    child { node[irn] {IrBinary($\geq$)}
      child { node[lf] {$\irexpression[1]$} }
      child { node[lf] {$0$} }
    }
    child { node[lf] {$\irexpression[1]$} };
\end{tikzpicture}
&
\begin{tikzpicture}[baseline=(current bounding box.center), grow=down,
    level distance=8mm]
  \node[irn] {IrClamp($\mathit{min}{=}0$)}
    child { node[lf] {$\irexpression[1]$} };
\end{tikzpicture}
\end{tabular}}\par\smallskip
{\footnotesize\itshape$P$: left factor is a Boolean mask;
  right equals masked expr; threshold is~0.}
\\
\hline
{\small\textsf{(R2)} Ternary-to-Multiply}\par\smallskip
\scalebox{0.77}{%
\begin{tabular}{@{}c@{\;$\Rightarrow_{P_t}$\;}c@{}}
\begin{tikzpicture}[baseline=(current bounding box.center), grow=down,
    level distance=8mm, level 1/.style={sibling distance=9mm}]
  \node[irn] {IrTernary}
    child { node[lf] {$\irexpression[b]$} }
    child { node[lf] {$\irexpression[2]$} }
    child { node[lf] {$\irexpression[3]$} };
\end{tikzpicture}
&
\begin{tikzpicture}[baseline=(current bounding box.center), grow=down,
    level distance=8mm,
    level 1/.style={sibling distance=32mm},
    level 2/.style={sibling distance=14mm},
    level 3/.style={sibling distance=10mm}]
  \node[irn] {IrBinary($+$)}
    child { node[irn] {IrBinary($\times$)}
      child { node[lf] {$\irexpression[b]$} }
      child { node[lf] {$\irexpression[2]$} }
    }
    child { node[irn] {IrBinary($\times$)}
      child { node[irn] {IrBinary($-$)}
        child { node[irn] {$\irconst(1)$} }
        child { node[lf] {$\irexpression[b]$} }
      }
      child { node[lf] {$\irexpression[3]$} }
    };
\end{tikzpicture}
\end{tabular}}\par\smallskip
{\footnotesize\itshape$P$: $e_2$ and $e_3$ contain no divisions.}
&
{\small\textsf{(R4)} Clamp-Upper ($\leq 0$)}\par\smallskip
\scalebox{0.77}{%
\begin{tabular}{@{}c@{\;$\Rightarrow_{P_c}$\;}c@{}}
\begin{tikzpicture}[baseline=(current bounding box.center), grow=down,
    level distance=8mm,
    level 1/.style={sibling distance=18mm},
    level 2/.style={sibling distance=9mm}]
  \node[irn] {IrBinary($\times$)}
    child { node[irn] {IrBinary($\leq$)}
      child { node[lf] {$\irexpression[1]$} }
      child { node[lf] {$0$} }
    }
    child { node[lf] {$\irexpression[1]$} };
\end{tikzpicture}
&
\begin{tikzpicture}[baseline=(current bounding box.center), grow=down,
    level distance=8mm]
  \node[irn] {IrClamp($\mathit{max}{=}0$)}
    child { node[lf] {$\irexpression[1]$} };
\end{tikzpicture}
\end{tabular}}\par\smallskip
{\footnotesize\itshape$P$: left factor is a Boolean mask;
  right equals masked expr; threshold is~0.}
\\
\hline
\end{tabular}%
}

\caption{Domain-specific rewrite rules. Preconditions appear beneath each rule.}
\vspace{-8mm}
\label{fig:rewrite-rules}
\end{figure}

\paragraph{Code generation.}
\label{sec:codegeneration}
The final IR is lowered to Python code. Statement nodes map directly to Python control flow:
$\irwhile$ $\to$ \texttt{while}, $\irite$ $\to$ \texttt{if/else},
$\irassignment$ $\to$ assignment, $\irseq$ $\to$ sequential composition.
The base expressions in the IR, such as $\irconst$ and $\irvar$, are lowered to Python constants and variables. Polyhedral and symbolic expressions are implemented as specialized Python classes, which internally store the constant and coefficient components using sparse tensors, as described in \S~\ref{sec:sparserepresentation}. These classes include methods to perform domain-specific operations such as component extraction, affine combination, etc.
For IR nodes involving tensor-level operations, the metadata computed during shape analysis records the expected tensor shapes, thereby enabling a direct mapping from IR expression nodes (e.g., binary operations) to appropriate tensor API. 

While the generated code is agnostic to the choice of tensor backend (Fig.~\ref{fig:gencode}) and can interface with standard libraries such as PyTorch or TensorFlow, we design a custom sparse tensor backend optimized for the sparsity patterns commonly seen in DNN certification. Preserving modularity, each tensor operation is compiled into a high-level API call provided by our backend. For example, $\irbinary$ is translated to an API call \texttt{binary(a,b,op)}. At runtime, if \texttt{a} and \texttt{b} are dense and \texttt{op} is \texttt{`+'}, the backend invokes the native PyTorch operation \texttt{a+b}. Or, if the operands are sparse and managed by our custom backend, an optimized implementation of the binary operation is used.
\subsection{Semantic Preservation}
\label{sec:correctness}

We began with a neuron-level \cf specification. Shape analysis
(\S\ref{sec:shapeanalysis}) turned it into an IR that computes over tensors, and the
rewrites of \S\ref{sec:rewrites} optimized that IR. To say what this IR
\emph{computes}, we equip it with an \emph{execution semantics} at the level of
tensors: a \emph{tensor environment} $\varrho$, the runtime symbol table mapping
IR variables to concrete tensors, with a judgement $\varrho \vdash i_e \Downarrow
T$ that evaluates an expression to a tensor and a judgement
$\varrho \vdash i_s \Downarrow \varrho'$ that executes a statement by updating
$\varrho$, exactly as an assignment updates a symbol table. The rules are standard: each IR node denotes the like-named tensor operation on its arguments (details in Appendix~\ref{app:proof-semanticpreservation}).

We can now state the correctness of the compilation. Two
semantics describe the same specification $P$: the neuron-level semantics of
prior work~\cite{constraintflow}, run independently on each of the $N$ neurons of a
layer, and \compiler's tensor execution semantics, run once on the whole layer.
Correctness is a \emph{bisimulation} between them, the two sides related by the
\emph{effective tensor} $\mathsf{eff}(T, m)$, the full tensor obtained by
logically expanding the broadcast dimensions recorded in $m$
(Appendix~\ref{app:proof-semanticpreservation}).

\begin{theorem}
\label{thm:correctness}
Let a well-typed \cf specification $P$ compile to $(i_s, i_e, m)$, and let
$\varrho_0$ be consistent with an abstract state $\mathbf{s}$ over a layer of $N$
neurons. If $\varrho_0 \vdash i_s \Downarrow \varrho$ and
$\varrho \vdash i_e \Downarrow T$, then for every neuron $i \in [N]$,
$\mathsf{eff}\bigl(T,\ m\bigr)[i] \;=\; v_i,$
where $v_i$ is the
result of the neuron-level \cf semantics of $P$ at neuron $i$.
That is, running the generated program once reproduces, for all $N$ neurons at once,
what prior work computes by running the specification $N$ times independently.
\end{theorem}

The proof (Appendix~\ref{app:proof-semanticpreservation}) is by induction on the \cf expression: it maintains the invariant that slicing each compiled sub-expression at neuron $i$ yields its neuron-level value $v_i$, using lemmas for shape alignment, the metadata stack, and \map-axis materialization.

\begin{theorem}
\label{thm:rewrites}
For each rule in Fig.~\ref{fig:rewrite-rules} and any tensor environment $\varrho$:
if the LHS IR node evaluates to a tensor $T$ and the rule's precondition holds,
then the RHS node also evaluates to $T$.
\end{theorem}



\begin{theorem}[Code Generation]
\label{thm:codegen}
Let $i_s$ be an IR program and $\mathcal{C}(i_s)$ the Python code generated for it, and assume that every backend call implements the tensor operation of the IR node it is generated for. If $\varrho \vdash i_s \Downarrow \varrho'$, then executing $\mathcal{C}(i_s)$ from a state that stores $\varrho$ terminates in a state that stores $\varrho'$.
\end{theorem}

The proofs are in Appendix~\ref{app:proof-rewriterules},~\ref{app:proof-codegen}.

\section{Sparse Tensor Representation}
\label{sec:sparserepresentation}

\S\ref{sec:sparsity-ov} showed that the coefficient tensors arising in DNN
certification are highly sparse, with sparsity patterns that vary by certifier
and DNN architecture. This section describes \sparse (Generalized Block
Compressed Sparse Representation), a novel sparse tensor representation that consists of blocks of varying sizes and internal sparsity.
Standard sparse formats such as CSR and BSR are ill-suited for DNN certifiers for
two reasons. First, they assume a background (default) value of zero, but
certification tensors often arise with a non-zero background. For instance, the bounds of all neurons are initialized as $+\infty$ or $-\infty$. Second, they assume either
uniform block sizes (BSR) or no block structure at all (CSR/COO), whereas
DNN certification produces tensors with variable-size blocks arising from layer sizes whose internal
structure reflects the DNN architecture (convolutional layer,
residual connections, activations). \sparse addresses both limitations: it supports a
configurable background value, and variable-sized blocks with diverse internal sparsity.

\subsection{Generalized Block Compressed Sparse Representation (\sparse)}
\label{sec:sparsedefs}

\begin{definition}[\sparse Tensor]
\label{def:sparsetensor}
A tensor in \sparse format is a tuple $\mathcal{T} = (d,\, S,\, B,\, Z)$, where:
\begin{itemize}
\item $d \in \mathbb{R}$ is the \emph{background value}, used for all entries
  not covered by any block;
\item $B = \{B_k\}_{k=1}^{m}$ is a list of non-overlapping \emph{sparse blocks}
  with $S = \{s_k\}_{k=1}^{m}$ as the starting indices;
\item $Z \in \mathbb{N}^r$ is the shape of the full $r$-dimensional tensor.
\end{itemize}
\end{definition}

A sparse block $B_k$ is a contiguous sub-tensor of $\mathcal{T}$
starting at index $s_k$. Entries outside all blocks take the background
value $d$. A sparse block is itself defined abstractly to support multiple
types:

\begin{definition}[Sparse Block]
\label{def:sparseblock}
A \emph{sparse block} is a tuple $B = (b,\, \Theta)$, where
$b \in \mathbb{R}^{d_1 \times \cdots \times d_r}$ is a \emph{core tensor}
and $\Theta$ is a set of structural parameters. Each block type defines an
injective index mapping $\varphi_\Theta$ from core indices to dense block
indices: $B_D[\varphi_\Theta(\mathbf{i})] = b[\mathbf{i}]$, where $\mathbf{i}$ is an index of $b$.
\end{definition}



Fig.~\ref{fig:sparseblocksmain} illustrates the three block types central to
our development: Stencil, Window, and Diagonal. Entries whose structure
matches none of the specialized types fall back to a \emph{Dense block}
(\dblock), whose core tensor equals the block itself and whose mapping
$\varphi$ is the identity. Two further specialized types, the Broadcast and
Uniform blocks are described in
Appendix~\ref{app:sparseblocks}.

\begin{figure*}[t]
\centering
\begin{subfigure}[b]{0.38\textwidth}
  \centering
  \mbox{\resizebox{!}{1.1cm}{
\begin{tikzpicture}[x=0.5cm,y=0.5cm]
  \def\W{20}    
  \def\H{6}     

  \foreach \r in {0,...,\numexpr\H-1\relax} {
    \foreach \c in {0,...,\numexpr\W-1\relax} {
      \pgfmathtruncatemacro\runA{and(\c>=\r,\c<\r+3)}
      \pgfmathtruncatemacro\runB{and(\c>=\r+6,\c<\r+9)}
      \pgfmathtruncatemacro\runC{and(\c>=\r+12,\c<\r+15)}

      \ifnum\runA=1
        \pgfmathtruncatemacro\lbl{(\c - \r) + 1}
      \else\ifnum\runB=1
        \pgfmathtruncatemacro\lbl{(\c - \r) - 2}
      \else\ifnum\runC=1
        \pgfmathtruncatemacro\lbl{(\c - \r) - 5}
      \else
        \def\lbl{0}
      \fi\fi\fi

      \ifnum\lbl>0
        \pgfmathsetmacro\shade{20 + (\lbl - 1)*10}  
        \fill[teal!\shade] ({\c},{-\r}) rectangle ++(1,-1);
      \else
        \fill[white]       ({\c},{-\r}) rectangle ++(1,-1);
      \fi

      \draw[gray] ({\c},{-\r}) rectangle ++(1,-1);
    }
  }
  \draw[line width=1pt] (0,0) rectangle (\W,-\H);
\end{tikzpicture}}\hspace{0.4em}%
  \resizebox{!}{0.55cm}{
\begin{tikzpicture}[x=0.5cm,y=0.5cm]
  \def\BX{0}   
  \def\BY{0}   

  \foreach \bi in {0,1,2} {
    \foreach \bj in {0,1,2} {
      \pgfmathtruncatemacro\lblb{\bi*3 + \bj + 1}
      \pgfmathsetmacro\shade{20 + (\lblb - 1)*10}
      \fill[teal!\shade]
        ({\BX+\bj},{\BY-\bi})
        rectangle ++(1,-1);
      \draw[gray]
        ({\BX+\bj},{\BY-\bi})
        rectangle ++(1,-1);
    }
  }
  \draw[line width=1pt]
    ({\BX},{\BY})
    rectangle
    ({\BX+3},{\BY-3});
\end{tikzpicture}}}
  \caption{Stencil Block}
  \label{fig:convblock}
\end{subfigure}
\hfill
\begin{subfigure}[b]{0.38\textwidth}
  \centering
  \mbox{\resizebox{!}{1.1cm}{
\begin{tikzpicture}[x=0.5cm,y=0.5cm]
  \def\W{20}    
  \def\H{6}     

  \def\rowColors{{%
  "blue!0!teal",
  "blue!14!teal",
  "blue!28!teal",
  "blue!42!teal",
  "blue!56!teal",
  "blue!70!teal"
  }}

  \foreach \r in {0,...,\numexpr\H-1\relax} {
    \pgfmathparse{\rowColors[\r]}
    \edef\rowColor{\pgfmathresult}

    \foreach \c in {0,...,\numexpr\W-1\relax} {
      \pgfmathtruncatemacro\runA{and(\c>=\r,\c<\r+3)}
      \pgfmathtruncatemacro\runB{and(\c>=\r+6,\c<\r+9)}
      \pgfmathtruncatemacro\runC{and(\c>=\r+12,\c<\r+15)}

      \ifnum\runA=1
        \pgfmathtruncatemacro\lbl{(\c - \r) + 1}
      \else\ifnum\runB=1
        \pgfmathtruncatemacro\lbl{(\c - \r) - 2}
      \else\ifnum\runC=1
        \pgfmathtruncatemacro\lbl{(\c - \r) - 5}
      \else
        \def\lbl{0}
      \fi\fi\fi

      \ifnum\lbl>0
        \pgfmathsetmacro\shade{20 + (\lbl - 1)*10}
        \fill[\rowColor!\shade] ({\c},{-\r}) rectangle ++(1,-1);
      \else
        \fill[white] ({\c},{-\r}) rectangle ++(1,-1);
      \fi

      \draw[gray] ({\c},{-\r}) rectangle ++(1,-1);
    }
  }
  \draw[line width=1pt] (0,0) rectangle (\W,-\H);
\end{tikzpicture}}\hspace{0.4em}%
  \resizebox{!}{1.1cm}{
\begin{tikzpicture}[x=0.5cm,y=0.5cm]
  \def\H{6}     
  \def\BX{0}    
  \def\BY{0}    

  \def\rowColors{{%
  "blue!0!teal",
  "blue!14!teal",
  "blue!28!teal",
  "blue!42!teal",
  "blue!56!teal",
  "blue!70!teal"
  }}

  \foreach \r in {0,...,\numexpr\H-1\relax} {
    \pgfmathparse{\rowColors[\r]}
    \edef\rowColor{\pgfmathresult}

    \foreach \c in {0,1,2} {
      \pgfmathtruncatemacro\lbl{(\c)+1} 
      \pgfmathsetmacro\shade{20 + (\lbl - 1)*10}

      \fill[\rowColor!\shade] ({\BX + \c}, {\BY - \r}) rectangle ++(1,-1);
      \draw[gray] ({\BX + \c}, {\BY - \r}) rectangle ++(1,-1);
    }

    \foreach \c in {0,1,2} {
      \pgfmathtruncatemacro\lbl{3 + (\c)+1} 
      \pgfmathsetmacro\shade{20 + (\lbl - 1)*10}

      \fill[\rowColor!\shade] ({\BX + 3 + \c}, {\BY - \r}) rectangle ++(1,-1);
      \draw[gray] ({\BX + 3 + \c}, {\BY - \r}) rectangle ++(1,-1);
    }

    \foreach \c in {0,1,2} {
      \pgfmathtruncatemacro\lbl{6 + (\c)+1} 
      \pgfmathsetmacro\shade{20 + (\lbl - 1)*10}

      \fill[\rowColor!\shade] ({\BX + 6 + \c}, {\BY - \r}) rectangle ++(1,-1);
      \draw[gray] ({\BX + 6 + \c}, {\BY - \r}) rectangle ++(1,-1);
    }
  }

  \foreach \block in {0,1,2} {
    \draw[line width=1pt]
      ({\BX + \block*3},{\BY})
      rectangle
      ({\BX + \block*3 + 3},{\BY - \H});
  }

\end{tikzpicture}}}
  \caption{Window Block}
  \label{fig:patchesblock}
\end{subfigure}
\hfill
\begin{subfigure}[b]{0.20\textwidth}
  \centering
  \mbox{\resizebox{!}{1.1cm}{
\begin{tikzpicture}[x=0.5cm, y=0.5cm, every node/.style={font=\small}]
  \def\size{6}

  \def\rowColors{{"teal!20!white", "teal!36!white", "teal!52!white", "teal!68!white", "teal!84!white", "teal!100!white"}}

  \foreach \i in {0,...,5} {
    \pgfmathparse{\rowColors[\i]}
    \edef\col{\pgfmathresult}
    \fill[\col] (\i,-\i) rectangle ++(1,-1);
  }

  \foreach \i in {0,...,\size} {
    \draw[gray] (0,-\i) -- (\size,-\i);     
    \draw[gray] (\i,0) -- (\i,-\size);      
  }
  \draw[line width=1pt, black] (0,0) rectangle (\size,-\size);
\end{tikzpicture}}\hspace{0.3em}%
  \resizebox{!}{1.1cm}{
\begin{tikzpicture}[x=0.5cm, y=0.5cm, every node/.style={font=\small}]
  \def\size{6}
  \def\BX{0}  

  \def\rowColors{{"teal!20!white", "teal!36!white", "teal!52!white", "teal!68!white", "teal!84!white", "teal!100!white"}}

  \foreach \i in {0,...,5} {
    \pgfmathparse{\rowColors[\i]}
    \edef\col{\pgfmathresult}
    \fill[\col] (\BX,-\i) rectangle ++(1,-1);
  }

  \foreach \i in {0,...,6} {
    \draw[gray] (\BX,-\i) -- (\BX+1,-\i);  
  }
  \draw[black] (\BX,0) -- (\BX,-6);         
  \draw[black] (\BX+1,0) -- (\BX+1,-6);     
  \draw[line width=1pt, black] (\BX,0) rectangle (\BX+1,-6);

\end{tikzpicture}}}
  \caption{Diagonal Block}
  \label{fig:diagblock}
\end{subfigure}
\caption{Dense and sparse representations of Stencil, Window, and Diagonal
  blocks. Different colors represent different values. In each pair, the left
  shows the dense block and the right shows its \sparse form.
  }
\label{fig:sparseblocksmain}
\end{figure*}

\noindent
\textbf{\kblock.}
In a convolutional layer, each output neuron is an affine combination of a
fixed-size patch of the input, with weights shared across all spatial
positions. Consequently, the polyhedral coefficient block for a convolutional
layer has a \emph{shifted-repeat} structure: row $i$ of the block is a sparse
vector of length equal to the input size, with $\lvert k \rvert$ non-zero
entries at positions shifted by $i \cdot \mathit{stride}$ relative to row 0
(where $k$ is the convolutional kernel). The core tensor $b \in \mathbb{R}^{\lvert k \rvert}$
stores the kernel values once, and $\Theta = \{\mathit{stride}, \mathit{padding}\}$
encodes the shift pattern. \\
\textbf{\pblock.}
For certifiers that maintain per-neuron polyhedral expressions that reference
patches of a previous layer (e.g., after backsubstituting through a
convolutional layer), the non-zero entries in each row occupy a contiguous
patch of the previous feature map, but the patches for different
neurons do not overlap and may have different sizes. The core tensor $b$
stores all patch values concatenated, and $\Theta = \{\mathit{stride},
\mathit{padding}\}$ encodes where each patch begins and ends. \\
\textbf{\diagblock.}
This captures the case where all the non-zero entries lie on a single diagonal. This arises when an abstract expression for a neuron only references one other neuron, e.g., when computing the bounds for a ReLU layer. The core tensor $b$ stores only the
diagonal entries, and $\Theta$ records the \emph{diagonal index}
$\delta$, the dimension along which the diagonal is defined.

\subsection{Operations over \sparse Tensors}
\label{sec:sparseops}

\compiler generates programs that call a tensor backend API; the \sparse
library implements this API for sparse tensors. We describe the element-wise
binary operations in detail.

\paragraph{Element-wise binary operations.}
Let $\mathcal{T}_1 = (d_1, S_1, B_1, Z)$ and $\mathcal{T}_2 = (d_2, S_2, B_2, Z)$
be two \sparse tensors, and $\mathcal{T}_3 = \mathcal{T}_1
\oplus \mathcal{T}_2$ for a binary operator $\oplus$. The background of the
result is $d_3 = d_1 \oplus d_2$, since every entry that is
background in both operands takes this value. To determine the
blocks of $\mathcal{T}_3$, a naive but sound approach would form a single large block covering
every index that either operand affects. However, it discards the inherent sparsity of the inputs and wastes computation.

We instead preserve sparsity by reasoning about which blocks interact. Two
blocks \emph{overlap} if they hold non-background values at a common index. An
entry of $\mathcal{T}_3$ is non-background exactly when at least one operand is
non-background there, so the blocks of $\mathcal{T}_3$ are computed by groups of
mutually overlapping blocks. Because overlap can chain, i.e., a block $B_i$ of
$\mathcal{T}_1$ may overlap $B_j$ of $\mathcal{T}_2$, which in turn overlaps
$B_k$ of $\mathcal{T}_1$, we compute the \emph{transitive closure} of the
overlap relation. Each resulting group gathers all blocks, from either operand,
that jointly compute a region of $\mathcal{T}_3$.

Each group must then be realized as one or more blocks of $\mathcal{T}_3$. We
can either merge the group into a single contiguous block or split it into
smaller disjoint blocks. Merging is simple but yields large, near-dense blocks
that give up sparsity, whereas splitting keeps blocks small at the cost of extra
bookkeeping. Finding the optimal choice is an open problem; in practice we use a
heuristic that works well: if all blocks in a group differ along at most one
dimension, we split, and otherwise we merge. These index computations determine the placement of sparse blocks before the corresponding tensor operations are executed. The numerical computation itself is performed on the blocks, ensuring that the internal sparsity is preserved when possible. 

After the index computation, this block-by-block computation can sometimes be avoided entirely. When a block
$B_i$ lies in only one operand, the operation reduces to combining $B_i$ with
the other operand's background value $d$, and the algebraic role of $d$ under
$\oplus$ fixes the result without touching the data. We formalize this,
we define identity and annihilator of a binary operator $\oplus$:

\begin{itemize}
\item The \emph{identity element} $\mathbf{1}_\oplus$ is the unique value such
  that $x \oplus \mathbf{1}_\oplus = \mathbf{1}_\oplus \oplus x = x$ for all $x$.
\item The \emph{annihilator element} $\mathbf{0}_\oplus$ is the unique value
  such that $x \oplus \mathbf{0}_\oplus = \mathbf{0}_\oplus \oplus x =
  \mathbf{0}_\oplus$ for all $x$.
\end{itemize}

When $B_i$ meets the background $d$ of the other operand, three cases arise:
\begin{itemize}
    \item If the constant is the identity, the result is simply the existing block (or a unary rewrite). 
    \item If it is the annihilator and matches the resulting dense constant, the block can be skipped.
    \item Otherwise, we must lift the constant to a block and perform the operation explicitly.
\end{itemize}

These optimizations avoid unnecessary computation and preserve sparsity, while ensuring the correctness of the resulting tensor.  The algorithm for matrix multiplication between $\mathcal{T}_1$ and $\mathcal{T}_2$ is similar. 
An example and the detailed discussion can be found in Appendix~\ref{app:sparse-example}, ~\ref{app:sparse-matmul}.
We prove that these operations compute the same result as their dense
counterparts (proof in Appendix~\ref{app:proof-sparseandrewrite}).

\begin{theorem}
\label{thm:sparse}
Let $\mathrm{dense}(\mathcal{T})$ be the dense tensor represented by $\mathcal{T}$.
For every \sparse operation $\odot$ and \sparse tensors $\mathcal{T}_1, \mathcal{T}_2$ of
appropriate shapes,
  $\mathrm{dense}(\mathcal{T}_1 \odot \mathcal{T}_2)
  = \mathrm{dense}(\mathcal{T}_1) \;\odot\; \mathrm{dense}(\mathcal{T}_2).$
\end{theorem}


\section{Evaluation}
\label{sec:eval}
We now evaluate \compiler's practicality and utility in developing DNN certifiers. In particular, we investigate whether neuron-level specifications can be systematically compiled into efficient tensor programs, enabling researchers to implement, experiment with, and extend certifier designs without manually constructing tensor-level programs.
To this end, we address the following:

\begin{description}
\item[RQ1 (Generality).] Can \compiler compile a diverse set of
  certifiers? (\S\ref{sec:eval-general})
\item[RQ2 (Exploration).] How does \compiler aid the exploration of precision-runtime tradeoffs? (\S\ref{sec:eval-explore})
\item[RQ3 (Performance).] How do the generated executables compare to
hand-written certifiers? (\S\ref{sec:eval-perf})
\item[RQ4 (Ablation).] How much do rewrite rules and \sparse contribute to performance? (\S\ref{sec:ablation})
\end{description}

\paragraph{Certifiers.}
We compile eight certifiers, chosen to span the space of abstract domains and to
include designs no existing tool can run. They fall into three groups.
\emph{Existing} certifiers have widely-used, hand-optimized implementations:
\textbf{IBP}~\cite{interval} propagates concrete interval bounds forward layer by layer; \textbf{DeepZ}~\cite{deepz} uses a zonotope domain that tracks
linear correlations between neurons; \textbf{DeepPoly}~\cite{deeppoly} uses a
polyhedral domain with \emph{backsubstitution};
and \textbf{CROWN-IBP}~\cite{crownibp} combines a
linear relaxation with IBP. \emph{Prototype} certifiers: \textbf{ReuseCert} and
\textbf{PolyZono}, were specified in prior work~\cite{constraintflow} but never
executed on real networks, for lack of a compiler. Finally, \textbf{SkipPoly} and
\textbf{ZID} are \emph{new} certifiers we introduce; we describe all four
non-existing certifiers in \S\ref{sec:eval-general}. The complete \cf
specifications of all eight certifiers are given in
Appendix~\ref{app:specifications}.

\paragraph{Setup.}
We evaluate on 32 DNNs from the ERAN benchmark suite~\cite{eran}, the standard
suite used in prior certification work and the VNN-COMP competition. The DNNs
cover two datasets (MNIST, CIFAR-10), and three training methods (standard, PGD
adversarial training, DiffAI certified training). 
The complete details of the DNNs are provided in Appendix~\ref{app:setup} (Table~\ref{tab:allnets}).
For clarity, in the main text for RQ1 and RQ2, we focus on the evaluation  on 6 representative DNNs (Table~\ref{tab:repnets}),
chosen to cover the benchmark's diversity systematically: one for each combination
of dataset (MNIST, CIFAR-10) and robustness-training regime (standard, PGD,
DiffAI), together spanning fully-connected and convolutional architectures. 
We compare \compiler's
generated executables against three hand-written libraries including auto\_LiRPA~\cite{lirpagithub}
IVAN~\cite{ivangithub}, 
and ERAN~\cite{eran}.
All experiments run on a dual-socket Intel Xeon Gold
6348 machine with $4$ cores, $1$\,TB of RAM,
running Ubuntu 24.04. 
We
impose a $200$\,s per-DNN timeout for all experiments.


\begin{table}[]
    \centering \footnotesize
    \caption{Six representative DNNs for evaluation.}
    \label{tab:repnets}
    \begin{tabular}{@{}lrrrrrrr@{}}
    \toprule
        DNN & Dataset &     Model &       Training Method &  Architecture & Layers & Perturbation ($\epsilon$) & Batch Size \\
    \midrule
     $N{10}$ &   MNIST &     6x500 &    PGD &           FCN &      6  & 0.005 & 100\\
    $N{11}$ &   MNIST &    4x1024 &              Standard &           FCN &      3 & 0.004 & 100\\
    $N{14}$ &   MNIST & ConvSmall &                DiffAI & Convolutional &      3  & 0.1 & 100\\
    $N{22}$ & CIFAR10 &     6x500 &              Standard &           FCN &      6  & 4e-5 & 100\\
    $N{30}$ & CIFAR10 &   ConvMed & PGD & Convolutional &      3 & 5e-4 & 10\\
    $N{32}$ & CIFAR10 &   ConvBig &                DiffAI & Convolutional &      6 & 0.002 & 10\\
    \bottomrule
    \end{tabular}
\end{table}


\begin{table}[t]
\centering\footnotesize
\caption{Certifiers supported by \compiler vs. existing libraries, and their \cf specification sizes.}
\label{tab:coverage}
\setlength{\tabcolsep}{4pt}
\begin{tabular}{@{}l cccc cc cc@{}}
\toprule
 & \multicolumn{4}{c}{Existing Implementations} & \multicolumn{2}{c}{Prototype Specifications} & \multicolumn{2}{c}{New Designs} \\
\cmidrule(lr){2-5}\cmidrule(lr){6-7}\cmidrule(l){8-9}
 & IBP & CROWN-IBP & DeepZ & DeepPoly & ReuseCert & PolyZono & SkipPoly & ZID \\
\midrule
\cf LOCs    & 11 & 27 & 14 & 27 & 25 & 35 & 22 & 67 \\
\midrule
auto\_LiRPA & \cmark & \cmark & \xmark & \cmark & \xmark & \xmark & \xmark & \xmark \\
IVAN        & \cmark & \xmark & \cmark & \cmark & \xmark & \xmark & \xmark & \xmark \\
ERAN        & \xmark & \xmark & \cmark & \cmark & \xmark & \xmark & \xmark & \xmark \\
\compiler   & \cmark & \cmark & \cmark & \cmark & \cmark & \cmark & \cmark & \cmark \\
\bottomrule
\end{tabular}
\end{table}

\subsection{RQ1: Supporting Diverse Certifiers}
\label{sec:eval-general}

\compiler compiles and executes all eight certifiers.
Table~\ref{tab:coverage} compares this coverage with existing libraries. Each
library supports only a fixed set of certifiers because each certifier requires a
separate hand-written tensor implementation. In contrast, \compiler accepts certifier specifications: each of the eight certifiers is only 11--67
lines of \cf (Table~\ref{tab:coverage}), with no user-written tensor operations.
\compiler synthesizes the corresponding tensor programs automatically. 

Four of
our certifiers are not supported by any
existing library: 
%
\textbf{SkipPoly} modifies DeepPoly's backsubstitution to reference neurons
two layers back instead of one. Because ReLU activations prevent cancellations
across layers, this preserves DeepPoly's precision while reducing the backward
pass on some DNNs.
SkipPoly matches DeepPoly's precision on all $32$ DNNs and is as fast or faster on $16$.
%
\textbf{ZID} combines multiple abstract domains: it applies DeepZ to early
layers, IBP to intermediate layers, and DeepPoly to the final layer. This provides an intermediate
precision/runtime tradeoff: ZID is more precise than IBP on $22/32$ networks,
while running faster than DeepPoly on all $32$ and faster than DeepZ on $28$.
Its mean precision ($51\%$) lies between IBP ($23\%$) and DeepPoly ($74\%$),
providing a tradeoff point unavailable in existing libraries.
%
\textbf{ReuseCert} caches intermediate expressions generated during
DeepPoly's backsubstitution and reuses them at later layers. It is faster than
DeepPoly on $24/32$ networks, at the cost of reduced precision ($53\%$ vs.\
$74\%$ mean precision).
%
\textbf{PolyZono} maintains both polyhedral and zonotope constraints for
each neuron simultaneously. Its precision matches DeepPoly's exactly, but its
runtime is higher on all 32 networks. 
This illustrates that \compiler 
facilitates evaluating alternative designs, including those that do not provide
a favorable precision/runtime tradeoff.

\begin{figure}[t]
\centering
\includegraphics[width=\columnwidth, trim={0mm 3mm 0mm 0mm}, clip]{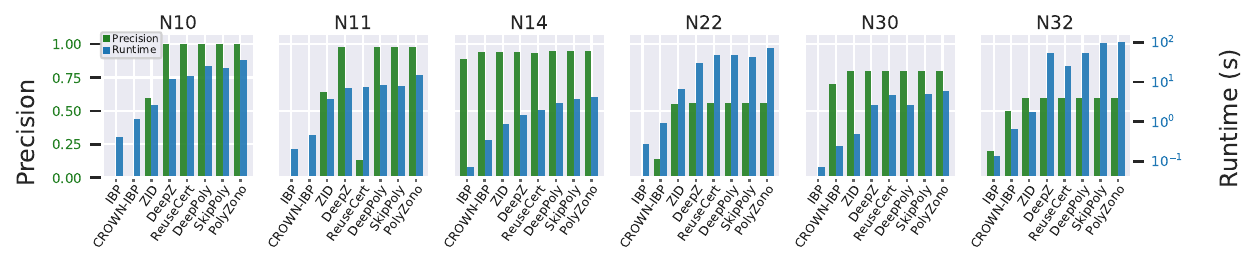}
\caption{Runtime and precision of new and existing DNN certifiers across various DNNs.}
\label{fig:designspace-pernet}
\vspace{-5mm}
\end{figure}

Fig.~\ref{fig:designspace-pernet} illustrates the design space exposed by
\compiler across six representative networks (Table~\ref{tab:repnets}), showing
runtime (blue) and precision (green) for all 8 certifiers. No single
certifier is uniformly best: inexpensive certifiers such as IBP suffice on some
DNNs, whereas more precise analyses provide substantial gains on others.
Intermediate designs such as ZID occupy useful tradeoff points between these
extremes. 
The results on all DNNs are shown in Appendix~\ref{app:evalall} (Figs.~\ref{fig:designspace-app1}, ~\ref{fig:designspace-app2}).
Overall, these results demonstrate that \compiler enables practical
exploration of a wide range of DNN certifiers without requiring separate
hand-written tensor implementations. Users specify analyses at the neuron
level and rely on the compiler to generate efficient tensor-level
implementations, substantially reducing the manual effort required to
design new certifiers.

\subsection{RQ2: Exploring a Family of Certifiers}
\label{sec:eval-explore}

A central benefit of treating a certifier as a specification is that a whole
\emph{family} of related certifiers becomes accessible by changing a parameter,
with \compiler compiling each variant automatically. Without \compiler,
implementing each variant requires a separate tensor-level implementation.

DeepPoly's precision comes from backsubstitution: to compute concrete
bounds for a neuron, it represents the neuron's bounds as a linear expression
over the previous layer's neurons and repeatedly substitutes those neurons'
bounds while traversing backward through the DNN. Each additional
backsubstitution step tightens the bound, but increases computation.
So, the number of backsubstitution layers represents a tunable
precision/runtime tradeoff. We define the family $\mathit{DeepPoly}_p$, which
stops backsubstitution after $p$ layers. Each variant is a one-line
change to the specification, and \compiler generates the
corresponding implementation automatically.

\begin{figure}[t]
\centering
\includegraphics[width=\columnwidth,trim={0in 3mm 0in 0in},clip]{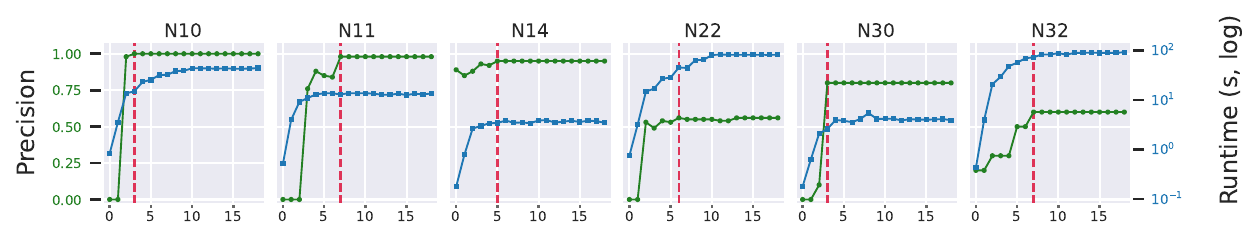}
\caption{Runtime and Precision of running the DeepPoly algorithm by the number of layers backsubstituted.}
\label{fig:psweep}
\vspace{-2mm}
\end{figure}

Fig.~\ref{fig:psweep} shows the runtime (blue) and precision (green) of
DeepPoly$_p$ as $p$ varies from $0$ to the DNN length. We report six
representative DNNs (Table~\ref{tab:repnets}) in the main text and all $32$ in
Appendix~\ref{app:evaldeeppoly} (Figs.~\ref{fig:psweep-all1},~\ref{fig:psweep-all2}). Runtime increases with $p$ due to the additional
backsubstitution steps. On standard-trained DNNs, precision generally improves
with larger $p$, while on PGD-trained DNNs it improves initially and then
plateaus, making further backsubstitution unnecessary. These trends hold across
both fully-connected and convolutional networks. 

\begin{figure}[t]
\centering
\includegraphics[width=\textwidth]{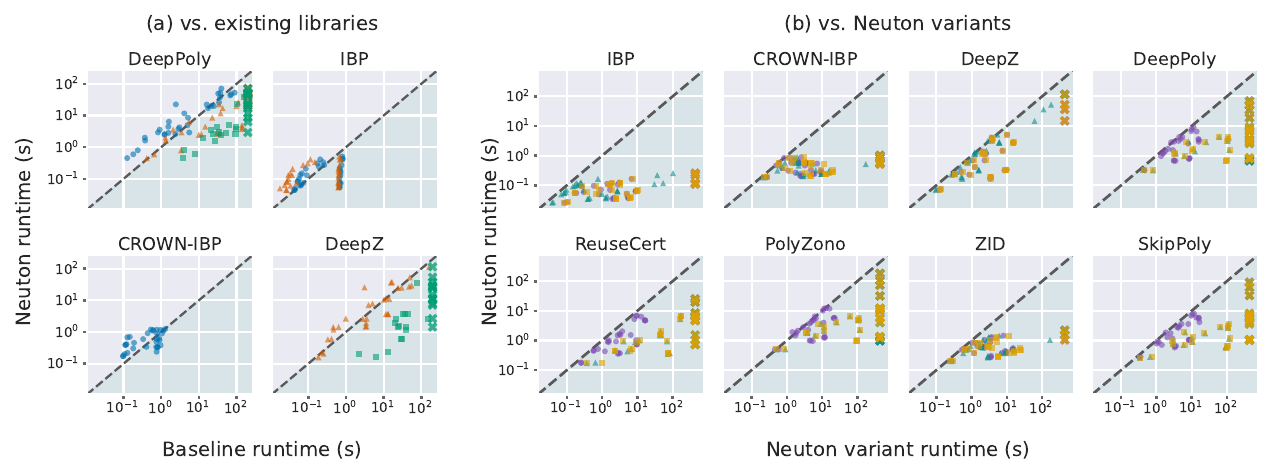}%
\makebox[0pt]{\begin{minipage}[t]{0pt}\vspace{0pt}\phantomsubcaption\label{fig:perf-alt}\phantomsubcaption\label{fig:ablation-scatter}\end{minipage}}
\caption[Per-DNN runtime of Neuton vs. existing libraries and vs. its variants.]{%
  Per-DNN runtime of \compiler vs.\ (a) existing libraries (\protect\mkAuto~$=$ auto\_LiRPA, \protect\mkIvan~$=$ IVAN,
  \protect\mkEran~$=$ ERAN) and (b) its variants (\protect\mkNS~$=$ \textbf{N--S}, \protect\mkNR~$=$ \textbf{N--R},
  \protect\mkNRS~$=$ \textbf{N--R--S}).
  Each point is one of the 32 DNNs (log--log axes); a
  point \emph{below} the dashed diagonal means \compiler is faster.
  \textsf{x} markers denote that the variant times out (200 s).}
\label{fig:perf-ablation}
\vspace{-3mm}
\end{figure}
\subsection{RQ3: Performance Against Hand-Optimized Libraries}
\label{sec:eval-perf}

We compare the performance of \compiler executables against hand-optimized implementations of existing certifiers in auto\_LiRPA~\cite{lirpagithub}, IVAN~\cite{ivangithub}, and ERAN~\cite{eran}.
Each of these hand-codes a fixed set of certifiers, whereas \compiler generates all eight from short neuron-level specifications (Table~\ref{tab:coverage}).
Fig.~\ref{fig:perf-alt} compares per-DNN runtimes; points below the diagonal indicate DNNs on which \compiler is faster. Appendix~\ref{app:rq3-tables} (Tables~\ref{tab:rq3-summary},~\ref{tab:rq3-full}) reports absolute runtimes for all benchmarks.

ERAN implements DeepZ and DeepPoly. \compiler is faster on every DNN that both tools complete, achieving mean speedups of $24.7\times$ for DeepZ and $13.3\times$ for DeepPoly. Moreover, ERAN times out on 30 of the 64 certifier--DNN combinations, whereas \compiler completes all of them.
IVAN implements IBP, DeepZ, and DeepPoly. \compiler achieves mean speedups of $2.7\times$, $1.3\times$, and $7.5\times$, respectively.
auto\_LiRPA implements IBP, CROWN-IBP, and DeepPoly. It implements algorithmic optimizations that are outside \compiler's scope. Even so, \compiler achieves mean speedups of $3.0\times$ on IBP and $1.0\times$ on CROWN-IBP, while reaching $0.73\times$ the speed of auto\_LiRPA on DeepPoly.

The remaining performance gap stems from \compiler's generality. Unlike existing libraries, which implement a fixed collection of dense analyses, \compiler supports arbitrary certifiers specified at the neuron level. This flexibility introduces two sources of overhead. First, \cf allows configurable traversal strategies and stopping conditions, requiring the generated code to evaluate user-specified stopping conditions during each iteration of the while-loop generated from the \traverse construct. Second, because sparsity patterns in polyhedral and symbolic expressions are not known statically, \compiler relies on \sparse, incurring modest overhead from sparse tensor operations. Despite these costs, \compiler is comparable with hand-optimized implementations while generating efficient tensor code for a broad class of certifiers from compact specifications.

\paragraph{Memory.}
We also measure the peak memory of every run. On the six representative DNNs, peak memory for IBP and CROWN-IBP is dominated by the PyTorch runtime itself, and all tools are within $5\%$ of each other. For the memory-intensive DeepZ and DeepPoly, \compiler uses less memory than the baselines: IVAN needs $1.29\times$ (DeepZ) and $2.13\times$ (DeepPoly) the memory of \compiler, and auto\_LiRPA $1.16\times$ (DeepPoly; geometric means). On N32, the largest DNN, \compiler's DeepPoly peaks at $21.3$\,GB against $24.0$\,GB for auto\_LiRPA, while IVAN times out. ERAN analyzes one input at a time; at batch size $1$ it uses $2.05\times$ (DeepZ) and $2.13\times$ (DeepPoly) the memory of \compiler. The same trends hold on all 32 DNNs (Appendix~\ref{app:memory}, Fig.~\ref{fig:mem-scatter} and Table~\ref{tab:mem-baselines}).

\subsection{RQ4: Ablation of \sparse and Rewrites}
\label{sec:ablation}

We evaluate the contributions of \sparse and the compiler's domain-specific rewrites through an ablation study. We compare four compiler variants: (i) \compiler (N), (ii) \textsc{\compiler-NoSparse} (N--S), (iii) \textsc{\compiler-NoRewrite} (N--R), and (iv) \textsc{\compiler-NoRewrite-NoSparse} (N--R--S). All variants are evaluated on every certifier and DNN with batch size $10$. Fig.~\ref{fig:ablation-scatter} reports per-DNN runtime relative to \compiler. Complete runtimes for every variant are provided in Appendix~\ref{app:rq4-tables} (Tables~\ref{tab:rq4-summary},~\ref{tab:rq4-fullA},~\ref{tab:rq4-fullB}).

Across all evaluations, removing \sparse increases runtime by a mean of $11.1\times$, removing the rewrites by $26.4\times$, and removing both by $30.1\times$. The variants also time out on several runs. Across the $256$ certifier--DNN combinations ($8$ certifiers $\times\ 32$ DNNs), N--S times out on an additional $30$ runs, N--R on $68$, and N--R--S on $71$.
The impact of the two optimizations differs across certifiers. Disabling the rewrites primarily affects the polyhedral and backsubstitution-based certifiers (DeepPoly, ReuseCert, SkipPoly, and PolyZono), which incur a mean slowdown of $32\times$ and fail or time out on an additional $12$--$18$ DNNs each. In contrast, removing \sparse degrades performance across all certifiers, with the largest effect on IBP (a mean slowdown of $49\times$, reaching $130$--$157\times$ on the largest convolutional DNNs) and mean slowdowns of $2.7$--$3.8\times$ for the polyhedral certifiers.

The N--R--S configuration serves as an approximation of a straightforward tensorized implementation of the neuron-level specification, without either compiler optimization. Even this variant performs tensor-level execution rather than neuron-by-neuron interpretation. \compiler is, on average, $30.1\times$ faster than N--R--S, indicating that these compiler optimizations contribute substantially to the efficiency of the generated implementations.

\paragraph{Memory.}
We measure the peak memory of all four variants for all eight certifiers on the six representative DNNs (detailed numbers in Appendix~\ref{app:memory}, Table~\ref{tab:mem-ablation}).
Without \sparse, peak memory grows by $8.6\times$ (geometric mean over completed runs), and N--S runs out of memory on N32 for every certifier. Without the rewrites, the backsubstitution-based certifiers materialize the large intermediate tensors of the repeat--multiply--reduce sequences, increasing memory by $31$--$130\times$; 10 of the 48 N--R runs and 12 of the 48 N--R--S runs run out of memory. \sparse and the rewrites are therefore what make the larger DNNs feasible at all.

\section{Related Work}
\label{sec:related}

\paragraph{DNN certification and its implementations.}
Numerous abstract interpretation-based DNN certifiers have been developed over the years. They span a range of abstract domains, including intervals~\cite{interval}, zonotopes~\cite{deepz}, and polyhedra~\cite{deeppoly,zhang2018crown,alphacrown,crownibp,wang2021beta,fastcrown}, together with refinements that trade cost for precision~\cite{refinezono,prima}.
These are among the most scalable approaches to formally certifying DNNs; complete
methods based on SMT or MILP, such as Reluplex~\cite{reluplex} and
Marabou~\cite{marabou}, give exact answers but scale to far smaller networks.
Efficient implementations of the abstract-interpretation certifiers are provided by
hand-optimized libraries such as ERAN~\cite{eran}, auto\_LiRPA~\cite{lirpagithub}, and
IVAN~\cite{ivangithub}. Each library supports
a fixed set of domains and transformers, and adding or modifying a certifier
requires re-implementing it in low-level tensor code. 
\cf~\cite{constraintflowsas} was recently proposed as a DSL for specifying DNN certifiers. Although it defines operational semantics~\cite{constraintflow} that can be used to build an interpreter, its neuron-wise execution model does not scale to practical DNNs.
\compiler complements these libraries and specification DSLs. Rather than hand-coding each certifier, it generates a tensor-level implementation from a neuron-level specification while achieving performance comparable with existing implementations (\S\ref{sec:eval-perf}).


\paragraph{Sparse tensor formats.}
Classical sparse formats like COO~\cite{coo}, CSR~\cite{csr}, and
CSF~\cite{csf} represent sparsity at the level of individual elements and do
not exploit the dense-block structure of certifier tensors. Block formats such as
BCSR~\cite{bcsr} compress at the block level but assume uniformly sized blocks;
because block sizes in DNN certifiers vary with layer type and size, enforcing
uniformity subdivides blocks and destroys internal sparsity. Doubly-compressed
formats like DCSR~\cite{dcsr} similarly rely on regular block structure. All of
these formats moreover assume a zero default, whereas certifier tensors use
non-zero defaults such as $\pm\infty$. \sparse is designed for exactly this
setting: variable-sized blocks over an arbitrary default value, each stored in its
most compact form (\S\ref{sec:sparserepresentation}).

\paragraph{Tensor compilers and frameworks.}
Tensor compilers like Halide~\cite{halide}, TVM~\cite{tvm},
XLA~\cite{xla}, Triton~\cite{triton}, NumPy~\cite{numpy}, PyTorch~\cite{pytorch} take programs already expressed as tensor operations
and lower them to efficient hardware code, optimizing schedules, fusion, and
memory layout. They assume the tensor-level program already exists and do not
infer it from a higher-level description. \compiler addresses the step that
precedes them: it synthesizes the tensor program including the shapes, broadcasts, and
reductions from a neuron-level specification. Its output could in turn be
handed to such a compiler for further hardware-specific optimization, which we
leave to future work.

\section{Conclusion}
\label{sec:conclusion}

We presented \compiler, a compiler that bridges the gap between the neuron-level
design of DNN certifiers and their tensor-level implementation. \compiler pairs a
stack-based IR and a shape analysis that lifts neuron-level \cf specifications into
efficient tensor programs with \sparse, a sparse tensor format tailored to the
layer-aligned, variable-block, non-zero-default sparsity that certifiers produce,
and with domain-specific rewrites that collapse common patterns into efficient
primitives. Across eight certifiers and 32 DNNs, \compiler compiles four
widely-used certifiers, produces the first executable implementations of four more: two prototypes previously specified but never run, and two new designs and
matches the performance of hand-optimized libraries while requiring only a short
specification per certifier. By making it practical to design, run, and compare
certifiers without manual tensor programming, \compiler turns certifier
development from a hand-implementation effort into a matter of writing a
specification.

\clearpage

\bibliographystyle{ACM-Reference-Format}
\bibliography{main}

\clearpage
\appendix

\section{IR Grammar and Metadata Auxiliary Functions}
\label{app:irgrammar}

\subsection*{A.1\enspace IR Grammar}

The complete grammar for the Intermediate Representation is given in
Fig.~\ref{fig:irgrammar}.  Each expression node also takes IR sub-expressions
as arguments; for example, $\irbinary(i_{e_1}, i_{e_2}, \oplus)$ applies
binary operator $\oplus$ to two IR expressions.  Nodes such as $\irrepeat$
and $\iradddim$ carry additional dimension attributes alongside their input
node.

\begin{figure}[H]
  \centering
  \begin{grammar}
    <IR Expression> $i_e$ ::=
      $\irconst \mid \irvar \mid \irnoise$
    \alt $\irbinary(i_{e_1},i_{e_2}) \mid \irunary(i_e) \mid \irternary(i_{e_1},i_{e_2}, i_{e_3})$
    \alt $\irdot(i_{e_1},i_{e_2}) \mid \irmult(i_{e_1},i_{e_2}) \mid \irinner(i_{e_1},i_{e_2})$
    \alt $\iradddim(i_{e_1}) \mid \irrepeat(i_{e_1}) \mid \iradddimconst(i_{e_1})$
    \alt $\irremovedim(i_{e_1}) \mid \irreduce(i_{e_1})$
    \alt $\irextractpolycoeff(i_{e_1}) \mid \irextractpolyconst(i_{e_1})$
    \alt $\irneuronpoly(i_{e_1}) \mid \irconstpoly(i_{e_1})
         \mid \ircombinepoly(i_{e_1},i_{e_2})$
    \alt $\irmapcoeff(i_{e_1}) \mid \irmapneuron(i_{e_1})
         \mid \irmapnoise(i_{e_1}) \mid \iraccess(i_{e_1})$
    \alt $\irextractsymcoeff(i_{e_1}) \mid \irextractsymconst(i_{e_1})$
    \alt $\irnoisesym(i_{e_1}) \mid \irconstsym(i_{e_1})
         \mid \ircombinesym(i_{e_1},i_{e_2})$

    <IR Statement> $i_s$ ::=
      $\irassignment(\irvar, i_e) \mid \irite(i_e, i_s)$
    \alt $\irwhile(i_e, i_{s_1}, i_{s_2})
         \mid \irtransretbasic([i_{e_1}, i_{e_2}, \cdots])$
    \alt $\irseq(i_{s_1}, i_{s_2}) \mid \irskip$
  \end{grammar}
  \caption{IR grammar.}
  \label{fig:irgrammar}
\end{figure}

\subsection*{A.2\enspace IR Metadata Auxiliary Functions}

An \emph{IrMetadata stack} $m = m_e :: m'$ is a cons-list of
IrMetadataElements (Def.~\ref{def:irme}).  The stack records one element per
logical ``nesting level'' contributed by \map and \traverse.
Fig.~\ref{fig:metadataaux} defines the auxiliary functions used by shape
analysis (Appendix~\ref{app:formalism}) and the proofs
(Appendix~\ref{app:proof-semanticpreservation}).

The \textsc{Ext}, \textsc{Lcm-elem-eq}, \textsc{Lcm-elem-uneq}, and
\textsc{Lcm-meta} rules are the formal specification of the metadata $\mathrm{ext}$
and $\lcm$ used by \matchdims\ (\S\ref{sec:alignshapes}, Fig.~\ref{fig:alignshapes}):
\textsc{Ext} pads a shorter element to match the taller one's rank,
\textsc{Lcm-elem-eq} handles same-rank elements, \textsc{Lcm-elem-uneq} handles
unequal-rank elements by padding the shorter one, and \textsc{Lcm-meta} lifts
these to full stacks element-wise.

\begin{figure}[H]
\centering
\resizebox{\linewidth}{!}{$
\begin{array}{c}
    \inferrule*[lab = \textsc{Height-1}]
    { m = \langle\rangle }
    { \height(m) = 0 }
    \hspace{1.5cm}
    \inferrule*[lab = \textsc{Height-2}]
    { m = m_e :: m_2 }
    { \height(m) = 1 + \height(m_2) }
    \\[6pt]
    \inferrule*[lab = \textsc{Get-shape-1}]
    { m = m_e :: \langle\rangle }
    { m(s) = m_e(s) }
    \hspace{1.5cm}
    \inferrule*[lab = \textsc{Get-shape-2}]
    { m = m_e :: m_2 }
    { m(s) = m_e(s) \mathbin{@} m_2(s) }
    \\[6pt]
    \inferrule*[lab = \textsc{Get-bcast-1}]
    { m = m_e :: \langle\rangle }
    { m(\beta) = m_e(\beta) }
    \hspace{1.5cm}
    \inferrule*[lab = \textsc{Get-bcast-2}]
    { m = m_e :: m_2 }
    { m(\beta) = m_e(\beta) \mathbin{@} m_2(\beta) }
    \\[6pt]
    \inferrule*[lab = \textsc{IsExp-1}]
    { }
    { \isexpanded(\langle\rangle) = \true }
    \hspace{1cm}
    \inferrule*[lab = \textsc{IsExp-2}]
    { m_e(\beta) = [1, \ldots, 1] }
    { \isexpanded(m_e) = \true }
    \hspace{1cm}
    \inferrule*[lab = \textsc{IsExp-3}]
    { \isexpanded(m_e) = t_1 \\ \isexpanded(m_2) = t_2 }
    { \isexpanded(m_e :: m) = t_1 \mathbin{\&} t_2 }
    \\[6pt]
    \inferrule*[lab = \textsc{Total-shape}]
    {
      s = [s_1, \ldots, s_n] \\
      \beta  = [\beta_1,  \ldots, \beta_n]
    }
    { s \otimes \beta = [s_1{\times}\beta_1, \ldots, s_n{\times}\beta_n] }
    \hspace{1cm}
    \inferrule*[lab = \textsc{Eq-meta}]
    { m_1(s) = m_2(s) \\ m_1(\beta) = m_2(\beta) }
    { m_1 \equiv m_2 }
    \\[6pt]
    \inferrule*[lab = \textsc{Ext}]
    {
      |m_{e'}| \geq |m_e| \\
      \forall j < |m_e|:\;
        m_e^*(s)[j] = m_e(s)[j] \;\wedge\; m_e^*(\beta)[j] = m_e(\beta)[j] \\
      \forall j \geq |m_e|:\;
        m_e^*(s)[j] = 1
        \;\wedge\; m_e^*(\beta)[j] = m_{e'}(s)[j]{\times}m_{e'}(\beta)[j]
    }
    { \mathrm{ext}(m_e, m_{e'}) = m_e^* }
    \\[6pt]
    \inferrule*[lab = \textsc{Lcm-elem-eq}]
    {
      |m_{e_1}| = |m_{e_2}| \quad
      m_{e_1}(s) \otimes m_{e_1}(\beta) = m_{e_2}(s) \otimes m_{e_2}(\beta) \\
      m_e = \langle \types, \isconst, s, \beta \rangle \\
      \forall i:\;
        m_{e_1}(s)[i] \neq m_{e_2}(s)[i]
        \implies m_{e_1}(s)[i] = 1 \vee m_{e_2}(s)[i] = 1 \\
      \forall i:\;
        m_{e_1}(s)[i] = m_{e_2}(s)[i]
        \implies s[i] = m_{e_1}(s)[i] \;\wedge\; \beta[i] = m_{e_1}(\beta)[i] \\
      \forall i:\;
        m_{e_1}(s)[i] \neq m_{e_2}(s)[i]
        \implies s[i] = m_{e_1}(s)[i] {\times} m_{e_1}(\beta)[i] \;\wedge\; \beta[i] = 1
    }
    { \lcm(m_{e_1}, m_{e_2}) = m_e }
    \\[6pt]
    \inferrule*[lab = \textsc{Lcm-elem-uneq}]
    {
      |m_{e_1}| < |m_{e_2}| \\
      m_{e_1}^{*} = \mathrm{ext}(m_{e_1}, m_{e_2})
    }
    { \lcm(m_{e_1}, m_{e_2}) = \lcm(m_{e_1}^{*}, m_{e_2}) }
    \\[6pt]
    \inferrule*[lab = \textsc{Lcm-meta}]
    {
      \height(m_1) = \height(m_2) \\
      m_e = \lcm(m_{e_1}, m_{e_2}) \quad
      m   = \lcm(m_1, m_2)
    }
    { \lcm(m_{e_1} :: m_1,\; m_{e_2} :: m_2) = m_e :: m }
\end{array}
$}
\caption{IR metadata auxiliary functions ($@$ = list concatenation).}
\label{fig:metadataaux}
\end{figure}

\begin{figure}[t]
\centering\small
  \[
    \inferrule*[Left=\textsc{AlignShapes}]{
      m_k = m_{e_k} :: m_k' \qquad 
      m_{e_k}^{*} = \mathrm{ext}(m_{e_k},\; m_{e_{3-k}}) \quad k=1,2 \\\\
      i_{e_k}^{*} =
        \iradddim(i_{e_k},\, m_{e_k}^{*})
        \;\text{if}\; |m_{e_k}| < |m_{e_{3-k}}|,
        \;\text{else}\; i_{e_k}
        \quad k=1,2 \\\\
      m = \lcm(m_{e_1}^{*} :: m_1',\; m_{e_2}^{*} :: m_2')
    }{
      \matchdims(i_{e_1},\; i_{e_2},\; m_1,\; m_2)
      = \bigl(\irrepeat(i_{e_1}^{*},\, m),\; \irrepeat(i_{e_2}^{*},\, m),\; m\bigr)
    }
    \]
\caption{Shape alignment rule.}
\label{fig:alignshapes}
\end{figure}

\newtheorem*{fmltheorem}{Theorem}

\newcommand{\fAS}{\mathsf{AS}}       
\newcommand{\fCP}{\mathsf{CP}}       
\newcommand{\fCS}{\mathsf{CS}}       
\newcommand{\fRd}{\mathsf{Rd}}       

\section{Shape Analysis: Inference Rules}
\label{app:formalism}

This appendix gives the shape analysis of \S\ref{sec:shapeanalysis} a
formal, inference-rule presentation.  It is a companion reference for
readers who prefer a judgement-based specification; the text of
\S\ref{sec:shapeanalysis} is self-contained and does not depend on it.
The IR grammar, metadata auxiliary functions, and the LCM and shape alignment
rules (including \matchdims, $\mathrm{ext}$, and $\lcm$) are in
Appendix~\ref{app:irgrammar} and Fig.~\ref{fig:alignshapes}.

\paragraph{Judgement form.}
We write
\[
  \Gamma,\;\sigma,\;F
  \;\vdash\; e
  \;\leadsto\; (i_s,\; i_e,\; m)
\]
for the \emph{compilation judgement}: under type context $\Gamma$,
IR store $\sigma$ (mapping \cf variable names to $(i_e, m)$ pairs), and
function store $F$, expression $e$ compiles to IR statement $i_s$, IR
expression $i_e$, and metadata stack $m$.
Well-formedness is guaranteed by the \cf type system throughout.
For space, we write $\fAS$ for \matchdims, $\fCP/\fCS$ for
$\ircombinepoly$/$\ircombinesym$, $\fRd$ for \irreduce, and
$e.C$/$e.k$ ($e.S$/$e.s$) for \polyexp (\symexp) coefficient/constant
extraction.

\subsection*{B.1\enspace Base Cases}

\begingroup\small
\begin{mathpar}
\inferrule*[Left=\textsc{Const}]{
  \tau_c = \mathrm{typeOf}(c)
}{
  \Gamma,\sigma,F \;\vdash\; c \;\leadsto\;
  \bigl(\irskip,\;
        \irconst(c),\;
        \langle\tau_c,\true,[1],[\currs]\rangle\bigr)
}
\\\\
\inferrule*[Left=\textsc{Var}]{
  \sigma(x) = (i_e,\, m)
}{
  \Gamma,\sigma,F \;\vdash\; x \;\leadsto\; (\irskip,\, i_e,\, m)
}
\end{mathpar}
\endgroup

\subsection*{B.2\enspace Binary Operations}

\noindent
For compactness we write $e.C \triangleq \irextractpolycoeff(e)$,
$e.k \triangleq \irextractpolyconst(e)$, and likewise
$e.S$/$e.s$ for the \symexp coefficient/constant.

\begingroup\small
\begin{mathpar}
\inferrule*[Left=\textsc{BinFloat}]{
  \Gamma,\sigma,F \vdash e_l \leadsto (i_{s_l},\, i_{e_l},\, m_l)
    \;\; (l=1,2) \\\\
  \fAS(i_{e_1},\, m_1,\, i_{e_2},\, m_2)
    = (i_{e_1}',\, i_{e_2}',\, m) \\\\
  \Gamma \vdash e_l : \tau \in \{\float,\bool\}
}{
  \Gamma,\sigma,F \;\vdash\; e_1 \oplus e_2
  \;\leadsto\;
  \bigl(\irseq(i_{s_1},\,i_{s_2}),\;
        \irbinary(i_{e_1}',\,i_{e_2}',\,\oplus),\;
        m\bigr)
}
\end{mathpar}
\endgroup

\noindent
For binary \polyexp expressions, the operation decomposes
component-wise on coefficient and constant parts:

\begingroup\small
\begin{mathpar}
\inferrule*[Left=\textsc{BinPoly}]{
  \Gamma,\sigma,F \vdash e_l \leadsto (i_{s_l},\, i_{e_l},\, m_l)
    \;\; (l=1,2) \\
  m_l^C = m_l[\types{\mapsto}\float,\;\irshape \mathbin{@} [\polys],\;\irbroadcast \mathbin{@} [1]]
    \;\; (l=1,2) \\
  m_l^k = m_l[\types{\mapsto}\float] \;\; (l=1,2) \\
  \fAS(i_{e_1}.C,\, m_1^C,\; i_{e_2}.C,\, m_2^C)
    = (i_{C1}',\, i_{C2}',\, m^C) \\
  \fAS(i_{e_1}.k,\, m_1^k,\; i_{e_2}.k,\, m_2^k)
    = (i_{k1}',\, i_{k2}',\, m^k)
}{
    \Gamma,\sigma,F \;\vdash\; e_1 \oplus e_2 : \polyexp \;\leadsto\;
    \bigl(\irseq(i_{s_1},i_{s_2}), \\
    \quad\fCP\bigl(\irbinary(i_{C1}',i_{C2}',\oplus),\;
                   \irbinary(i_{k1}',i_{k2}',\oplus)\bigr),\;
    m^k[\types{\mapsto}\polyexp]\bigr)
}
\end{mathpar}
\endgroup

\noindent
Binary \symexp expressions decompose in the same way, with $\ircombinesym$, $e.S$, $e.s$, and $\syms$ replacing their \polyexp counterparts:

\begingroup\small
\begin{mathpar}
\inferrule*[Left=\textsc{BinSym}]{
  \Gamma,\sigma,F \vdash e_l \leadsto (i_{s_l},\, i_{e_l},\, m_l)
    \;\; (l=1,2) \\
  m_l^S = m_l[\types{\mapsto}\float,\;\irshape \mathbin{@} [\syms],\;\irbroadcast \mathbin{@} [1]]
    \;\; (l=1,2) \\
  m_l^s = m_l[\types{\mapsto}\float] \;\; (l=1,2) \\
  \fAS(i_{e_1}.S,\, m_1^S,\; i_{e_2}.S,\, m_2^S)
    = (i_{S1}',\, i_{S2}',\, m^S) \\
  \fAS(i_{e_1}.s,\, m_1^s,\; i_{e_2}.s,\, m_2^s)
    = (i_{s1}',\, i_{s2}',\, m^s)
}{
    \Gamma,\sigma,F \;\vdash\; e_1 \oplus e_2 : \symexp \;\leadsto\;
    \bigl(\irseq(i_{s_1},i_{s_2}), \\
    \quad\fCS\bigl(\irbinary(i_{S1}',i_{S2}',\oplus),\;
                   \irbinary(i_{s1}',i_{s2}',\oplus)\bigr),\;
    m^s[\types{\mapsto}\symexp]\bigr)
}
\end{mathpar}
\endgroup

\subsection*{B.3\enspace Map on \polyexp}

\noindent
Write $m^{bc}$ for $m$ with the \polys dimension moved to the
broadcast factor, and $m[\types \mapsto \tau']$ for $m$ with its type
component replaced.

\begingroup\small
\begin{mathpar}
\inferrule*[Left=\textsc{Map}]{
  \Gamma,\sigma,F \vdash e' \leadsto (i_{s_1},\, i_{e_1},\, m_1) \quad
  F(f) = (n_f,\, c_f,\, e_f) \\
  m_e^{\mathrm{map}} = \langle\typeneuron,\false,[\polys],[1]\rangle \\
  m_n = m_e^{\mathrm{map}}{::}m_1^{bc} \quad
  m_c = m_e^{\mathrm{map}}{::}m_1 \\
  \sigma^{+} = \sigma\bigl[n_f \mapsto (\irmapneuron(i_{e_1}),\,m_n),\;
                            c_f \mapsto (\irmapcoeff(i_{e_1}),\,m_c)\bigr] \\
  \Gamma,\sigma^{+},F \vdash e_f
    \leadsto (i_{s_f},\; i_{e_f},\; m_e^{f}{::}m_{\mathit{out}}) \\
  i_{e_{pc}} = \fRd(i_{e_f}.C,\, m_e^{f}) \quad
  i_{e_{pz}} = \fRd(i_{e_f}.k,\, m_e^{f}) \\\\
  \fAS\ \bigl(i_{e_1}.k,\,
    m_1[\types\!\mapsto\!\float],\,
    i_{e_{pz}},\,
    m_{\mathit{out}}\bigr)
  = (i_{e_k}',\, i_{e_{pz}}',\, m')
}{
  \Gamma,\sigma,F \;\vdash\; e'.\map(f) \;\leadsto\;
  \Bigl(\irseq(i_{s_1},\,i_{s_f}),\;
    \fCP\!\bigl(
      i_{e_{pc}},\;
      \irbinary(i_{e_k}',\,i_{e_{pz}}',\,+)\bigr),\;
    m'\Bigr)
}
\end{mathpar}
\endgroup

\noindent
The premise $\Gamma,\sigma^+,F \vdash e_f \leadsto (\ldots,
m_e^{f}{::}m_{\mathit{out}})$ asserts that the returned stack has \emph{some}
element $m_e^{f}$ on top — the map-level slot introduced by $m_e^{\mathrm{map}}$
is preserved, though $m_e^{f}$ may differ from $m_e^{\mathrm{map}}$ (Lemma~\ref{lem:stackinv}).

\noindent
The \textsc{MapSym} rule for \symexp is structurally identical, with \polys
replaced by \syms, \typeneuron replaced by \typesym,
$\irmapneuron$ by $\irmapnoise$, and $\fCP$ by $\fCS$:

\begingroup\small
\begin{mathpar}
\inferrule*[lab=\textsc{MapSym}]{
  \Gamma,\sigma,F \vdash e' \leadsto (i_{s_1},\, i_{e_1},\, m_1) \quad
  F(f) = (n_f,\, c_f,\, e_f) \\\\
  m_e^{\mathrm{map}} = \langle\typesym,\false,[\syms],[1]\rangle \\\\
  m_n = m_e^{\mathrm{map}}{::}m_1^{bc} \quad
  m_c = m_e^{\mathrm{map}}{::}m_1 \\\\
  \sigma^{+} = \sigma\bigl[n_f \mapsto (\irmapnoise(i_{e_1}),\,m_n),\;
                            c_f \mapsto (\irmapcoeff(i_{e_1}),\,m_c)\bigr] \\
  \Gamma,\sigma^{+},F \vdash e_f
    \leadsto (i_{s_f},\; i_{e_f},\; m_e^{f}{::}m_{\mathit{out}}) \\
  i_{e_{pc}} = \fRd(i_{e_f}.S,\, m_e^{f}) \quad
  i_{e_{pz}} = \fRd(i_{e_f}.s,\, m_e^{f}) \\\\
  \fAS\ \bigl(i_{e_1}.s,\,
    m_1[\types\!\mapsto\!\float],\,
    i_{e_{pz}},\,
    m_{\mathit{out}}\bigr)
  = (i_{e_s}',\, i_{e_{pz}}',\, m')
}{
  \Gamma,\sigma,F \;\vdash\; e'.\map(f) \;\leadsto\;
  \Bigl(\irseq(i_{s_1},\,i_{s_f}),\;
    \fCS\!\bigl(
      i_{e_{pc}},\;
      \irbinary(i_{e_s}',\,i_{e_{pz}}',\,+)\bigr),\;
    m'\Bigr)
}
\end{mathpar}
\endgroup

\subsection*{B.4\enspace Traverse}

Because \traverse compiles to a loop, we use an auxiliary
\emph{loop-body judgement}
\[
  \Gamma,\;\sigma,\;F,\;f,\;\mathit{stop}
  \;\vdash_{\mathrm{body}}\; (i_e,\; m)
  \;\leadsto_\ell\; (i_e',\; m',\; i_s),
\]
where $\polys_\ell$ is the symbolic layer size at iteration $\ell$,
populated from the DNN graph at runtime
(\S\ref{sec:shapeanalysis-traverse}).

\begingroup\small
\begin{mathpar}
\inferrule*[lab=\textsc{TravBody}]{
  i_e^s = \textsc{IrGetPolyexpStop}(i_e,\;\mathit{stop}) \quad
  i_e^a = \textsc{IrGetPolyexpNotStop}(i_e,\;\mathit{stop}) \\\\
  m_e^{\mathrm{map}} = \langle\typeneuron,\false,[\polys_\ell],[1]\rangle \quad
  F(f) = (n_f,\, c_f,\, e_f) \\
  m_n = m_e^{\mathrm{map}}{::}m^{bc} \quad
  m_c = m_e^{\mathrm{map}}{::}m \\\\
  \sigma_\ell = \sigma\bigl[n_f \mapsto (\irmapneuron(i_e^a),\,m_n),\;
                            c_f \mapsto (\irmapcoeff(i_e^a),\,m_c)\bigr] \\
  \Gamma,\sigma_\ell,F \vdash e_f
    \leadsto (i_{s_f},\; i_{e_f},\; m_e^{\mathrm{map}}{::}m_{\mathit{out}}) \\
  i_{e_{pc}} = \fRd(i_{e_f}.C,\, m_e^{\mathrm{map}}) \quad
  i_{e_{pz}} = \fRd(i_{e_f}.k,\, m_e^{\mathrm{map}}) \\
  i_{e_{\mathrm{new}}} = \fCP(i_{e_{pc}},\; i_{e_{pz}}) \\
  \fAS(i_e^s,\, m,\; i_{e_{\mathrm{new}}},\, m_{\mathit{out}})
    = (i_s',\, i_n',\, m') \\
  i_{e,c}' = \irbinary(i_s'.C,\, i_n'.C,\, +) \quad
  i_{e,z}' = \irbinary(i_s'.k,\, i_n'.k,\, +) \\
  i_e' = \fCP(i_{e,c}',\; i_{e,z}')
}{
  \Gamma,\sigma,F,f,\mathit{stop}
  \;\vdash_{\mathrm{body}}\; (i_e,\, m)
  \;\leadsto_\ell\; (i_e',\; m',\; i_{s_f})
}
\end{mathpar}
\endgroup

\noindent
The body rule partitions the loop-carried expression into stopped lane
$i_e^s$ and active lane $i_e^a$, applies map machinery to the active
lane, and accumulates the stopped lane into the result.

\begingroup\small
\begin{mathpar}
\inferrule*[lab=\textsc{Traverse}]{
  \Gamma,\sigma,F \vdash e \leadsto (i_{s_0},\, i_{e_0},\, m_0) \quad
  F(f) = (n_f,\, c_f,\, e_f) \\
  \bigl\{\Gamma,\sigma,F,f,\mathit{stop}
    \;\vdash_{\mathrm{body}}\; (i_{e_\ell},\, m_\ell)
    \;\leadsto_\ell\; (i_{e_{\ell+1}},\, m_{\ell+1},\, i_{s_\ell})
    \bigr\}_{\ell\geq 0}
}{
  \Gamma,\sigma,F \;\vdash\; e.\traverse(f,\mathit{dir},\mathit{pri},\mathit{stop})
  \;\leadsto\;
  \bigl(\irseq(i_{s_0},\,\irwhile(i_{e_0},m_0,f,\mathit{stop})),\;
    i_{e_{\ell^*}},\; m_{\ell^*}\bigr)
}
\end{mathpar}
\endgroup

\noindent
$\ell^* = \min\{\ell \geq 0
  \mid \textsc{IrGetPolyexpNotStop}(i_{e_\ell},\mathit{stop}) = \mathbf{0}\}$
is the termination index, finite by DNN acyclicity
(Appendix~\ref{app:proof-semanticpreservation}, traverse case).

\subsection*{B.5\enspace Formal Correctness}

The compilation judgement makes Theorem~\ref{thm:correctness} precise.

\begin{fmltheorem}
  If\; $\Gamma,\sigma,F \vdash P \leadsto (i_s,\, i_e,\, m)$\; and
  $\varrho$ is consistent with $\sigma$
  (equation~\eqref{eq:consistent}), then $\mathcal{L}(P,\, i_e,\, m)$
  holds (equation~\eqref{eq:lifting}).
\end{fmltheorem}

\noindent
\emph{Proof.}
By induction on the derivation of
$\Gamma,\sigma,F \vdash P \leadsto (i_s,i_e,m)$.
Each rule above corresponds to exactly one case in
Appendix~\ref{app:proof-semanticpreservation}:
\textsc{Const}/\textsc{Var} $\to$ constant/variable cases;
\textsc{BinFloat}/\textsc{BinPoly} $\to$ binary operation cases;
\textsc{Map} $\to$ \map case (using Lemmata~\ref{lem:alignshapes}
and~\ref{lem:stackinv});
\textsc{TravBody}+\textsc{Traverse} $\to$ traverse case
(the body rule corresponds to the inductive step of $\mathcal{I}_\ell$).
\qed

\clearpage
\section{Proof of Theorem~\ref{thm:correctness} (Semantic Preservation)}
\label{app:proof-semanticpreservation}

\providecommand{\sem}[1]{[\![#1]\!]}

\paragraph{Proof overview.}
The proof uses induction
on the \cf expression. 
We define the \emph{correspondence invariant}
$\mathcal{L}$: after a
sub-expression's statements execute, evaluating its result and slicing at neuron $i$
yields $v_i$. The base cases hold by consistency of
$\varrho_0$, and three lemmas carry the rest. For binary operations,
Lemma~\ref{lem:alignshapes} (\matchdims\ Correctness) shows that shape alignment
preserves $\mathcal{L}$.
The \map\ case reduces over the axis introduced when
the \map\ is entered, and its correctness follows from two facts:
Lemma~\ref{lem:stackinv} (Stack Invariant) shows the map level stays on top of the
metadata stack, even under nested \map{}s; and Lemma~\ref{lem:mapaxis} (Map-Axis Materialization) shows
that axis is physically \emph{stored} rather than broadcast, so the reduction sums
the correct per-pair contributions. Finally, \traverse\
applies the \map\ argument at each loop iteration.

\subsection{\enspace Notation and Formal Semantics}

\paragraph{Layer sizes and index sets.}
Fix a DNN layer with $N$ current-layer neurons
($N = \currs$ at runtime) and $N'$ previous-layer neurons ($N' = \prevs$).
Let $[k]$ denote $\{0, 1, \ldots, k-1\}$.
Write $P$ for the polyhedral expression size ($\polys$).
We use $i \in [N]$ for neuron indices and $j \in [P]$ for polyhedral component indices.

\paragraph{Value domains.}
Each \cf type $\types$ has a value domain $\mathcal{V}_\types$:
$\mathcal{V}_\float = \mathbb{R}$;
$\mathcal{V}_\polyexp = \mathbb{R}^P \times \mathbb{R}$ (coefficient vector and constant);
$\mathcal{V}_\typeneuron = [N']$ (previous-layer neuron index);
and analogously for \symexp and \bool.

\paragraph{CF neuron-level semantics.}
The big-step semantics of \cf is given in~\cite{constraintflow}.
A \emph{neuron-level environment} $\rho$ maps \cf variable names to elements of $\mathcal{V}_\types$.
We write $(\Gamma, \fstore) \vdash e, \rho \Downarrow^{\mathrm{nl}} v$ for the judgment
that expression $e$ evaluates to $v \in \mathcal{V}_\types$ under type context $\Gamma$,
function store $\fstore$, and environment $\rho$.
For a well-typed \cf specification $P$ and abstract state $\mathbf{s}$, the
neuron-level result for neuron $i$ is:
\[
  (\Gamma, \fstore) \vdash P,\; \rho_i(\mathbf{s}) \Downarrow^{\mathrm{nl}} v_i,
\]
where $\rho_i(\mathbf{s})$ maps each \cf variable $x$ to its value at neuron $i$ in $\mathbf{s}$.
The \cf big-step semantics is deterministic, so we may write
$\sem{e}^{\mathrm{nl}}_{\rho}$ for the unique value $v$ with
$(\Gamma,\fstore)\vdash e,\rho \Downarrow^{\mathrm{nl}} v$, and likewise
$\sem{\irexpression}^{\mathrm{ir}}_{\varrho}$ for the unique tensor $T$ with
$\varrho \vdash \irexpression \Downarrow^{\mathrm{ir}} T$.
We use these functional forms wherever a value is required.

\paragraph{IR execution semantics.}
The generated program is executed over a \emph{tensor environment} $\varrho$ --- the
runtime symbol table mapping IR variables to concrete tensors. Its meaning is given
by two mutually recursive judgements: expression evaluation
$\varrho \vdash \irexpression \Downarrow^{\mathrm{ir}} T$, which reads $\varrho$ and
returns a tensor, and statement execution
$\varrho \vdash i_s \Downarrow^{\mathrm{ir}} \varrho'$, which updates $\varrho$
exactly as an assignment updates a symbol table. Figure~\ref{fig:exec-semantics}
gives the rules for the core nodes.
The remaining tensor expression nodes evaluate to the correspondingly named backend
operation applied to their evaluated arguments, with axis and shape parameters read
from the metadata $m_e$:
$\irrepeat(\irexpression, m_e)$ replicates $T$ along the axes specified by
$m_e$'s broadcast factor $\irbroadcast$;
$\iradddim(\irexpression, m_e)$ inserts a singleton axis;
$\irreduce(\irexpression, m_e)$ sums along the axis identified by $m_e$;
and $\ircombinepoly, \irextractpolycoeff, \irextractpolyconst$ pack and unpack the
$(C, k)$ representation of polyhedral expressions.

\begin{figure}[h]
\centering\small
\begin{mathpar}
\inferrule*[lab=\textsc{E-Const}]{ }{\varrho \vdash \irconst(c) \Downarrow^{\mathrm{ir}} \bar{c}}
\and
\inferrule*[lab=\textsc{E-Var}]{ }{\varrho \vdash x \Downarrow^{\mathrm{ir}} \varrho(x)}
\and
\inferrule*[lab=\textsc{E-Bin}]{
  \varrho \vdash i_{e_1} \Downarrow^{\mathrm{ir}} T_1 \\ \varrho \vdash i_{e_2} \Downarrow^{\mathrm{ir}} T_2
}{
  \varrho \vdash \irbinary(i_{e_1}, i_{e_2}, \oplus) \Downarrow^{\mathrm{ir}} T_1 \mathbin{\oplus} T_2
}
\and
\inferrule*[lab=\textsc{E-Reduce}]{
  \varrho \vdash i_e \Downarrow^{\mathrm{ir}} T
}{
  \varrho \vdash \irreduce(i_e, m_e) \Downarrow^{\mathrm{ir}} \mathrm{sum}(T,\, \mathrm{axis}(m_e))
}
\\\\
\inferrule*[lab=\textsc{S-Skip}]{ }{\varrho \vdash \irskip \Downarrow^{\mathrm{ir}} \varrho}
\and
\inferrule*[lab=\textsc{S-Assign}]{
  \varrho \vdash i_e \Downarrow^{\mathrm{ir}} T
}{
  \varrho \vdash \irassignment(x, i_e) \Downarrow^{\mathrm{ir}} \varrho[x \mapsto T]
}
\and
\inferrule*[lab=\textsc{S-Seq}]{
  \varrho \vdash i_{s_1} \Downarrow^{\mathrm{ir}} \varrho' \\ \varrho' \vdash i_{s_2} \Downarrow^{\mathrm{ir}} \varrho''
}{
  \varrho \vdash \irseq(i_{s_1}, i_{s_2}) \Downarrow^{\mathrm{ir}} \varrho''
}
\\\\
\inferrule*[lab=\textsc{S-While-T}]{
  \varrho \vdash i_e \Downarrow^{\mathrm{ir}} \mathbf{true} \\
  \varrho \vdash i_s \Downarrow^{\mathrm{ir}} \varrho' \\
  \varrho' \vdash \irwhile(i_e, i_s) \Downarrow^{\mathrm{ir}} \varrho''
}{
  \varrho \vdash \irwhile(i_e, i_s) \Downarrow^{\mathrm{ir}} \varrho''
}
\and
\inferrule*[lab=\textsc{S-While-F}]{
  \varrho \vdash i_e \Downarrow^{\mathrm{ir}} \mathbf{false}
}{
  \varrho \vdash \irwhile(i_e, i_s) \Downarrow^{\mathrm{ir}} \varrho
}
\end{mathpar}
\caption{IR execution semantics: expression evaluation
  $\varrho \vdash i_e \Downarrow^{\mathrm{ir}} T$ (reads $\varrho$) and statement
  execution $\varrho \vdash i_s \Downarrow^{\mathrm{ir}} \varrho'$ (updates
  $\varrho$). $\bar{c}$ is the scalar tensor $c$; $\oplus$ is applied element-wise;
  $\irite$ is analogous to \textsc{S-While}. Nodes not shown evaluate to the
  like-named backend op on their evaluated arguments.}
\label{fig:exec-semantics}
\end{figure}

\paragraph{Effective tensor.}
For $m_e = \langle \types, \isconst, [s_1,\ldots,s_r],
[\beta_1,\ldots,\beta_r]\rangle$ and $T$ of shape $[s_1,\ldots,s_r]$,
define the \emph{effective tensor}
$\mathsf{eff}(T, m_e) \in (\mathcal{V}_\types)^{s_1\beta_1 \times \cdots \times s_r\beta_r}$ by:
\begin{equation}
\label{eq:eff}
  \mathsf{eff}(T, m_e)[j_1,\ldots,j_r]
  \;\triangleq\; T\!\left[\left\lfloor\tfrac{j_1}{\beta_1}\right\rfloor,\ldots,
                           \left\lfloor\tfrac{j_r}{\beta_r}\right\rfloor\right],
  \qquad j_l \in [0, s_l\beta_l).
\end{equation}
This is the tensor obtained by logically replicating each stored entry
$\beta_1 \times \cdots \times \beta_r$ times.
For a stack of height 0, $\mathsf{eff}$ is the identity.

\paragraph{Stacked effective tensor.}
For a metadata stack $m = m_{e,k}{::}\cdots{::}m_{e,1}$ of height~$k$
(top at left), where $m_{e,l}$ has rank~$r_l$, the tensor $T$ has axes ordered
from the \emph{bottom} element ($m_{e,1}$, outermost/first axis) to the
\emph{top} element ($m_{e,k}$, innermost/last axis).
We extend $\mathsf{eff}$ to stacks:
\begin{equation}
\label{eq:effstack}
  \mathsf{eff}(T,\;m_{e,k}{::}\cdots{::}m_{e,1})
  \bigl[\mathbf{j}^{(1)},\ldots,\mathbf{j}^{(k)}\bigr]
  \;\triangleq\;
  T\!\left[
    \left\lfloor\tfrac{j^{(1)}_1}{\beta_{1,1}}\right\rfloor,
    \ldots,
    \left\lfloor\tfrac{j^{(1)}_{r_1}}{\beta_{1,r_1}}\right\rfloor,
    \;\ldots,\;
    \left\lfloor\tfrac{j^{(k)}_1}{\beta_{k,1}}\right\rfloor,
    \ldots,
    \left\lfloor\tfrac{j^{(k)}_{r_k}}{\beta_{k,r_k}}\right\rfloor
  \right],
\end{equation}
where $j^{(l)}_d \in [0,s_{l,d}\beta_{l,d})$.
For $k=1$ this is exactly~\eqref{eq:eff}.

\paragraph{Index convention.}
All statements below use the \emph{whole-stack} effective tensor
$\mathsf{eff}(T,m)$ of~\eqref{eq:effstack}, whose axes run from the bottom stack
element (outermost, first axis) to the top element (innermost, last axis). We
write $\mathsf{eff}(T,m)[i]$ with $i\in[N]$ for the slice at the outermost
\emph{neuron} axis --- the first axis, governed by the bottom element, of
effective size~$N$. The notation $[i]$ may abbreviate a multi-index when inner
axes remain: the remaining indices then range over the axes contributed by the
higher stack elements and carry the internal structure of the value at neuron~$i$
(e.g.\ its polyhedral coefficients).
Inside a \map body the stack carries the map element on top,
$\irme[\mathit{map}]{::}m_{\mathit{out}}$: by~\eqref{eq:effstack} the bottom
element $m_{\mathit{out}}$ governs the outer neuron axis and $\irme[\mathit{map}]$
the inner map axis, so we write
$\mathsf{eff}(T,\irme[\mathit{map}]{::}m_{\mathit{out}})[i,j]$ with $i\in[N]$ and
$j\in[P]$.
Using the whole stack rather than $\mathrm{top}(m)$ is essential, not cosmetic:
when the stack has height~$>1$ the tensor $T$ carries one axis-block per stack
element, so its rank is the \emph{sum} of the element ranks and strictly exceeds
$\mathrm{rank}(\mathrm{top}(m))$. Consequently $\mathsf{eff}(\cdot,\mathrm{top}(m))$
--- the single-element form~\eqref{eq:eff}, which requires its tensor argument to
have exactly $\mathrm{rank}(\mathrm{top}(m))$ axes --- is undefined on such a $T$.
The whole-stack form~\eqref{eq:effstack} is the well-defined object throughout;
for a height-$1$ stack the two coincide.

\paragraph{Consistent environments.}
We say that a tensor environment $\varrho$ is \emph{consistent} with per-neuron
CF environments $\{\rho_i\}_{i \in [N]}$ and IR store $\sstore$ if for
every \cf variable $x$ with $\sstore(x) = (\irexpression[x], \irm_x)$:
\begin{equation}
\label{eq:consistent}
  \forall\, i \in [N]:\quad
  \mathsf{eff}(\sem{\irexpression[x]}^{\mathrm{ir}}_\varrho,\; \irm_x)[i]
  \;=\; \rho_i(x),
\end{equation}
where $\mathsf{eff}(\cdot,\irm_x)$ is the whole-stack effective
tensor~\eqref{eq:effstack} and $[i]$ denotes the $i$-th slice along the first
(outermost, neuron) effective dimension of size $N$.

\paragraph{Correspondence invariant.}
Let $\mathcal{L}(e, \irexpression, \irm)$ denote the predicate:
\emph{for all tensor environments $\varrho$ consistent with
$\{\rho_i\}_{i \in [N]}$, and for all $i \in [N]$,}
\begin{equation}
\label{eq:lifting}
  \varrho \vdash \irexpression \Downarrow^{\mathrm{ir}} T
  \;\implies\;
  \mathsf{eff}(T,\; \irm)[i]
  \;=\; v_i,
  \quad\text{where }(\Gamma,\fstore) \vdash e, \rho_i \Downarrow^{\mathrm{nl}} v_i.
\end{equation}
Theorem~\ref{thm:correctness} runs the compiled statement $i_s$ before evaluating
the result $i_e$. Every shape-analysis rule emits $i_s$ as a sequence of
assignments to \emph{fresh} IR variables (temporaries and loop variables), so by
\textsc{S-Assign}/\textsc{S-Seq} (Fig.~\ref{fig:exec-semantics}) the execution
$\varrho_0 \vdash i_s \Downarrow^{\mathrm{ir}} \varrho$ only \emph{extends} the
environment with those bindings, preserving the consistency~\eqref{eq:consistent}
of all pre-existing bindings. It therefore suffices to establish, for the top-level
$(\hat{\irexpression}, \irm)$ under the resulting consistent $\varrho$, the
\emph{correspondence invariant} $\mathcal{L}(P, \hat{\irexpression}, \irm)$, which
immediately gives $\mathsf{eff}(T,\irm)[i] = v_i$ for all
$i \in [N]$. Accordingly the induction below is phrased on expressions, with the
understanding that each sub-expression's statements are executed --- extending
$\varrho$ with fresh bindings --- before its result expression is read.

\subsection{\enspace Lemma: \matchdims Correctness}

\begin{lemma}[\matchdims\ Correctness]
\label{lem:alignshapes}
Suppose $\mathcal{L}(e_1, \irexpression[1], \irm_1)$ and
$\mathcal{L}(e_2, \irexpression[2], \irm_2)$ hold. \\
If $\matchdims(\irexpression[1], \irexpression[2], \irm_1, \irm_2)$
returns $(\irexpression[1]', \irexpression[2]', \irm)$, then
$\mathcal{L}(e_1, \irexpression[1]', \irm)$ and
$\mathcal{L}(e_2, \irexpression[2]', \irm)$ both hold.
\end{lemma}

\begin{proof}
By induction on $\max(\height(\irm_1), \height(\irm_2))$.

\smallskip
\textit{Base case} $(\irm_1 = \irm_2 = \langle\rangle)$:
Both stacks are empty; the algorithm returns $(\irexpression[1], \irexpression[2], \langle\rangle)$
unchanged. The effective tensors are scalars and $\mathsf{eff}(T, \langle\rangle)=T$, so
$\mathcal{L}(e_l, \irexpression[l], \langle\rangle)$ is inherited from the hypotheses. 

\smallskip
\textit{Inductive step}:
Write $\irm_l = m_{e,l} :: \irm_l'$ for $l = 1, 2$.
Let $m_{e,l} = \langle \types_l, \isconst_l,
[s_1^{(l)}, \ldots, s_{r_l}^{(l)}],
[\beta_1^{(l)}, \ldots, \beta_{r_l}^{(l)}]\rangle$.

\textit{Whole-stack bookkeeping.}
Each step of \matchdims\ modifies only the \emph{top} element $m_{e,l}$ and wraps
$\irexpression[l]$ in $\iradddim$/$\irrepeat$ nodes acting on the corresponding
innermost axis-block, leaving the tail $\irm_l'$ and its axis-blocks untouched.
We therefore split a whole-stack effective index as
$(\mathbf{J}';\,\mathbf{a})$, where $\mathbf{J}'$ ranges over the outer axes
governed by $\irm_l'$ and $\mathbf{a}$ over the innermost axes governed by the top
element; by~\eqref{eq:effstack} the two blocks are read independently. It suffices
to verify each alignment identity on the top block $\mathbf{a}$ for an arbitrary
fixed $\mathbf{J}'$: the whole-stack identity then follows, and with it the neuron
slice $[i]$ --- a component of $\mathbf{J}'$ when the stack has height $>1$, or of
$\mathbf{a}$ when the top element is the sole element. Below we display the
top-block computation; the tail index $\mathbf{J}'$ is carried unchanged.

\textit{Step 1: Rank alignment.}
Suppose without loss of generality that $r_1 < r_2$.
The algorithm appends $r_2 - r_1$ dimensions to $m_{e,1}$, setting
$s_j^{(1)} \gets 1$ and $\beta_j^{(1)} \gets s_j^{(2)}\beta_j^{(2)}$
for $j > r_1$, and wraps $\irexpression[1]$ in $\iradddim$.
Let the updated element be $m_{e,1}^+$ and the updated expression $\irexpression[1]^+$.

We claim $\mathcal{L}(e_1, \irexpression[1]^+, m_{e,1}^+ :: \irm_1')$.
Fix any consistent $\varrho$ and let $\varrho \vdash \irexpression[1] \Downarrow^{\mathrm{ir}} T_1$
and $\varrho \vdash \irexpression[1]^+ \Downarrow^{\mathrm{ir}} T_1^+$.
By the semantics of $\iradddim$, $T_1^+$ is $T_1$ with $r_2 - r_1$ singleton dimensions
prepended/appended as specified. For any $j_1,\ldots,j_{r_2}$ with $j_l \in [s_l^+\beta_l^+)$:
\[
  \mathsf{eff}(T_1^+, m_{e,1}^+)[j_1,\ldots,j_{r_2}]
  = T_1^+\!\left[\left\lfloor\tfrac{j_1}{\beta_1^+}\right\rfloor,\ldots,
               \left\lfloor\tfrac{j_{r_2}}{\beta_{r_2}^+}\right\rfloor\right].
\]
For indices $j > r_1$ we have $s_j^+ = 1$, so $\lfloor j_j / \beta_j^+ \rfloor = 0$,
meaning those indices collapse to 0 in $T_1^+$ — the same as reading $T_1$ at that position.
For $j \leq r_1$ the broadcast factor is unchanged, so:
\[
  \mathsf{eff}(T_1^+, m_{e,1}^+)[j_1,\ldots,j_{r_2}]
  = \mathsf{eff}(T_1, m_{e,1})[j_1,\ldots,j_{r_1}].
\]
This top-block identity holds for every fixed tail index $\mathbf{J}'$, and
$\iradddim$ leaves the tail read of~\eqref{eq:effstack} unchanged, so the
whole-stack effective tensors agree:
$\mathsf{eff}(T_1^+, m_{e,1}^+{::}\irm_1') = \mathsf{eff}(T_1, m_{e,1}{::}\irm_1')$.
In particular, taking the neuron slice (the first effective dimension, size $N$):
\[
  \mathsf{eff}(T_1^+, m_{e,1}^+{::}\irm_1')[i]
  = \mathsf{eff}(T_1, \irm_1)[i]
  = \sem{e_1}^{\mathrm{nl}}_{\rho_i},
\]
where the last equality holds by hypothesis $\mathcal{L}(e_1, \irexpression[1], \irm_1)$.
Hence $\mathcal{L}(e_1, \irexpression[1]^+, m_{e,1}^+ :: \irm_1')$ holds.

\textit{Step 2: Broadcast alignment.}
After rank alignment, both top elements have rank $r = r_1 = r_2$.
The algorithm computes $m_e = \lcm(m_{e,1}^+, m_{e,2})$ dimension-by-dimension.
For each dimension $j \in [r]$:
\begin{itemize}[topsep=1pt,itemsep=0pt]
\item If $s_j^{(1)} = s_j^{(2)}$ (same intrinsic size): the LCM leaves this
  dimension unchanged, and $\irexpression[l]$ is not wrapped.
\item If $s_j^{(1)} = 1$ and $s_j^{(2)} > 1$ (or vice versa): the algorithm
  sets $s_j = s_j^{(2)}$ and $\beta_j = \beta_j^{(2)}$ in $m_e$, and
  wraps $\irexpression[1]^+$ in $\irrepeat(\cdot, m_e)$ along axis $j$.
\end{itemize}
Let $\irexpression[1]'' = \irrepeat(\irexpression[1]^+, m_e)$.
We claim $\mathcal{L}(e_1, \irexpression[1]'', m_e :: \irm_1')$.
For any consistent $\varrho$ and $\varrho \vdash \irexpression[1]'' \Downarrow^{\mathrm{ir}} T_1''$:
by the semantics of $\irrepeat$, a stored axis $j$ of intrinsic size $s_j^{(1)}=1$
is replicated into copies of its single slice, so for every physical index $p_j$
into $T_1''$ along axis $j$,
\[
  T_1''\!\left[\ldots, p_j, \ldots\right] = T_1^+\!\left[\ldots, 0, \ldots\right],
\]
because the source index is $p_j \bmod s_j^{(1)} = 0$.
Reading off the effective tensors, $\mathsf{eff}(T_1'', m_e)$ accesses axis $j$ at
physical index $\lfloor j_j/\beta_j^{(2)}\rfloor$, while
$\mathsf{eff}(T_1^+, m_{e,1}^+)$ accesses it at $\lfloor j_j/\beta_j^{(1)}\rfloor = 0$
(since $j_j < s_j^{(1)}\beta_j^{(1)} = \beta_j^{(1)}$). Both therefore read
$T_1^+[\ldots,0,\ldots]$, so
\[
  \mathsf{eff}(T_1'', m_e)[\mathbf{a}]
  = \mathsf{eff}(T_1^+, m_{e,1}^+)[\mathbf{a}]
  \qquad(\mathbf{a} = j_1,\ldots,j_r),
\]
on the top block for every fixed tail index $\mathbf{J}'$. As in Step~1, $\irrepeat$
leaves the tail read unchanged, so the whole-stack tensors agree,
$\mathsf{eff}(T_1'', m_e{::}\irm_1') = \mathsf{eff}(T_1^+, m_{e,1}^+{::}\irm_1')$,
and taking the neuron slice:
\[
  \mathsf{eff}(T_1'', m_e{::}\irm_1')[i]
  = \mathsf{eff}(T_1^+, m_{e,1}^+{::}\irm_1')[i]
  = \sem{e_1}^{\mathrm{nl}}_{\rho_i}.
\]
The argument for $\irexpression[2]$ is symmetric.

\textit{Step 3: Recursive tail.}
The algorithm calls $\matchdims$ recursively on
$(\irexpression[1]'', \irexpression[2]'', \irm_1', \irm_2')$,
which have strictly smaller max-height.
By the inductive hypothesis, the output $(\irexpression[1]''', \irexpression[2]''', \irm')$
satisfies $\mathcal{L}(e_l, \irexpression[l]''', \irm')$ for $l=1,2$.
The final output is $(\irexpression[1]''', \irexpression[2]''', m_e :: \irm')$.
We claim $\mathcal{L}(e_l, \irexpression[l]''', m_e :: \irm')$, i.e.,
$\mathsf{eff}(T_{\mathrm{final}}, m_e{::}\irm')[i] = v_i$ where $T_{\mathrm{final}}$
is the evaluation of $\irexpression[l]'''$.

Steps~1 and~2 established
$\mathsf{eff}(T'', m_e{::}\irm_l')[i] = v_i$ for $\irexpression[l]''$.
The recursive call wraps $\irexpression[l]''$ into $\irexpression[l]'''$ by introducing
$\irrepeat$ and $\iradddim$ nodes.
Crucially, each such node carries a metadata element from the \emph{tail} stacks
$\irm_1'$ or $\irm_2'$, not from $m_e$: the recursive call only processes the
sub-stacks below $m_e$.
By the IR semantics, $\irrepeat(\cdot, m_{e}')$ and $\iradddim(\cdot, m_{e}')$
with $m_e' \neq m_e$ affect only the axes identified by $m_e'$ (the outer axes,
belonging to the tail block $\mathbf{J}'$), leaving the $m_e$-indexed top block
$\mathbf{a}$ intact.
Hence the top-block reading is unchanged while the tail is realigned from $\irm_l'$
to $\irm'$; by the inductive hypothesis this realignment preserves the whole-stack
effective tensor, so
$\mathsf{eff}(T_{\mathrm{final}}, m_e{::}\irm')[i] = \mathsf{eff}(T'', m_e{::}\irm_l')[i] = v_i$.
This gives $\mathcal{L}(e_l, \irexpression[l]''', m_e :: \irm')$.
\end{proof}

\subsection{\enspace Lemma: Stack Invariant for \map}

\begin{lemma}[Stack Invariant]
\label{lem:stackinv}
In the \textsc{Map} rule (Fig.~\ref{fig:map-rule}), $f$'s body $e_f$ is compiled by
the premise
$\Gamma,\sigma^+,F \vdash e_f \leadsto (i_{s_f},\, i_{e_f},\, m_e^{f}{::}m_{\mathit{out}})$,
starting from the stack $\sigma^+$ places on top: a stack of height $h$ whose top
element is $\irme[\mathit{map}]$. The derived stack $m_e^{f}{::}m_{\mathit{out}}$
also has height $h$, for some top element $m_e^{f}$ (not necessarily equal to
$\irme[\mathit{map}]$).
\end{lemma}

\begin{proof}
By induction on the nesting depth $d$ of \map constructs in $e_f$.

\smallskip
\textit{Base case} $(d=0$, no \map in $e_f)$:
the derivation of
$\Gamma,\sigma^+,F \vdash e_f \leadsto (\cdots)$ uses only the base-case, binary,
and \traverse\ rules. Inspecting each:
\begin{itemize}[topsep=1pt,itemsep=0pt]
\item \textit{Constants and variables} (\textsc{Const}, \textsc{Var}): these read
  $\sstore$ and return the stored metadata stack unchanged; they neither push nor pop.
\item \textit{Binary operations} (the \textsc{BinFloat}/\textsc{BinPoly} rules,
  \S\ref{sec:shapeanalysis}): both sub-expressions
  are derived recursively (the rule's premises; by the $d=0$ hypothesis, each yields
  a stack of height $h$),
  then \matchdims is applied.
  \matchdims operates element-wise on the two stacks and returns a combined stack of
  the same height $h$.
  Stack depth is therefore preserved: the returned stack has height $h$ and decomposes as
  some $m_e^{f}{::}m_{\mathit{out}}$.
  (The top element $m_e^{f}$ may differ from $\irme[\mathit{map}]$ if \matchdims
  expanded the shape — for example, extracting polynomial coefficients inside the body
  extends the rank of the map-level element, as shown in Fig.~\ref{fig:irfinal}.)
\item \textit{Unary operations and \dott}: process a single sub-expression;
  the stack is returned unchanged from that sub-expression's derivation.
\item \textit{\traverse} (the \textsc{Traverse}/\textsc{TravBody} rules): these emit
  an $\irwhile$ \emph{statement} node whose loop-carried expression uses fresh
  symbolic metadata variables populated from the DNN graph.
  Critically, they do not push or pop any element on the \emph{outer} map metadata
  stack: they read from $\sstore$ to obtain the loop input and write new IR variables
  for the loop body, but the map-dimension bookkeeping structure $\irm$ is separate
  and traverse never modifies it. (Internally, \textsc{TravBody} pushes its own map
  element and pops it via $\irreduce$, so the loop-carried stack is height-preserving
  by the same argument as the \map\ case below.)
  Hence \traverse\ changes neither the height nor the top element of the outer map
  metadata stack.
\item \textit{Function calls}: compiling a function call saves $\sstore$, rebinds
  formal parameters to IR arguments (updating stored IR nodes but not the map
  stack), recursively derives the function body, and restores $\sstore$.
  Under the $d=0$ assumption, the function body contains no nested \map, so
  the base-case argument applies to the body's derivation: it preserves the stack
  height.
  Restoring $\sstore$ does not alter the metadata stack.
  Hence the function call leaves the height unchanged (and, as with binary
  operations, may only refine the top element).
\end{itemize}
None of these cases changes the stack height, so the base case holds.

\smallskip
\textit{Inductive step} $(d \geq 1)$:
Assume the claim holds for all map bodies with nesting depth $< d$.
Let $\expr'.\map(g)$ be a \map construct encountered within $e_f$,
with $g$'s body having nesting depth $d' < d$.

When the derivation reaches $\expr'.\map(g)$, the \textsc{Map} rule applies again.
Its store $\sigma^+$ pushes a fresh map element
$\irme[\mathit{inner}] = \langle \typeneuron, \false, [P], [1] \rangle$
onto the stack, increasing the height by~1.
Since $g$'s body has nesting depth $d' < d$, the inductive hypothesis applies:
the premise deriving $g$'s body returns a stack of the same height $h+1$.
The $\irreduce$ premises then consume that top element $m_e^{g}$ to identify the
reduction axis, and the rule's conclusion returns the stack popped to height $h$.
The element below (previously $\irme[\mathit{map}]$, possibly refined by prior binary operations) is now the new top.

This argument applies to every nested \map encountered during the processing of $f$'s body.
Between nested \map calls, only base cases and binary operations are processed; by the
base-case argument these do not change the stack height.
Hence after processing the entire body of $f$, the stack has the original height $h$ and can be written
as $m_e^{f}{::}m_{\mathit{rest}}$ for some top element $m_e^{f}$.
\end{proof}

\subsection{\enspace Lemma: Map-Axis Materialization}

\begin{lemma}[Map-Axis Materialization]
\label{lem:mapaxis}
If the body of $\expr'.\map(f)$ compiles to $(i_{s_f},\, \irexpression[f],\,
m_e^{f}{::}\irm_{\mathit{out}})$ (height $\geq 2$, as required for the reduction of
Fig.~\ref{fig:map-rule}), then the top element $m_e^{f}$ consists solely of the map
axis: it has rank $1$, and that axis is \emph{stored}, with intrinsic size $P$ and
broadcast factor $1$.
\end{lemma}

\begin{proof}
We establish the three claims (rank $1$; intrinsic $P$; broadcast $1$) in turn.

\emph{Rank $1$.} Because $f$ returns a \polyexp, its value at each map position is a single
polyhedral number, whose constant component is a scalar. Hence the constant part of
$\irexpression[f]$ carries exactly one axis at the map level --- the map axis --- so
$m_e^{f}$ has rank $1$. (This is exactly the precondition that $\irreduce$ imposes: after
$\irextractpolycoeff$ appends the output coefficient axis, the top element must have rank
$\leq 2$, i.e.\ the pre-extraction top must have rank $\leq 1$; a body that violated this
would fail to compile.)

\emph{Intrinsic size $P$ (storedness).} The map arguments are introduced with the map axis
stored: \\ $\irmapcoeff$ and $\irmapneuron$ append a fresh top element
$\langle\cdot,\cdot,[P],[1]\rangle$, and for the neuron argument the intrinsic-to-broadcast
demotion is applied only to the \emph{lower} stack elements, never to this new axis. It
therefore suffices to show that no shape-analysis node ever moves the map axis from the
intrinsic shape into the broadcast factor. Inspecting the nodes: variable lookup and
pointwise/unary nodes leave the top element unchanged; coefficient/constant extraction and
$\iradddim$ only \emph{append} inner dimensions to the top element; $\irreduce$ removes the
top element entirely; and $\irrepeat$ moves a dimension from the broadcast factor
\emph{into} the intrinsic shape. The only node that merges two top elements is the $\lcm$
inside $\matchdims$, whose dimension-wise rule (Fig.~\ref{fig:alignshapes}) maps a
stored/broadcast pair only in the direction $1 \mapsto s$ (materialise the singleton via
$\irrepeat$) and leaves an equal-intrinsic pair unchanged; it never maps a stored size
$s>1$ to a singleton. Hence a dimension stored in some operand stays stored in the result.

It remains to check that the map axis is never introduced in broadcast form in the first
place. It originates only from $c_f$ and $n_f$, where it is stored, and it can propagate only
through $\matchdims$, which combines two \emph{equal-height} stacks (or a stack with a
constant, in which case the constant inherits the other operand's metadata verbatim). Its
rank-padding step ($\iradddim$) copies dimension~$0$ unchanged from the operand that already
carries the map axis and appends only \emph{inner} broadcast dimensions, so it never turns
the map axis into a broadcast dimension. Enclosing variables cannot supply a rival,
non-stored map axis: $\sstore^{+}$ extends $\sstore$, so they retain their pre-\map,
height-$1$ metadata (no map axis at all), and a nested \map or \traverse returns its result
at the height of its own input. Thus wherever the map axis appears it is stored, and in
particular $m_e^{f}$'s map axis has intrinsic size $P$.

\emph{Broadcast $1$.} The map axis has effective size $P$ (the number of coefficient pairs);
no shape-analysis node changes an axis's effective size --- $\matchdims$ only redistributes
a dimension between its intrinsic and broadcast components, preserving their product.
Since $\text{effective} = \text{intrinsic}\times\text{broadcast} = P$ and the intrinsic size
is $P$, the broadcast factor is $1$.
\end{proof}

\subsection{\enspace Proof of Theorem~\ref{thm:correctness}}

We prove the Correspondence Invariant~\eqref{eq:lifting} by structural induction on the
\cf expression $e$.
The induction is well-founded: \cf function definitions are non-recursive, so
function bodies are structurally smaller than the call sites.

For brevity, we give detailed proofs for the constructs that are the focus of
the main text (\S\ref{sec:shapeanalysis}): the base cases, binary operations on
\float and on \polyexp/\symexp, \map, and \traverse. Together these exercise
every mechanism that shape analysis relies on for correctness --- deferred
broadcast through \matchdims (Lemma~\ref{lem:alignshapes}), the metadata-stack
discipline (Lemma~\ref{lem:stackinv}), and the iterative loop lifting. The
remaining \cf constructs handled by the full implementation are discharged by
the same techniques; we summarize them at the end
(\textbf{Remaining cases}) and omit the routine details.

\medskip
\noindent\textbf{Case} $e = c$ (\textup{constant}):
Shape analysis emits $\irexpression = \irconst(c)$ with
$\irm = \langle \types_c, \true, [1], [N] \rangle$.
For any consistent $\varrho$:
$\varrho \vdash \irconst(c) \Downarrow^{\mathrm{ir}} c$ (scalar).
By~\eqref{eq:eff}: $\mathsf{eff}(c, \langle \types_c, \true, [1],[N]\rangle)[i] = c$
for all $i \in [N]$ (the broadcast factor $N$ replicates the scalar).
The CF semantics of a constant gives $(\Gamma,\fstore)\vdash c, \rho_i \Downarrow^{\mathrm{nl}} c$
for every $\rho_i$. Hence $\mathcal{L}(c, \irconst(c), \irm)$ holds. 

\medskip
\noindent\textbf{Case} $e = x$ (\textup{variable}):
Shape analysis returns $\sstore(x) = (\irexpression[x], \irm_x)$.
By the consistency condition~\eqref{eq:consistent} of the initial environment,
$\mathsf{eff}(\sem{\irexpression[x]}^{\mathrm{ir}}_\varrho, \irm_x)[i] = \rho_i(x)$
for all $i \in [N]$, which is exactly $\sem{x}^{\mathrm{nl}}_{\rho_i}$.
Hence $\mathcal{L}(x, \irexpression[x], \irm_x)$ holds. 

\medskip
\noindent\textbf{Case} $e = e_1 \oplus e_2$ (\textup{binary \float operation},
the \textsc{BinFloat} rule (\S\ref{sec:shapeanalysis})):
By the inductive hypothesis (IH):
$\mathcal{L}(e_1, \irexpression[1], \irm_1)$ and
$\mathcal{L}(e_2, \irexpression[2], \irm_2)$.

By Lemma~\ref{lem:alignshapes}, \matchdims produces
$(\irexpression[1]', \irexpression[2]', \irm)$ with
$\mathcal{L}(e_l, \irexpression[l]', \irm)$ for $l=1,2$.

The output expression is $\irexpression = \irbinary(\irexpression[1]', \irexpression[2]', \oplus)$.
For any consistent $\varrho$, let
$\varrho \vdash \irexpression[l]' \Downarrow^{\mathrm{ir}} T_l$.
Then $\varrho \vdash \irexpression \Downarrow^{\mathrm{ir}} T$ where
$T[k_1,\ldots,k_r] = T_1[k_1,\ldots,k_r] \oplus T_2[k_1,\ldots,k_r]$
(the IR binary node applies $\oplus$ element-wise).
Therefore:
\begin{align*}
  \mathsf{eff}(T, \irm)[i]
  &= \mathsf{eff}(T_1, \irm)[i]
     \;\oplus\;
     \mathsf{eff}(T_2, \irm)[i]
  &&\text{(\textsc{eff} commutes with pointwise $\oplus$ under shared $\irm$)} \\
  &= \sem{e_1}^{\mathrm{nl}}_{\rho_i}
     \;\oplus\;
     \sem{e_2}^{\mathrm{nl}}_{\rho_i}
  &&\text{(IH via Lemma~\ref{lem:alignshapes})} \\
  &= \sem{e_1 \oplus e_2}^{\mathrm{nl}}_{\rho_i}.
  &&\text{(CF semantics of $\oplus$)}
\end{align*}
Hence $\mathcal{L}(e_1\oplus e_2, \irexpression, \irm)$ holds. 

\medskip
\noindent\textbf{Case} $e = e_1 \oplus e_2$ (\textup{binary \polyexp or \symexp operation}):
The \textsc{BinPoly} rule decomposes into component-wise arithmetic on the coefficient
and constant parts via $\irextractpolycoeff$ / $\irextractpolyconst$, applies the \float
argument above to each component, and reassembles with $\ircombinepoly$.

Formally, let $v = (C, k)$ denote a polyhedral value with coefficient vector $C \in \mathbb{R}^P$
and constant $k \in \mathbb{R}$, and let $(e)_C, (e)_k$ denote the coefficient and
constant projections of a polyhedral expression's value.
By the IH applied to the extracted components,
$\mathcal{L}\bigl((e_1)_C, \irextractpolycoeff(\irexpression[1]), \irm_1\bigr)$ and
$\mathcal{L}\bigl((e_1)_k, \irextractpolyconst(\irexpression[1]), \irm_1\bigr)$
hold (and likewise for $e_2$).
Applying the binary-\float argument to each component gives:
\[
  \bigl(\mathsf{eff}(T, \irm)[i]\bigr)_C
  = \bigl(\sem{e_1}^{\mathrm{nl}}_{\rho_i}\bigr)_C
    \oplus
    \bigl(\sem{e_2}^{\mathrm{nl}}_{\rho_i}\bigr)_C,
\]
and similarly for the constant component.
Since $\ircombinepoly$ is the inverse of the extract operations and both are lossless,
the assembled result satisfies $\mathcal{L}(e_1\oplus e_2, \irexpression, \irm)$. 

\medskip
\noindent\textbf{Case} $e = \expr'.\map(f)$ (\textup{Fig.~\ref{fig:map-rule}}):

\noindent\textit{Setup.}
By the IH, $\mathcal{L}(\expr', \irexpression[1], \irm_1)$, i.e., for all $i \in [N]$:
\[
  \mathsf{eff}(\sem{\irexpression[1]}^{\mathrm{ir}}_\varrho, \irm_1)[i]
  = \sem{\expr'}^{\mathrm{nl}}_{\rho_i}
  \;=:\; (N_i, C_i, k_i),
\]
where $N_i \in ([N'])^P$ is the neuron list, $C_i \in \mathbb{R}^P$ the coefficient vector,
and $k_i \in \mathbb{R}$ the constant, of the polyhedral expression at neuron $i$.

\noindent\textit{Step 1: Extract components.}
Write $\irm_1^{\mathsf{bc}}$ for $\irm_1$ with the intrinsic $P$-dimension
of $\mathrm{top}(\irm_1)$ moved to its broadcast factor (reflecting that the neuron list
is shared across coefficient slices).
The following hold by the semantics of $\irmapneuron$, $\irmapcoeff$, and
$\irextractpolyconst$:
\begin{align}
  \mathsf{eff}(\sem{\irexpression[N]}^{\mathrm{ir}}_\varrho,\, \irme[\mathit{map}]{::}\irm_1^{\mathsf{bc}})[i,j]
  &= N_i[j],
  \quad i\in[N],\, j\in[P], \label{eq:mN}\\
  \mathsf{eff}(\sem{\irexpression[C]}^{\mathrm{ir}}_\varrho,\, \irme[\mathit{map}]{::}\irm_1)[i,j]
  &= C_i[j],
  \quad i\in[N],\, j\in[P], \label{eq:mC}\\
  \mathsf{eff}(\sem{\irexpression[k]}^{\mathrm{ir}}_\varrho,\, \irm_k)[i]
  &= k_i,
  \quad i\in[N], \label{eq:mk}
\end{align}
where $\irme[\mathit{map}] = \langle \typeneuron,\false,[P],[1]\rangle$ is the map
dimension element, and $[i,j]$ indexes first into the $N$-dimension (from the
remaining stack below $\irme[\mathit{map}]$) then into the $P$-dimension.
Equation~\eqref{eq:mN} holds because $\irmapneuron$ extracts the neuron list for
each polyhedral component, with the $N$-dimension broadcast (a single stored copy).
Equation~\eqref{eq:mC} holds because $\irmapcoeff$ extracts the coefficient for
each $(i,j)$ pair.
Equation~\eqref{eq:mk} holds because $\irextractpolyconst$ extracts the constant
scalar, independent of $j$.

\noindent\textit{Step 2: Apply $f$ pointwise.}
Define enriched environments:
$\varrho^+ = \varrho[\irexpression[N] \mapsto \varrho(\irexpression[N]),\;
                      \irexpression[C] \mapsto \varrho(\irexpression[C])]$,
and $\sstore^+ = \sstore[n_{\arg} \mapsto (\irexpression[N], \irme[\mathit{map}]{::}\irm_1^{\mathsf{bc}}),\;
                          c_{\arg} \mapsto (\irexpression[C], \irme[\mathit{map}]{::}\irm_1)]$.
By~\eqref{eq:mN} and~\eqref{eq:mC}, $\varrho^+$ is consistent with $\sstore^+$
for the two-index family $\{(\rho^{(i,j)})\}_{i\in[N],j\in[P]}$ of per-$(i,j)$ CF
environments, where $\rho^{(i,j)}$ extends $\rho_i$ with $n_{\arg} \mapsto N_i[j]$
and $c_{\arg} \mapsto C_i[j]$.

Since $f$'s body is structurally smaller than $\expr'.\map(f)$, the IH applies
to $f$'s body $\expr_f$:
\[
  \Gamma, \sstore^+, \fstore \vdash \expr_f \leadsto (i_{s_f},\, \irexpression[f],\,
  m_e^{f}{::}\irm_{\mathit{out}}),
\]
where, by Lemma~\ref{lem:stackinv}, the returned stack has height $2$ with some top
element $m_e^{f}$.
By Lemma~\ref{lem:mapaxis}, the top element $m_e^{f}$ has rank $1$ and is the map axis,
stored with intrinsic size $P$ and broadcast $1$ --- exactly as $\irme[\mathit{map}]$; the
two agree on this axis and differ only in the type field ($\irme[\mathit{map}]$ is a
\typeneuron, whereas $m_e^{f}$ carries the body's result typing). This agreement on the map
axis is what lets Step~3 identify the reduction axis.
The Correspondence Invariant (extended to height-$2$ stacks via~\eqref{eq:effstack})
gives, for all $i \in [N]$, $j \in [P]$:
\begin{equation}
\label{eq:fij}
  \mathsf{eff}(\varrho^+ \vdash \irexpression[f], m_e^{f}{::}\irm_{\mathit{out}})[i,j]
  = f(N_i[j], C_i[j]) \;=:\; (\gamma_{ij},\, \delta_{ij}) \in \mathbb{R}^{P'}\times\mathbb{R},
\end{equation}
where $\gamma_{ij}$ is the coefficient vector (of the output size $P'$) and $\delta_{ij}$
the constant of the polyhedral value $f(N_i[j], C_i[j])$, and
$(\Gamma,\fstore)\vdash \expr_f, \rho^{(i,j)} \Downarrow^{\mathrm{nl}} f(N_i[j], C_i[j])$
by the IH on $\expr_f$.

\noindent\textit{Step 3: Reduce over the map dimension.}
The $\irreduce$ premises of the \textsc{Map} rule reduce the coefficient and constant
parts of $\irexpression[f]$ over the map axis. Extraction treats the two parts differently, so we track the tensor axes
explicitly.

\emph{Coefficient part.}
$\irexpression[f].C = \irextractpolycoeff(\irexpression[f])$ retypes the top element to
\float and \emph{appends} the output coefficient axis of size $P'$ as its innermost
dimension. Writing $m_e^{f,C}$ for $m_e^{f}$ with type \float and intrinsic shape
$[P, P']$ (broadcast $[1,1]$), the extracted tensor satisfies, for $i\in[N]$, $j\in[P]$,
$j'\in[P']$:
\begin{equation}
\label{eq:extractC}
  \mathsf{eff}\bigl(\sem{\irexpression[f].C}^{\mathrm{ir}}_{\varrho^+},\;
    m_e^{f,C}{::}\irm_{\mathit{out}}\bigr)[i,j,j']
  = \gamma_{ij}[j'].
\end{equation}
$\irreduce$ then reduces over tensor axis $r$, where $r$ is the total number of tensor
axes contributed by the stack \emph{below} the top element; this is exactly the flat
position of the map axis $j$, whereas the appended coefficient axis $j'$ sits at position
$r{+}1$ and is left intact. By Lemma~\ref{lem:mapaxis} the map axis is stored with
intrinsic size $P$, so this reduction sums $P$ genuine per-pair contributions (rather than a
single broadcast copy). The surviving $j'$ axis descends into the (now top) element,
which is retyped \polyexp; write $m_C'$ for the resulting whole stack
($\irm_{\mathit{out}}$ with its top element extended by the $P'$ axis).
Denoting the result $T_{\mathit{pc}}$, by~\eqref{eq:extractC}:
\begin{equation}
\label{eq:reduce}
  \mathsf{eff}(T_{\mathit{pc}}, m_C')[i,j']
  = \sum_{j=0}^{P-1} \gamma_{ij}[j']
  \qquad\text{i.e.}\qquad
  \mathsf{eff}(T_{\mathit{pc}}, m_C')[i]
  = \sum_{j=0}^{P-1} \bigl(f(N_i[j], C_i[j])\bigr)_C
  \in \mathbb{R}^{P'}.
\end{equation}
It is essential that the reduction targets the map axis $j$ and \emph{not} the
coefficient axis $j'$: summing $j'$ would collapse the output polyhedral expression
into a scalar rather than accumulating the $P$ per-pair contributions.

\emph{Constant part.}
$\irexpression[f].k = \irextractpolyconst(\irexpression[f])$ only retypes the top element
to \float, leaving its rank-$1$ shape $[P]$ unchanged, so
$\mathsf{eff}(\sem{\irexpression[f].k}^{\mathrm{ir}}_{\varrho^+}, m_e^{f}{::}\irm_{\mathit{out}})[i,j] = \delta_{ij}$.
Here $\irreduce$ sums the sole map axis, giving a scalar per neuron:
$\mathsf{eff}(T_{\mathit{pz}}, \irm_{\mathit{out}})[i]
= \sum_{j=0}^{P-1}\delta_{ij} = \bigl(\sum_j f(N_i[j],C_i[j])\bigr)_k$.
The original constant $k_i$ of $\expr'$ (Eq.~\eqref{eq:mk}) is then aligned with
$\matchdims$ (Lemma~\ref{lem:alignshapes}) and added (step~3 of Fig.~\ref{fig:map-rule}),
yielding $\mathsf{eff}(T_{\mathit{pz}}', \irm_{\mathit{out}})[i]
= \bigl(\sum_j f(N_i[j],C_i[j])\bigr)_k + k_i$.

\noindent\textit{Conclusion for the \map case.}
$\ircombinepoly$ assembles the reduced coefficient tensor $T_{\mathit{pc}}$ and constant
tensor $T_{\mathit{pz}}'$ into a single \polyexp with metadata stack $\irm'$ (its top
element is $m_C'$ retyped \polyexp). By~\eqref{eq:reduce} and the constant computation above:
\begin{align}
\mathsf{eff}\!\left(\varrho^+ \vdash
  \ircombinepoly(\irexpression[\mathit{pc}], \irexpression[\mathit{pz}]''),\,
  \irm'\right)[i]
&= \Bigl(\underbrace{\textstyle\sum_{j} \bigl(f(N_i[j], C_i[j])\bigr)_C}_{\text{coefficient}},\;
          \underbrace{\textstyle\sum_{j} \bigl(f(N_i[j], C_i[j])\bigr)_k + k_i}_{\text{constant}}\Bigr) \\
&= \sum_{j=0}^{P-1} f(N_i[j], C_i[j]) \;+\; k_i .
\end{align}
where the final sum is polyhedral addition and $k_i$ is the original constant of $\expr'$.

\emph{Well-formedness condition (linearity of the map function).}
The reduction above sums $f$ over \emph{all} $P$ lanes, whereas CF's \map\
semantics applies $f$ only to the genuine $(\text{neuron},\text{coefficient})$
pairs, evaluating any zero-coefficient term to $0$~\cite{constraintflow}. Since the
tensor representation carries a full width-$P$ coefficient axis --- most of whose
lanes have coefficient $0$ --- the two coincide exactly when those zero-coefficient
lanes contribute nothing, i.e.\ when $f(n,0)=0$. We adopt this as a
\emph{well-formedness condition} on the map function: it is linear in its
coefficient argument, of the form $f(n,c)=c\cdot g(n)$ (optionally branching on
$\operatorname{sign}(c)$, as in DeepPoly's
$c\ge 0\,?\,c\cdot n[\texttt{L}]:c\cdot n[\texttt{U}]$). This holds for every \map\
(and \traverse) use in DNN certification. Under it --- and using that CF's \map\
carries the polyhedral constant unchanged --- summing over all $P$ lanes equals
summing over the non-zero pairs, and the right-hand side equals
$\sem{\expr'.\map(f)}^{\mathrm{nl}}_{\rho_i}$.
Hence $\mathcal{L}(\expr'.\map(f), \ircombinepoly(\irexpression[\mathit{pc}],
\irexpression[\mathit{pz}]''), \irm')$ holds.

\medskip
\noindent\textbf{Case} $e.\traverse(f, \mathit{dir}, \mathit{priority}, \mathit{stop})$:

\noindent\textit{CF semantics of \traverse.}
Let $(N_i^\ell, c_i^\ell, k_i^\ell)$ denote the PolyExp value at neuron $i$ after
$\ell$ iterations.
Define $e_0 = e$ (the traverse argument) and, for $\ell \geq 0$:
\begin{equation}
\label{eq:trav-step}
  e_{\ell+1,i}
  \;\triangleq\;
  \underbrace{\sum_{j:\;\mathrm{stop}(N_i^\ell[j])} c_{i,j}^\ell \cdot N_i^\ell[j]
    + k_i^\ell}_{\text{stopped part}}
  \;+\;
  \underbrace{\sum_{j:\;\neg\mathrm{stop}(N_i^\ell[j])} f\!\bigl(N_i^\ell[j],\, c_{i,j}^\ell\bigr)}_{\text{active part after }f},
\end{equation}
where the stopped part collects contributions from neurons that already satisfy
$\mathit{stop}$, and $f$ is applied to each active (non-stopped) neuron.
The loop terminates when $\neg\mathrm{stop}(N_i^\ell[j])$ holds for no $j$
(all neurons stopped).

\noindent\textit{Loop invariant.}
The compiler emits an $\irwhile$ loop with loop-carried expression
$\irexpression[\ell]$ and metadata stack $\irm_\ell$.
We prove by induction on $\ell$:
\[
  \mathcal{I}_\ell:\quad \mathcal{L}(e_\ell,\; \irexpression[\ell],\; \irm_\ell).
\]

\textit{Base case} $(\ell=0)$:
$\irexpression[0]$ is the IR expression for the traverse argument $e$.
By the outer structural IH on $e$, $\mathcal{L}(e, \irexpression[0], \irm_0)$,
i.e., $\mathcal{I}_0$ holds.

\textit{Inductive step} (assume $\mathcal{I}_\ell$):
Each loop iteration executes four sub-steps.

\smallskip
\textit{Sub-step 1: Partition.}
The compiler emits
$\irexpression[\mathrm{stop}] = \textsc{IrGetPolyexpStop}(\irexpression[\ell], \mathit{stop})$ 
and \\
$\irexpression[\mathrm{act}] = \textsc{IrGetPolyexpNotStop}(\irexpression[\ell], \mathit{stop})$. \\
By the semantics of these IR nodes, for all $i \in [N]$:
\begin{align}
  \mathsf{eff}(\sem{\irexpression[\mathrm{stop}]}^{\mathrm{ir}}_\varrho,
    \irm_\ell)[i]
  &= \bigl(N_i^\ell[j]_{j:\mathrm{stop}},\;
            c_{i,\cdot}^\ell\!\!\restriction_{\mathrm{stop}},\; k_i^\ell\bigr),
  \label{eq:trav-stop}\\
  \mathsf{eff}(\sem{\irexpression[\mathrm{act}]}^{\mathrm{ir}}_\varrho,
    \irm_\ell)[i]
  &= \bigl(N_i^\ell[j]_{j:\neg\mathrm{stop}},\;
            c_{i,\cdot}^\ell\!\!\restriction_{\neg\mathrm{stop}},\; 0\bigr).
  \label{eq:trav-act}
\end{align}
That is, the stop-node retains the original constant and all stopped-neuron
coefficients; the active-node takes the complementary coefficients with constant
$0$.
In particular $\mathcal{L}(e_\ell^{\mathrm{act}}, \irexpression[\mathrm{act}],
\irm_\ell)$ holds, where $e_\ell^{\mathrm{act}}$ is the active-neuron sub-expression.

\smallskip
\textit{Sub-step 2: Apply $f$ to active neurons via map.}
The \textsc{TravBody} rule (Appendix~\ref{app:formalism}) applies the same map
machinery as the \textsc{Map} rule to $\irexpression[\mathrm{act}]$ with function $f$.
By $\mathcal{I}_\ell$ and~\eqref{eq:trav-act}, the precondition of the \map case
(proved above) is satisfied with $\irexpression[\mathrm{act}]$ in place of $\irexpression[1]$.
The compiled active part maps $f$ over all $P$ lanes; by~\eqref{eq:trav-act} the
stopped lanes carry coefficient $0$, so it evaluates
$\sum_{j\in[P]} f(N_i^\ell[j], c_{i,j}^{\mathrm{act}})$, where
$c_{i,j}^{\mathrm{act}} = c_{i,j}^\ell$ on active lanes and $0$ on stopped lanes.
By the same well-formedness condition used in the \map\ case
(linearity of $f$, so $f(n,0)=0$), the stopped lanes contribute nothing, and this
equals the active-only sum $\sum_{j:\neg\mathrm{stop}} f(N_i^\ell[j], c_{i,j}^\ell)$
of~\eqref{eq:trav-step}.
The \map case gives output $\irexpression[\mathrm{new}]$ with metadata $\irm_\ell^{\mathrm{new}}$
satisfying, for all $i \in [N]$:
\begin{equation}
\label{eq:trav-new}
  \mathsf{eff}(\sem{\irexpression[\mathrm{new}]}^{\mathrm{ir}}_\varrho,
    \irm_\ell^{\mathrm{new}})[i]
  = \sum_{j:\;\neg\mathrm{stop}(N_i^\ell[j])} f\!\bigl(N_i^\ell[j],\; c_{i,j}^\ell\bigr).
\end{equation}

\smallskip
\textit{Sub-step 3: Accumulate.}
The compiler adds the stopped and new parts:
using $\matchdims$ to align shapes (Lemma~\ref{lem:alignshapes}) followed by
$\irbinary(\cdot,\cdot,+)$ on the coefficient and constant components, then
$\ircombinepoly$.
Applying Lemma~\ref{lem:alignshapes} and the binary-float case to each component,
and combining~\eqref{eq:trav-stop} with~\eqref{eq:trav-new}:
\begin{align*}
  \mathsf{eff}(\sem{\irexpression[\ell{+}1]}^{\mathrm{ir}}_\varrho,
    \irm_{\ell+1})[i]
  &= \bigl(\text{stopped part for }i\bigr)
   + \sum_{j:\;\neg\mathrm{stop}} f\!\bigl(N_i^\ell[j],\; c_{i,j}^\ell\bigr)
  = e_{\ell+1,i},
\end{align*}
where the last equality is exactly~\eqref{eq:trav-step}.
Hence $\mathcal{I}_{\ell+1}$ holds.

\textit{Symbolic shape variables.}
$\irm_\ell$ uses symbolic shape variables (e.g., for the growing polySize) populated
at runtime from the DNN graph (Section~\ref{sec:shapeanalysis-traverse}).
Since $\mathcal{L}$ is quantified over all consistent environments — which include
all valid shape instantiations — $\mathcal{I}_\ell$ holds for every concrete DNN.

\textit{Termination.}
The DNN is acyclic. In the direction of $\mathit{dir}$, each active neuron
$N_i^\ell[j]$ at network depth $d_j$ is replaced by neurons strictly closer
to the input ($d < d_j$ in forward traversal, $d > d_j$ in backward).
The maximum depth of any active neuron strictly decreases at each iteration.
Since the DNN has finitely many layers, the loop terminates.

\textit{Loop exit.}
At exit iteration $\ell^*$, all neurons in $\irexpression[\ell^*]$ satisfy
$\mathit{stop}$, so the active part is empty and $e_{\ell^*} = \sem{e.\traverse(\ldots)}^{\mathrm{nl}}$
is the fixed point reached by the neuron-level semantics.
By $\mathcal{I}_{\ell^*}$, the loop-carried IR output $\irexpression[\ell^*]$ with
metadata $\irm_{\ell^*}$ satisfies
$\mathsf{eff}(\sem{\irexpression[\ell^*]}^{\mathrm{ir}}_\varrho, \irm_{\ell^*})[i]
= e_{\ell^*,i} = \sem{e.\traverse(\ldots)}^{\mathrm{nl}}_{\rho_i}$
for all $i \in [N]$.
Hence $\mathcal{L}(e.\traverse(\ldots), \irexpression[\ell^*], \irm_{\ell^*})$ holds. 

\medskip
\noindent\textbf{Remaining cases.}
The constructs not detailed above are handled by the shape analysis with the
same machinery, and their cases follow the same pattern; we omit the routine
details. Each falls into one of the three templates already established:
\begin{itemize}[topsep=2pt,itemsep=1pt]
\item \emph{Base-case-like} (\dott on a \typeneuron, element access $\iraccess$,
  type conversions $\irneuronpoly$, $\irconstpoly,\irnoisesym$, unary operations):
  these either read $\sstore$ and return the stored metadata unchanged, or wrap a
  single sub-expression in a node that acts pointwise on the effective tensor.
  The correspondence invariant is preserved exactly as in the constant/variable and
  unary arguments, since $\mathsf{eff}$ commutes with any pointwise IR node.
\item \emph{Component-wise} (\dott on a \polyexp/\symexp, ternary/conditional):
  the operand is split with $\irextractpolycoeff$/$\irextractpolyconst$ (or the
  \textsf{Sym} variants), the operation is applied to each component, and the
  result is reassembled with $\ircombinepoly$/$\ircombinesym$ --- identical in
  structure to the binary \polyexp/\symexp case.
\item \emph{Map-style reduce} (\map on a \symexp): identical to the \polyexp
  \map case (Fig.~\ref{fig:map-rule}), with $\irmapnoise$ and
  $\irextractsymcoeff$/$\irextractsymconst$ replacing their \polyexp
  counterparts and the reduction taken over the zonotope noise axis instead of
  the neuron axis. Lemma~\ref{lem:stackinv} applies verbatim.
\end{itemize}
Function calls additionally save and restore $\sstore$ around the body's compilation;
since this introduces no new tensor axis and $\sstore$ is non-recursive, the
invariant carries through the body by the induction hypothesis. In each case the
correspondence invariant is preserved by the same arguments used above.

\medskip
\noindent\textbf{Conclusion.}
By structural induction, $\mathcal{L}(e, \irexpression, \irm)$ holds for all
well-typed \cf expressions $e$ processed by the shape analysis pass.
Instantiating at the top-level expression of $P$:
\[
  \mathsf{eff}(\sem{\hat{P}}^{\mathrm{ir}}_{\varrho(\mathbf{s})}, \irm_P)[i]
  = \sem{P}^{\mathrm{nl}}_{\rho_i(\mathbf{s})} = v_i
  \quad \forall\, i \in [N],
\]
which is precisely the statement of Theorem~\ref{thm:correctness}.
\qed

\section{Correctness Proofs: Rewrite Rules}
\label{app:proof-rewriterules}
\subsection{\enspace Proof of Theorem~\ref{thm:rewrites}: R1 (Contraction-to-MatMul)}

\begin{proof}[Proof of R1]
The rewrite rule collapses \\
$\irreduce(\irbinary(\irrepeat(i_A,\,\mathit{last}),\,\irrepeat(i_B,\,\mathit{first}),\,\times),\,+)$
to $\irinner(i_A,i_B)$.

Let $i_A$ evaluate to $A\in\mathbb{R}^{B\times m\times k}$ and $i_B$ evaluate
to $B\in\mathbb{R}^{B\times k\times n}$ under some IR environment.

\medskip
\noindent\textbf{Step 1: Repeat semantics.}
$\irrepeat(i_A,\mathit{last})$ inserts a new axis of size $n$ as the last
dimension, broadcast-replicating $A$ along it:
\[
  A'[b,i,\ell,j] = A[b,i,\ell]
  \quad\forall\; b\in[B],\;i\in[m],\;\ell\in[k],\;j\in[n].
\]
$\irrepeat(i_B,\mathit{first})$ inserts a new axis of size $m$ as the second
dimension, broadcast-replicating $B$:
\[
  B'[b,i,\ell,j] = B[b,\ell,j]
  \quad\forall\; b,i,\ell,j.
\]

\noindent\textbf{Step 2: Element-wise product.}
$\irbinary(A',B',\times)$ computes
\[
  C[b,i,\ell,j]
  = A'[b,i,\ell,j]\times B'[b,i,\ell,j]
  = A[b,i,\ell]\times B[b,\ell,j].
\]

\noindent\textbf{Step 3: Reduction.}
$\irreduce(C,+)$ sums over the $\ell$-axis (dimension 2):
\[
  D[b,i,j]
  = \sum_{\ell=0}^{k-1} C[b,i,\ell,j]
  = \sum_{\ell=0}^{k-1} A[b,i,\ell]\times B[b,\ell,j].
\]

\noindent\textbf{Step 4: Inner product.}
By definition, $\irinner(i_A,i_B)$ evaluates to the batched matrix product
$A\cdot B$:
\[
  (A\cdot B)[b,i,j]
  = \sum_{\ell=0}^{k-1} A[b,i,\ell]\times B[b,\ell,j].
\]

\noindent\textbf{Conclusion.}
$D[b,i,j]=(A\cdot B)[b,i,j]$ for all $b,i,j$, so both sides of R1 evaluate
to the same tensor. 

The precondition checked by the implementation
(\texttt{rewrite\_expr\_5} in \texttt{rewrite.py}) is that one operand is
repeated along the last axis and the other along the first; together these
ensure $A'$ and $B'$ have the same shape and the reduction is over the shared
inner axis~$\ell$.
\end{proof}

\subsection{\enspace Proof of Theorem~\ref{thm:rewrites}: R2 (Ternary-to-Multiply)}

\begin{proof}[Proof of R2]
The rewrite replaces $(\irexpression[b]\;?\;\irexpression[2]\;:\;\irexpression[3])$
with $\irexpression[b]\times\irexpression[2] + (1-\irexpression[b])\times\irexpression[3]$,
subject to precondition $P_t$ (operand $\irexpression[b]$ has Boolean type).

Let $b,x,y$ be the values of $\irexpression[b],\irexpression[2],\irexpression[3]$
respectively.  The Boolean typing guarantees $b\in\{0,1\}$ element-wise.

\medskip
\noindent\textbf{Case $b=1$ (true).}
\[
  b\cdot x + (1-b)\cdot y = 1\cdot x + 0\cdot y = x + 0 = x.
\]
The ternary $(1\;?\;x:y)=x$. 

\medskip
\noindent\textbf{Case $b=0$ (false).}
\[
  b\cdot x + (1-b)\cdot y = 0\cdot x + 1\cdot y = 0 + y = y.
\]
The ternary $(0\;?\;x:y)=y$. 

\medskip
\noindent\emph{Precondition.}
The rewritten form $b\cdot x + (1-b)\cdot y$ evaluates \emph{both} branches
eagerly, whereas the ternary evaluates only the taken branch. This is sound only
when both branches are total. The implementation (\texttt{rewrite\_expr\_1})
therefore fires only on \emph{division-free} ternaries, guarded by
\texttt{not is\_div}: a branch containing a division may be undefined on the
untaken side, so such ternaries are left intact (and evaluated lazily). With no
partial operation in either branch, $x$ and $y$ are both always well-defined, and
the two forms agree for every $b\in\{0,1\}$. 

The \texttt{IrConvertBoolToFloat} node inserted by the implementation converts
the Boolean tensor $b$ to its $\{0.0, 1.0\}$ floating-point representation
before the arithmetic, which is exactly the $b\in\{0,1\}$ assumption used
above.
\end{proof}

\subsection{\enspace Proof of Theorem~\ref{thm:rewrites}: R3 and R4 (Clamp Rewrites)}

\begin{proof}[Proof of R3]
R3 rewrites $(\irexpression[1]\geq 0)\times\irexpression[1]$ to
$\irclamp(\irexpression[1],\mathit{min}{=}0)$. \\
Let $x$ be the value of $\irexpression[1]$.
$\irclamp(x,\mathit{min}{=}0)$ is defined as $\max(x,0)$.

\medskip
\noindent\textbf{Case $x\geq 0$.}
The comparison $(x\geq 0)$ evaluates to $1$ (true as a Boolean, i.e., $1.0$
as a float after conversion).  Hence:
\[
  (x\geq 0)\times x = 1\times x = x = \max(x,0). \checkmark
\]

\medskip
\noindent\textbf{Case $x<0$.}
The comparison $(x\geq 0)$ evaluates to $0$.  Hence:
\[
  (x\geq 0)\times x = 0\times x = 0 = \max(x,0). \checkmark
\]
In both cases $(x\geq 0)\times x = \max(x,0) = \irclamp(x,\mathit{min}{=}0)$.
\end{proof}

\begin{proof}[Proof of R4]
R4 rewrites $(\irexpression[1]\leq 0)\times\irexpression[1]$ to
$\irclamp(\irexpression[1],\mathit{max}{=}0)$.
Let $x$ be the value of $\irexpression[1]$.
$\irclamp(x,\mathit{max}{=}0)=\min(x,0)$.

\medskip
\noindent\textbf{Case $x\leq 0$.}
$(x\leq 0)=1$, so $(x\leq 0)\times x = x = \min(x,0)$. 

\medskip
\noindent\textbf{Case $x>0$.}
$(x\leq 0)=0$, so $(x\leq 0)\times x = 0 = \min(x,0)$. 

The implementation applies these rules in two forms, both requiring the
comparison to be against~$0$ (checked by \texttt{get\_const(cond\_rhs) == 0}).
The bare product
$\mathrm{boolToFloat}(\irexpression[1]\geq 0)\times\irexpression[1]$ is collapsed
to $\irclamp(\irexpression[1],\mathit{min}{=}0)$ by \texttt{rewrite\_expr\_clamp},
which is exactly the $(x\geq 0)\times x = \max(x,0)$ case above.
The two-branch ReLU-relaxation ternary
$(\irexpression[1]\geq 0)\;?\;\irexpression[1]\!\cdot a : \irexpression[1]\!\cdot b$,
whose branches scale the compared value by different coefficients, is collapsed by \\
\texttt{rewrite\_expr\_ternary\_to\_clamp} to
$\irclamp(\irexpression[1],\mathit{min}{=}0)\cdot a
 + \irclamp(\irexpression[1],\mathit{max}{=}0)\cdot b
 = \max(x,0)\cdot a + \min(x,0)\cdot b$.
By the case analysis above this reproduces the ternary for both signs of $x$
($x\geq 0$ gives $x\cdot a$; $x<0$ gives $x\cdot b$); the $\leq$ direction is
symmetric. Both forms fire only when the ternary is division-free
(\texttt{not is\_div}).
\end{proof}

\section{Correctness Proof: Code Generation}
\label{app:proof-codegen}

\begin{proof}[Proof of Theorem~\ref{thm:codegen}]
By induction on $i_s$. Code generation is syntax-directed and one-to-one (\S\ref{sec:codegeneration}): $\irassignment$, $\irseq$, $\irwhile$, and $\irite$ become Python assignment, sequencing, \texttt{while}, and \texttt{if}, whose semantics match the IR statement rules, and each expression node becomes a backend call that, by assumption, evaluates to the same tensor. For the \sparse backend, the assumption holds by Theorem~\ref{thm:sparse}; the Python and PyTorch code itself is not formally verified.
\end{proof}

\section{Additional \sparse Details and Correctness Proofs}
\label{app:sparse}

This appendix supplements \S\ref{sec:sparserepresentation} with the remaining
block-type illustrations, a worked example of a binary operation, the
matrix-multiplication operation, and full correctness proofs for
Theorem~\ref{thm:sparse}. The proofs are grounded in the actual \sparse
implementation (files \texttt{sparse\_tensor.py} and \texttt{sparse\_block.py}).
Full proofs for the rewrite rules (Theorem~\ref{thm:rewrites}) are in
Appendix~\ref{app:proof-rewriterules}.

\subsection{Additional Sparse Blocks}
\label{app:sparseblocks}
Besides the Stencil, Window, and Diagonal blocks presented in
\S\ref{sec:sparsedefs}, \sparse uses two further specialized block types, the
Broadcast and Uniform blocks, together with the Dense block that serves as a
fallback. Figure~\ref{fig:sparseblocks} shows the dense and \sparse forms of the
Broadcast and Uniform blocks.

\paragraph{Dense Block (\dblock).}
The core tensor $b$ equals the block itself, and $\varphi$ is the identity.
Dense blocks are the fallback for entries whose structure is not captured by
any of the specialized types. They arise, for example, in the coefficient
tensors of zonotope certifiers after multiple layers of backsubstitution.

\paragraph{Broadcast Block (\rblock).}
A Broadcast block models the situation where all rows (or slices along one axis) of
a sub-tensor are identical. This arises when a scalar or vector value is
broadcast across a dimension --- for example, the bias term in an affine layer,
which is the same for all inputs in a batch. The core tensor $b$ stores a single
copy, and $\Theta = \{\mathit{rep\_axis}, \mathit{rep\_count}\}$ records the
repetition.

\paragraph{Uniform Block (\cblock).}
The simplest block type: all entries in the block equal a single scalar
$b \in \mathbb{R}$, and $\Theta = \emptyset$. This arises after applying a
scalar operation (e.g., $\times 0$ or $+ c$) uniformly to a region. Uniform
blocks interact cleanly with the background value $d$, and are often merged into
$d$ directly.

\begin{figure}[h]
    \centering
    \begin{subfigure}{0.35\textwidth}
        \begin{minipage}{0.75\textwidth}
        \centering
        \resizebox{\linewidth}{!}{
\begin{tikzpicture}[x=0.5cm, y=0.5cm, every node/.style={font=\small}]
  \def\size{6}

  \def\rowColors{{"teal!20!white", "teal!36!white", "teal!52!white", "teal!68!white", "teal!84!white", "teal!100!white"}}

  \foreach \i in {0,...,5} {
    \pgfmathparse{\rowColors[\i]}
    \edef\col{\pgfmathresult}
    \foreach \j in {0,...,5} {
      \fill[\col] (\j,-\i) rectangle ++(1,-1);
      \draw[gray] (\j,-\i) rectangle ++(1,-1);
    }
  }

  \draw[line width=1pt] (0,0) rectangle (\size,-\size);

\end{tikzpicture}}
      \end{minipage}
      \begin{minipage}{0.15\textwidth}
        \centering
        \resizebox{\linewidth}{!}{
\begin{tikzpicture}[x=0.5cm, y=0.5cm, every node/.style={font=\small}]
  \def\size{6}
  \def\BX{0} 

  \def\rowColors{{"teal!20!white", "teal!36!white", "teal!52!white", "teal!68!white", "teal!84!white", "teal!100!white"}}

  \foreach \i in {0,...,5} {
    \pgfmathparse{\rowColors[\i]}
    \edef\col{\pgfmathresult}
    \fill[\col] (\BX,-\i) rectangle ++(1,-1);
    \draw[gray] (\BX,-\i) rectangle ++(1,-1);
  }

  \draw[line width=1pt] (\BX,0) rectangle (\BX+1,-\size);

\end{tikzpicture}}
      \end{minipage}
      \caption{Broadcast Block}
      \label{fig:repeatblock}
    \end{subfigure}
    \hfill
    \begin{subfigure}{0.35\textwidth}
        \begin{minipage}{0.75\textwidth}
        \centering
        \resizebox{\linewidth}{!}{
\begin{tikzpicture}[x=0.5cm, y=0.5cm, every node/.style={font=\small}]
  \def\size{6}

  \def\constColor{teal!70!white}

  \foreach \i in {0,...,5} {
    \foreach \j in {0,...,5} {
      \fill[\constColor] (\j,-\i) rectangle ++(1,-1);
      \draw[gray] (\j,-\i) rectangle ++(1,-1);
    }
  }

  \draw[line width=1pt] (0,0) rectangle (\size,-\size);

\end{tikzpicture}}
      \end{minipage}
      \begin{minipage}{0.15\textwidth}
        \centering
        \resizebox{\linewidth}{!}{
\begin{tikzpicture}[x=0.5cm, y=0.5cm, every node/.style={font=\small}]
  \def\size{6}
  \def\BX{0} 

  \def\constColor{teal!70!white}

  \fill[\constColor] (\BX, -2.5) rectangle ++(1,-1);
  \draw[gray]        (\BX, -2.5) rectangle ++(1,-1);
  \draw[line width=1pt] (\BX, -2.5) rectangle ++(1,-1);

\end{tikzpicture}}
      \end{minipage}
      \caption{Uniform Block}
      \label{fig:constblock}
    \end{subfigure}
  \caption{Dense and \sparse forms of the Broadcast (Fig.~\ref{fig:repeatblock})
    and Uniform (Fig.~\ref{fig:constblock}) blocks. In each pair the left is the
    dense block and the right is its \sparse representation.}
\label{fig:sparseblocks}
\end{figure}

\subsection{Example of a Binary Operation over \sparse Tensors}
\label{app:sparse-example}
Let $\mathcal{T}_1$ and $\mathcal{T}_2$ be two sparse tensors in \sparse format.
$\mathcal{T}_1$ has three blocks $B_1: \dblock$, $B_2: \dblock$, and
$B_3: \diagblock$ at start indices $s_1, s_2, s_3$, with background value $0$.
$\mathcal{T}_2$ has two blocks $B'_1: \cblock$ with constant $1$ and
$B'_2: \diagblock$ at start indices $s_1, s_2$, also with background value $0$.
Assume $B_1$ and $B'_1$ fully overlap, and $B_2$ and $B'_2$ fully overlap. We
compute $\mathcal{T}_3 = \mathcal{T}_1 \times \mathcal{T}_2$.
Since $B_3$ does not overlap any block in $\mathcal{T}_2$ and the background is
$0$ (the annihilator of multiplication), $B_3$ is multiplied by nothing and is
simply replaced by $0$.
Since $B'_1$ is a uniform block with value $1$ (the identity of
multiplication), the result of $B_1 \times B'_1$ is $B_1$.
Finally, $B_2 \times B'_2$ multiplies a dense block by a diagonal block; the
result is a diagonal block, computed by multiplying only the diagonal elements
of $B_2$ and $B'_2$ in $O(n)$ time rather than $O(n^2)$.

\subsection{Matrix Multiplication over \sparse Tensors}
\label{app:sparse-matmul}
Matrix multiplication follows the same principle as the element-wise binary
operations (\S\ref{sec:sparseops}), with overlap redefined to respect the
contraction. For $\mathcal{T}_3 = \mathcal{T}_1 \cdot \mathcal{T}_2$, each entry
is $\mathcal{T}_3[\mathbf{i}, \mathbf{k}] = \sum_j \mathcal{T}_1[\mathbf{i}, j]
\cdot \mathcal{T}_2[j, \mathbf{k}]$, so a block of $\mathcal{T}_1$ interacts with
a block of $\mathcal{T}_2$ whenever their inner index ranges are compatible for
multiplication rather than identical. We again take the transitive closure of
this relation, and within each group compute all valid pairwise products of the
participating core tensors and sum them into the corresponding output block. The
identity and annihilator shortcuts carry over unchanged: multiplying against a
background of $0$ annihilates a block and can be skipped, while adding a
background of $0$ leaves a block untouched.

\subsection{Element-Access Semantics of \sparse Tensors}

We first make the pointwise semantics of \sparse explicit, since the
theorem statements in \S\ref{sec:sparserepresentation} use $\mathcal{T}[\mathbf{i}]$
informally.

\begin{definition}[Element Access]
\label{def:elemaccess}
For a \sparse tensor $\mathcal{T}=(d,S,B,Z)$ and index $\mathbf{i}\in[Z]$:
\[
  \mathcal{T}[\mathbf{i}] \;:=\;
  \begin{cases}
    b_k\!\left[\varphi_{\Theta_k}^{-1}(\mathbf{i}-s_k)\right]
      & \text{if } \exists k:\;
        \mathbf{i}-s_k \in \mathrm{range}(\varphi_{\Theta_k}), \\
    d
      & \text{otherwise.}
  \end{cases}
\]
The case is unique: blocks are non-overlapping (Def.~\ref{def:sparsetensor}),
and each $\varphi_{\Theta_k}$ is injective (Def.~\ref{def:sparseblock}).
\end{definition}

\noindent
We also note that for a \diagblock with parameter $\delta$
(the pair of identical dimensions), the index mapping is:
\begin{equation}
\label{eq:diagmap}
  \varphi_\delta(j_0,\ldots,j_{\delta-1},j_{\delta+1},\ldots,j_{r-1})
  = (j_0,\ldots,j_{\delta-1},j_{\delta-1},j_{\delta+1},\ldots,j_{r-1}),
\end{equation}
i.e., dimension $\delta-1$ of the core is replicated at both positions $\delta-1$
and $\delta$ in the dense block.  Injectivity is immediate: if
$\varphi_\delta(\mathbf{j})=\varphi_\delta(\mathbf{j}')$ then $j_k=j_k'$ for
every component $k$.

\subsection{Proof of Theorem~\ref{thm:sparse} (Sparse Operation Correctness)}
\label{app:proof-sparseandrewrite}

\begin{proof}
By Def.~\ref{def:elemaccess}, $\mathrm{dense}(\mathcal{T})[\mathbf{i}]
=\mathcal{T}[\mathbf{i}]$, so the equalities
$\mathrm{dense}(\mathcal{T}_1\odot\mathcal{T}_2)
=\mathrm{dense}(\mathcal{T}_1)\odot\mathrm{dense}(\mathcal{T}_2)$
of Theorem~\ref{thm:sparse} are equivalent to the pointwise identities
established below.
We first prove $(\mathcal{T}_1\oplus\mathcal{T}_2)[\mathbf{i}]
=\mathcal{T}_1[\mathbf{i}]\oplus\mathcal{T}_2[\mathbf{i}]$
for every $\mathbf{i}\in[Z]$.

Let $\mathcal{T}_3=(d_3,S_3,B_3,Z)$ with $d_3=d_1\oplus d_2$ be the result.
Fix $\mathbf{i}$; there are four cases.

\medskip
\noindent\textbf{Case 1 (no block covers $\mathbf{i}$).}
$\mathcal{T}_l[\mathbf{i}]=d_l$ for $l=1,2$ by Def.~\ref{def:elemaccess}.
The algorithm produces no block in $\mathcal{T}_3$ covering~$\mathbf{i}$, so
$\mathcal{T}_3[\mathbf{i}]=d_3=d_1\oplus d_2
=\mathcal{T}_1[\mathbf{i}]\oplus\mathcal{T}_2[\mathbf{i}]$. 

\medskip
\noindent\textbf{Case 2 ($\mathbf{i}$ covered only by block $B_{k_1}\in\mathcal{T}_1$).}
$\mathcal{T}_1[\mathbf{i}]=b_{k_1}[\varphi_{\Theta_{k_1}}^{-1}(\mathbf{i}-s_{k_1})]$
and $\mathcal{T}_2[\mathbf{i}]=d_2$.
The \texttt{binary} method of \texttt{SparseTensor} applies $\oplus$ between
$B_{k_1}$ and the scalar $d_2$ entry-by-entry.  The result block
$B_{k_1}^{(3)}$ retains the same $\varphi_{\Theta_{k_1}}$ with core
$(b_{k_1}\oplus d_2)$:
\[
  \mathcal{T}_3[\mathbf{i}]
  = (b_{k_1}\oplus d_2)\!\left[\varphi_{\Theta_{k_1}}^{-1}(\mathbf{i}-s_{k_1})\right]
  = b_{k_1}\!\left[\varphi_{\Theta_{k_1}}^{-1}(\mathbf{i}-s_{k_1})\right]\oplus d_2
  = \mathcal{T}_1[\mathbf{i}]\oplus\mathcal{T}_2[\mathbf{i}]. 
\]

\noindent\emph{Identity shortcut.}
If $d_2=\mathbf{1}_\oplus$, then $b_{k_1}[\mathbf{j}]\oplus d_2
=b_{k_1}[\mathbf{j}]$ for every core index $\mathbf{j}$, so the implementation
copies $B_{k_1}$ unchanged to $\mathcal{T}_3$.  Correctness follows from Case~2
with $b_{k_1}\oplus\mathbf{1}_\oplus=b_{k_1}$.

\noindent\emph{Annihilator shortcut.}
If $d_2=\mathbf{0}_\oplus$, then for every $\mathbf{j}$,
$b_{k_1}[\mathbf{j}]\oplus\mathbf{0}_\oplus=\mathbf{0}_\oplus$.
Since $d_3=d_1\oplus\mathbf{0}_\oplus=\mathbf{0}_\oplus$, the result block
would be identically equal to the background $d_3$; the implementation correctly
omits it.  Every $\mathbf{i}$ covered only by $B_{k_1}$ then receives value
$d_3=\mathbf{0}_\oplus$ from the background, which equals
$\mathcal{T}_1[\mathbf{i}]\oplus\mathcal{T}_2[\mathbf{i}]
=b_{k_1}[\cdots]\oplus\mathbf{0}_\oplus=\mathbf{0}_\oplus$. 

\medskip
\noindent\textbf{Case 3 ($\mathbf{i}$ covered only by block $B_{k_2}\in\mathcal{T}_2$).}
Symmetric to Case~2. 

\medskip
\noindent\textbf{Case 4 ($\mathbf{i}$ covered by $B_{k_1}\in\mathcal{T}_1$ and
$B_{k_2}\in\mathcal{T}_2$).}
The \texttt{union\_tensors} routine computes the transitive closure of
overlapping block pairs.  Since $B_{k_1}$ and $B_{k_2}$ share index
$\mathbf{i}$, they are placed in the same overlap group.  Within the group, the
merge/split heuristic computes $v_1\oplus v_2$ at each position in the union,
where $v_l$ is the value from $B_{k_l}$ (or $d_l$ if the position falls
outside $B_{k_l}$).  At $\mathbf{i}$, both blocks contribute:
\[
  v_1 = b_{k_1}[\varphi_{\Theta_{k_1}}^{-1}(\mathbf{i}-s_{k_1})]
       =\mathcal{T}_1[\mathbf{i}],
  \qquad
  v_2 = b_{k_2}[\varphi_{\Theta_{k_2}}^{-1}(\mathbf{i}-s_{k_2})]
       =\mathcal{T}_2[\mathbf{i}],
\]
so the output block stores $v_1\oplus v_2
=\mathcal{T}_1[\mathbf{i}]\oplus\mathcal{T}_2[\mathbf{i}]$ at the appropriate
position, giving $\mathcal{T}_3[\mathbf{i}]
=\mathcal{T}_1[\mathbf{i}]\oplus\mathcal{T}_2[\mathbf{i}]$. 

\medskip
\noindent\textbf{Matmul case.}
We prove $(\mathcal{T}_1\cdot\mathcal{T}_2)[\mathbf{i},\mathbf{k}]
=\sum_j\mathcal{T}_1[\mathbf{i},j]\cdot\mathcal{T}_2[j,\mathbf{k}]$.
The block-based algorithm below is sound only when both operands have background
value $0$, because otherwise the indices left uncovered by the blocks carry a
non-zero value that contributes to the sum
($d_1\cdot\mathcal{T}_2[j,\mathbf{k}]$ or $\mathcal{T}_1[\mathbf{i},j]\cdot d_2$)
yet is never visited by a block-only enumeration. The runtime therefore
dispatches on the backgrounds. The left operand $\mathcal{T}_1$ always has
background $0$: it is a coefficient matrix, whose absent entries are $0$ (the
implementation asserts this). If the right operand $\mathcal{T}_2$ also has
background $0$, the block-based algorithm is used, and we prove it correct
below. If $\mathcal{T}_2$ has a non-zero background, the implementation instead
materializes both operands as dense blocks and performs a standard dense
multiplication, which computes the sum by definition; correctness is then
immediate. It thus suffices to prove the block-based algorithm assuming both
backgrounds are $0$.

With both backgrounds $0$, any index not covered by a
block in $\mathcal{T}_1$ contributes
$0\cdot\mathcal{T}_2[j,\mathbf{k}]=0$ to the sum, and symmetrically an uncovered
$\mathcal{T}_2$ index contributes $\mathcal{T}_1[\mathbf{i},j]\cdot 0=0$.  Hence
only block pairs contribute.

The algorithm iterates over all pairs $(B_{k_1}^{(1)},B_{k_2}^{(2)})$ whose
inner index ranges intersect (i.e., the column range of $B_{k_1}^{(1)}$
overlaps the row range of $B_{k_2}^{(2)}$).  For each such pair, it computes
the partial product $B_{k_1}^{(1)}\cdot B_{k_2}^{(2)}$ restricted to the
overlapping inner range.  The \texttt{find\_connected\_blocks} post-processing
step sums all partial products at each output position $(\mathbf{i},\mathbf{k})$.

Let $J(\mathbf{i},\mathbf{k})$ denote the set of $j$ covered by some
$B_{k_1}^{(1)}$ in row $\mathbf{i}$ and some $B_{k_2}^{(2)}$ in column
$\mathbf{k}$.  For $j\in J(\mathbf{i},\mathbf{k})$, the pair that covers $j$
is included in the enumeration and contributes
$\mathcal{T}_1[\mathbf{i},j]\cdot\mathcal{T}_2[j,\mathbf{k}]$.
For $j\notin J(\mathbf{i},\mathbf{k})$, at least one of $\mathcal{T}_l[\cdot,j]$
is $0$ (background), contributing $0$.  Hence:
\[
  (\mathcal{T}_1\cdot\mathcal{T}_2)[\mathbf{i},\mathbf{k}]
  = \sum_{j\in J(\mathbf{i},\mathbf{k})}
      \mathcal{T}_1[\mathbf{i},j]\cdot\mathcal{T}_2[j,\mathbf{k}]
  = \sum_j \mathcal{T}_1[\mathbf{i},j]\cdot\mathcal{T}_2[j,\mathbf{k}]. 
\]
This completes the proof of Theorem~\ref{thm:sparse}.
\end{proof}

\subsection{\diagblock{}+\dblock Conjunctive Binary Operation}

The following lemma grounds the \texttt{conjunctive\_binary} special case
from \texttt{SparseBlock.binary} in the implementation.

\begin{lemma}[\diagblock Conjunctive Binary]
\label{lem:diagbin}
Let $D$ be a \diagblock with diagonal-pair parameter $\delta$ and core $b$,
and let $A$ be a \dblock with core tensor $a$ (same block shape as $D$).
Let $\oplus$ be a binary operator whose annihilator is $0$ (the global
background value). Then:
\[
  (D \oplus A)[\mathbf{i}]
  = \begin{cases}
      b[\mathbf{j}]\oplus a[\varphi_\delta(\mathbf{j})]
        & \text{if } \mathbf{i} = \varphi_\delta(\mathbf{j})
          \text{ for some core index } \mathbf{j}, \\
      0
        & \text{otherwise}.
    \end{cases}
\]
Equivalently, $(D\oplus A)$ is a \diagblock with the same parameter $\delta$
and core $b' = b \oplus \mathit{diag}_\delta(a)$, where
$\mathit{diag}_\delta(a)[\mathbf{j}]
:= a[\varphi_\delta(\mathbf{j})]$ extracts the $\delta$-diagonal of $a$.
\end{lemma}

\begin{proof}
Fix any index $\mathbf{i}$ in the block's index range.

\medskip
\noindent\textbf{Off-diagonal ($\mathbf{i}\notin\mathrm{range}(\varphi_\delta)$).}
By Def.~\ref{def:elemaccess}, $D[\mathbf{i}]=0$ (the background, since
$\mathbf{i}$ is not in the range of $\varphi_\delta$).
Since $0=\mathbf{0}_\oplus$:
$D[\mathbf{i}]\oplus A[\mathbf{i}]=0\oplus A[\mathbf{i}]=0$.
The new block must store $0$ at $\mathbf{i}$; since the background is also $0$,
the position contributes no entry to the result block. 

\medskip
\noindent\textbf{Diagonal ($\mathbf{i}=\varphi_\delta(\mathbf{j})$ for unique $\mathbf{j}$).}
$D[\mathbf{i}]=b[\mathbf{j}]$ (by Def.~\ref{def:elemaccess} and injectivity of
$\varphi_\delta$) and $A[\mathbf{i}]=a[\mathbf{i}]=a[\varphi_\delta(\mathbf{j})]$.
The result is $b[\mathbf{j}]\oplus a[\varphi_\delta(\mathbf{j})]$, which equals
$b'[\mathbf{j}]=(b\oplus\mathit{diag}_\delta(a))[\mathbf{j}]$. 

\medskip
The implementation (\texttt{sparse\_block.py}, \texttt{conjunctive\_binary})
extracts the diagonal of $a$ via \texttt{get\_diagonal(sp\_block.block,\,$\delta$)}
and multiplies element-wise with $b$, matching the formula above.
\end{proof}

\clearpage
\section{Additional Evaluation and Results}
\label{app:eval-extra}

\subsection{Experimental Setup}
\label{app:setup}
\paragraph{Benchmark networks.}
Table~\ref{tab:allnets} lists the full details of all 32 DNNs used in the
evaluation, of which Table~\ref{tab:repnets} shows six representatives. The
networks span two datasets, three training regimes (standard, PGD, DiffAI), and
fully-connected, convolutional, and residual architectures.

\begin{table}[h]
    \centering
    \small
    \begin{tabular}{llllllll}
    \toprule
        Name & Dataset & Model & Training Method & Architecture & Layers & $\epsilon$ & Batch \\
    \midrule
     $N_{1}$  & MNIST   & 3x50      & Standard              & FCN           & 3  & 0.004  & 100\\
     $N_{2}$  & MNIST   & 3x100     & Standard              & FCN           & 3  & 0.004  & 100\\
     $N_{3}$  & MNIST   & 5x100     & DiffAI                & FCN           & 6  & 0.004  & 100\\
     $N_{4}$  & MNIST   & 6x100     & Standard              & FCN           & 6  & 0.004  & 100\\
     $N_{5}$  & MNIST   & 9x100     & Standard              & FCN           & 9  & 0.004  & 100\\
     $N_{6}$  & MNIST   & 6x200     & Standard              & FCN           & 6  & 0.004  & 100\\
     $N_{7}$  & MNIST   & 9x200     & Standard              & FCN           & 9  & 0.004  & 100\\
     $N_{8}$  & MNIST   & 6x500     & Standard              & FCN           & 6  & 0.004  & 100\\
     $N_{9}$  & MNIST   & 6x500     & PGD-$\epsilon{=}0.1$  & FCN           & 6  & 0.005  & 100\\
     $N_{10}$ & MNIST   & 6x500     & PGD-$\epsilon{=}0.3$  & FCN           & 6  & 0.005  & 100\\
     $N_{11}$ & MNIST   & 4x1024    & Standard              & FCN           & 3  & 0.004  & 100\\
     $N_{12}$ & MNIST   & ConvSmall & Standard              & convolutional & 3  & 0.004  & 10\\
     $N_{13}$ & MNIST   & ConvSmall & PGD                   & convolutional & 3  & 0.005  & 100\\
     $N_{14}$ & MNIST   & ConvSmall & DiffAI                & convolutional & 3  & 0.1    & 100\\
     $N_{15}$ & MNIST   & ConvMed   & Standard              & convolutional & 3  & 0.004  & 10\\
     $N_{16}$ & MNIST   & ConvBig   & DiffAI                & convolutional & 6  & 0.2    & 10\\
     $N_{17}$ & MNIST   & ConvSuper & DiffAI                & convolutional & 6  & 0.14   & 10\\
     $N_{18}$ & MNIST   & Skip      & DiffAI                & Residual      & 6  & 0.14   & 10\\
     $N_{19}$ & CIFAR10 & 4x100     & Standard              & FCN           & 5  & 4e-5   & 100\\
     $N_{20}$ & CIFAR10 & 6x100     & Standard              & FCN           & 7  & 4e-5   & 100\\
     $N_{21}$ & CIFAR10 & 9x200     & Standard              & FCN           & 10 & 4e-5   & 100\\
     $N_{22}$ & CIFAR10 & 6x500     & Standard              & FCN           & 6  & 4e-5   & 100\\
     $N_{23}$ & CIFAR10 & 6x500     & PGD-$\epsilon{=}0.0078$ & FCN         & 6  & 5e-4   & 100\\
     $N_{24}$ & CIFAR10 & 6x500     & PGD-$\epsilon{=}0.0313$ & FCN         & 6  & 5e-4   & 100\\
     $N_{25}$ & CIFAR10 & 7x1024    & Standard              & FCN           & 6  & 4e-5   & 100\\
     $N_{26}$ & CIFAR10 & ConvSmall & Standard              & convolutional & 3  & 4e-5   & 100\\
     $N_{27}$ & CIFAR10 & ConvSmall & PGD                   & convolutional & 3  & 5e-4   & 100\\
     $N_{28}$ & CIFAR10 & ConvSmall & DiffAI                & convolutional & 3  & 0.004 & 100\\
     $N_{29}$ & CIFAR10 & ConvMed   & Standard              & convolutional & 3  & 4e-5   & 100\\
     $N_{30}$ & CIFAR10 & ConvMed   & PGD-$\epsilon{=}0.0078$ & convolutional & 3 & 5e-4  & 10\\
     $N_{31}$ & CIFAR10 & ConvMed   & PGD-$\epsilon{=}0.0313$ & convolutional & 3 & 5e-4  & 100\\
     $N_{32}$ & CIFAR10 & ConvBig   & DiffAI                & convolutional & 6  & 0.002  & 10\\
    \bottomrule
    \end{tabular}
    \caption{Details of all 32 DNNs used in the evaluation (\S\ref{sec:eval}).
      Table~\ref{tab:repnets} lists the six representative networks.}
    \label{tab:allnets}
\end{table}
\clearpage
\subsection{Design space across all 32 networks.}
\label{app:evalall}
\begin{figure}[h]
\centering
\includegraphics[width=\columnwidth]{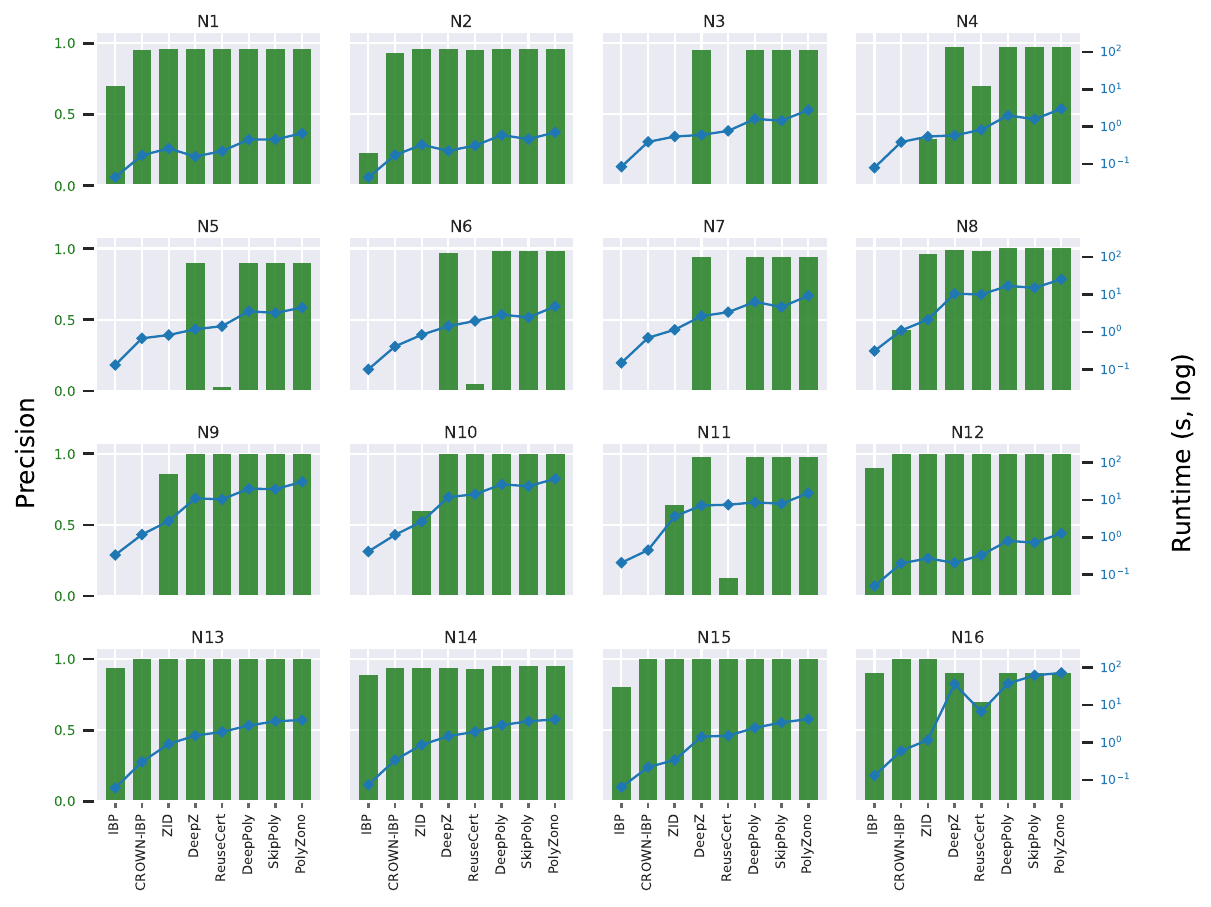}
\caption{Per-network precision--runtime design space, networks N1--N16. For each
  certifier, the green bar is its precision (left axis) and the blue marker is its
  runtime (right axis; log scale, shared with Fig.~\ref{fig:designspace-app2}).}
\label{fig:designspace-app1}
\end{figure}

\begin{figure}[h]
\centering
\includegraphics[width=\columnwidth]{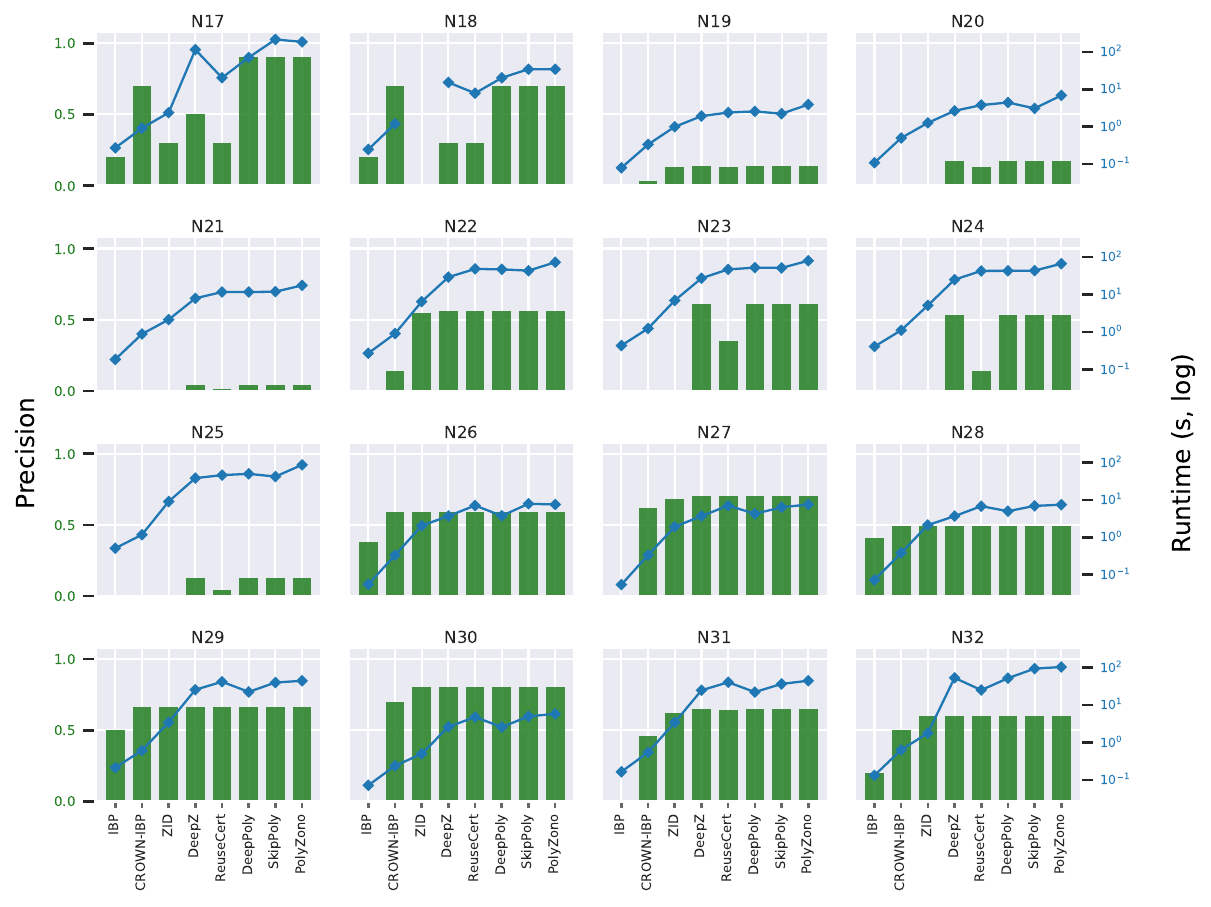}
\caption{Per-network precision--runtime design space, networks N17--N32
  (continued from Fig.~\ref{fig:designspace-app1}, same axes).}
\label{fig:designspace-app2}
\end{figure}

Figures~\ref{fig:designspace-app1} and~\ref{fig:designspace-app2} show the
precision--runtime design space of the eight \compiler-generated certifiers on
each of the 32 DNNs individually (the six networks in
Fig.~\ref{fig:designspace-pernet} are a representative subset), split into
networks N1--N16 and N17--N32. The per-network view confirms the pattern from the
main text: the cheap certifiers (IBP, CROWN-IBP) are far less precise on most
networks; the precise certifiers cluster at high precision but at widely different
cost; and ZID consistently occupies either an intermediate operating point or the
cheapest slot among equally-precise certifiers --- one that no existing library
offers. Such an exploration, comparing eight certifiers (four of which no library
supports) across the full benchmark suite, is only possible because each is a
short specification compiled by \compiler.

The remaining networks, N17--N32, are shown in Fig.~\ref{fig:designspace-app2} on
the same axes.

\clearpage
\subsection{$\mathit{DeepPoly}_p$ across all 32 networks.}
\label{app:evaldeeppoly}
\begin{figure}[h]
\centering
\includegraphics[width=\columnwidth]{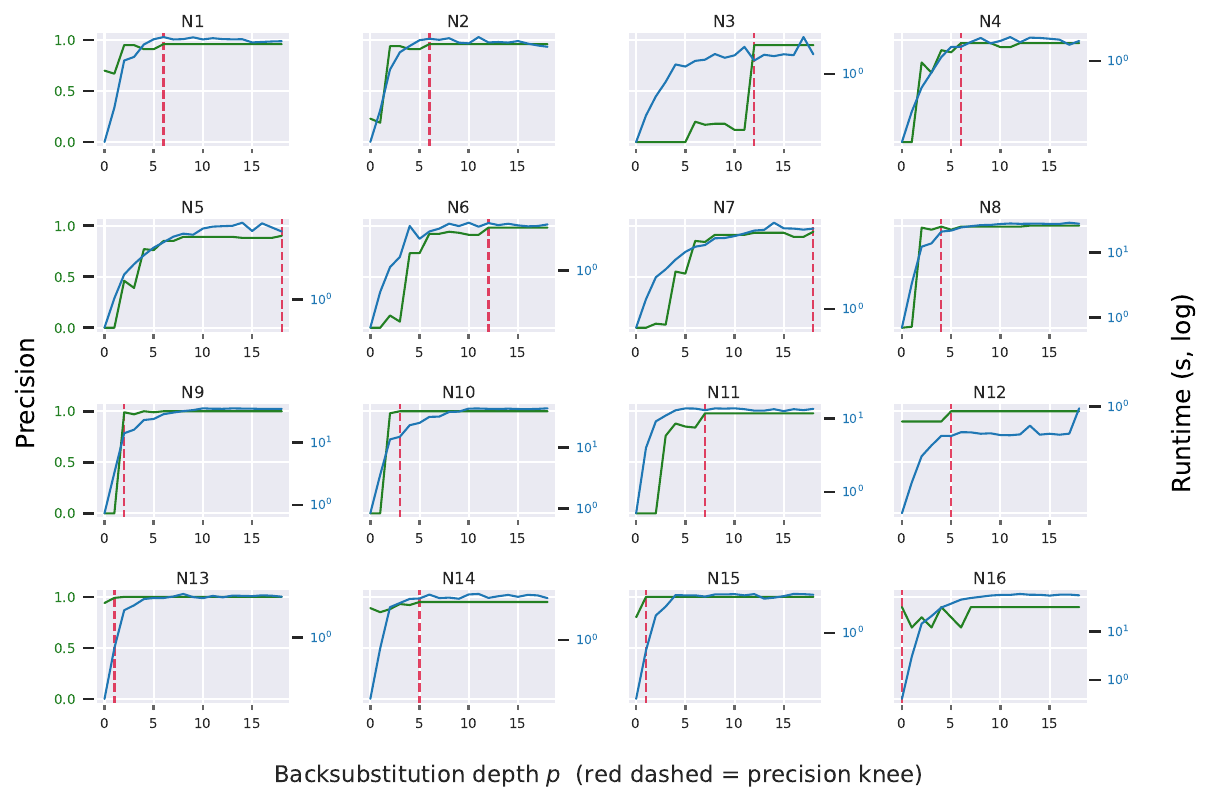}
\caption{$\mathit{DeepPoly}_p$ on networks N1--N16: precision (green) and runtime
  (blue, log) as backsubstitution depth $p$ increases; the red dashed line marks
  the smallest $p$ at which precision saturates. Continued in
  Fig.~\ref{fig:psweep-all2}.}
\label{fig:psweep-all1}
\end{figure}

\begin{figure}[h]
\centering
\includegraphics[width=\columnwidth]{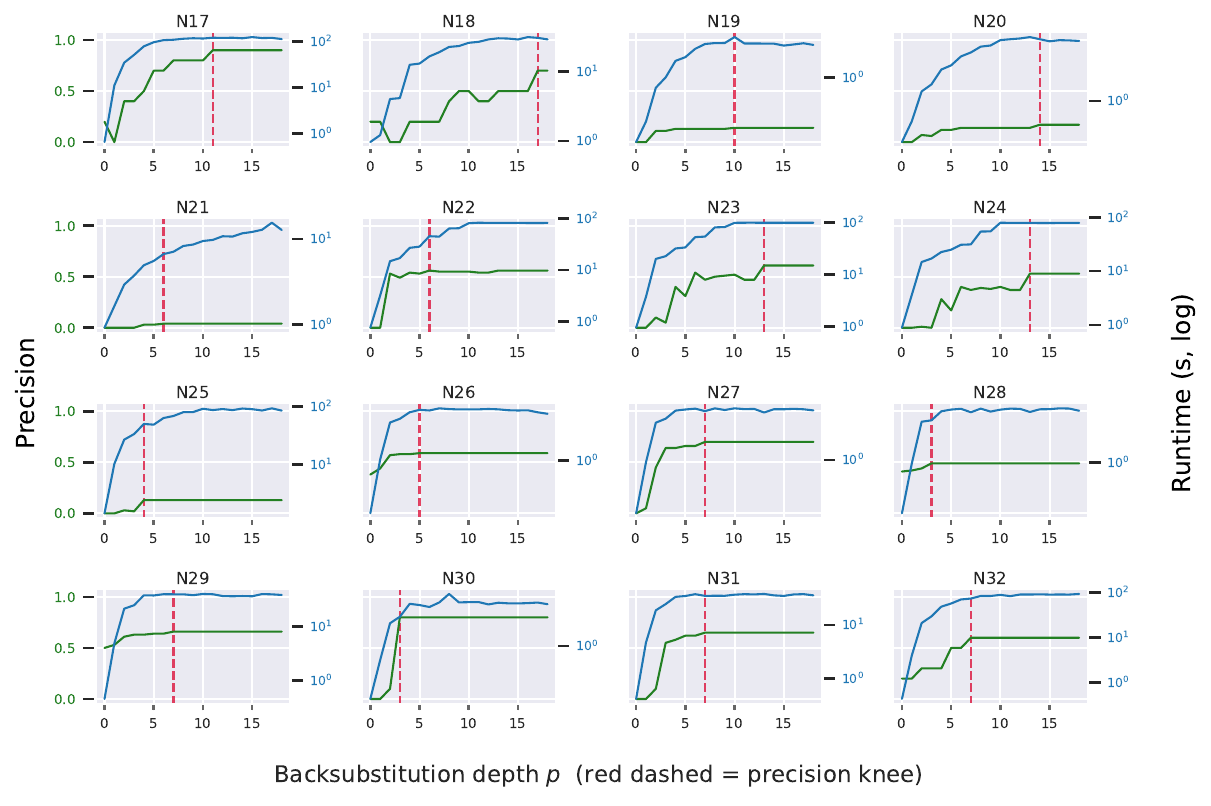}
\caption{$\mathit{DeepPoly}_p$ on networks N17--N32 (continued from
  Fig.~\ref{fig:psweep-all1}).}
\label{fig:psweep-all2}
\end{figure}

We report the backsubstitution-depth sweep of \S\ref{sec:eval-explore} over all
32 DNNs. The depth at which precision saturates ranges from $p=0$ to full depth
and cannot be predicted from the training regime; on the networks that saturate
early, stopping at the knee saves substantial runtime (up to $139\times$) at no
precision cost.
Figures~\ref{fig:psweep-all1} and~\ref{fig:psweep-all2} show the per-network
precision and runtime curves of $\mathit{DeepPoly}_p$ for all 32 DNNs individually
(the six networks in Fig.~\ref{fig:psweep} are a representative subset), split into
networks N1--N16 and N17--N32.

The remaining networks, N17--N32, are shown in Fig.~\ref{fig:psweep-all2}.

\clearpage
\subsection{Per-Certifier Runtime Comparison}
\label{app:rq3-tables}

This appendix supplements \S\ref{sec:eval-perf} (RQ3) with the absolute per-DNN runtimes underlying Fig.~\ref{fig:perf-alt}. Table~\ref{tab:rq3-summary} summarises, for each of the four certifiers that at least one baseline library implements, the aggregate speedup of \compiler over each baseline; Table~\ref{tab:rq3-full} then reports the full per-DNN wall-clock times (in seconds) and speedups on each of the 32 DNNs of Table~\ref{tab:allnets}. The speedup of \compiler over a baseline is the baseline runtime divided by the \compiler runtime, so a value above $1$ means \compiler is faster. In Table~\ref{tab:rq3-full}, TO marks a run exceeding the $200$\,s reporting threshold of \S\ref{sec:eval-perf}, and -- marks a configuration the baseline does not implement or errors on.

\begin{table}[h]
\centering\small
\begin{tabular}{ll rrr r}
\toprule
Certifier & Baseline & Mean & GeoMean & Median & $n$ \\
\midrule
IBP & auto\_LiRPA & 3.0 & 1.6 & 1.0 & 32 \\
 & IVAN & 2.7 & 1.1 & 0.60 & 31 \\
\addlinespace[2pt]
DeepZ & IVAN & 1.3 & 0.96 & 1.0 & 31 \\
 & ERAN & 25 & 17 & 16 & 17 \\
\addlinespace[2pt]
DeepPoly & auto\_LiRPA & 0.73 & 0.58 & 0.63 & 32 \\
 & IVAN & 7.5 & 3.5 & 3.4 & 24 \\
 & ERAN & 13 & 11 & 12 & 17 \\
\addlinespace[2pt]
CROWN-IBP & auto\_LiRPA & 1.0 & 0.77 & 0.62 & 32 \\
\bottomrule
\end{tabular}
\caption{Aggregate speedup of \compiler over each baseline (speedup $=$ baseline$/$\compiler runtime; ${>}1$ means \compiler is faster), summarising Table~\ref{tab:rq3-full}. Each row aggregates over the $n$ DNNs on which both \compiler and the baseline finished within the $200$\,s timeout. Table~\ref{tab:rq3-full} gives the underlying per-DNN times.}
\label{tab:rq3-summary}
\end{table}

\begin{table*}[h]
\centering\footnotesize\setlength{\tabcolsep}{3.5pt}
\begin{tabular}{l|rrrr|rrrr|rrrr|rrrr|}
\toprule
 & \multicolumn{4}{c|}{\textbf{IBP}} & \multicolumn{4}{c|}{\textbf{DeepZ}} & \multicolumn{4}{c|}{\textbf{DeepPoly}} & \multicolumn{4}{c|}{\textbf{CROWN-IBP}} \\
\cmidrule(lr){2-5} \cmidrule(lr){6-9} \cmidrule(lr){10-13} \cmidrule(lr){14-17}
Network & \textsc{n} & aL & IV & ER & \textsc{n} & aL & IV & ER & \textsc{n} & aL & IV & ER & \textsc{n} & aL & IV & ER \\
\midrule
$N_{1}$ & 0.04 & 0.04 & \textbf{0.03} & -- & \textbf{0.16} & -- & 0.19 & 6.45 & 0.45 & \textbf{0.12} & 0.41 & 3.73 & 0.17 & \textbf{0.10} & -- & -- \\
$N_{2}$ & 0.04 & 0.04 & \textbf{0.03} & -- & \textbf{0.23} & -- & 0.23 & 13.3 & 0.60 & \textbf{0.18} & 0.65 & 8.05 & 0.17 & \textbf{0.10} & -- & -- \\
$N_{3}$ & 0.09 & 0.05 & \textbf{0.03} & -- & 0.60 & -- & \textbf{0.41} & 30.7 & 1.61 & \textbf{0.38} & 1.32 & 27.3 & 0.39 & \textbf{0.11} & -- & -- \\
$N_{4}$ & 0.08 & 0.05 & \textbf{0.03} & -- & 0.59 & -- & \textbf{0.33} & 28.3 & 2.00 & \textbf{0.37} & 1.29 & 13.3 & 0.39 & \textbf{0.11} & -- & -- \\
$N_{5}$ & 0.13 & 0.06 & \textbf{0.03} & -- & 1.18 & -- & \textbf{0.47} & 38.7 & 3.58 & \textbf{0.91} & 2.07 & 21.3 & 0.68 & \textbf{0.13} & -- & -- \\
$N_{6}$ & 0.10 & 0.06 & \textbf{0.03} & -- & 1.43 & -- & \textbf{0.64} & TO & 2.92 & \textbf{0.82} & 3.46 & TO & 0.41 & \textbf{0.12} & -- & -- \\
$N_{7}$ & 0.15 & 0.07 & \textbf{0.03} & -- & 2.66 & -- & \textbf{0.99} & TO & 6.39 & \textbf{1.60} & 5.53 & TO & 0.70 & \textbf{0.15} & -- & -- \\
$N_{8}$ & 0.31 & 0.17 & \textbf{0.10} & -- & 10.5 & -- & \textbf{10.2} & TO & \textbf{17.0} & 17.5 & 42.9 & TO & 1.09 & \textbf{0.61} & -- & -- \\
$N_{9}$ & 0.34 & 0.25 & \textbf{0.15} & -- & \textbf{11.0} & -- & 14.2 & TO & \textbf{19.8} & 25.0 & 99.2 & TO & 1.18 & \textbf{0.76} & -- & -- \\
$N_{10}$ & 0.41 & 0.21 & \textbf{0.11} & -- & 11.8 & -- & \textbf{11.6} & TO & \textbf{25.9} & 34.3 & 109 & TO & 1.16 & \textbf{0.60} & -- & -- \\
$N_{11}$ & 0.21 & 0.09 & \textbf{0.04} & -- & 7.07 & -- & \textbf{3.97} & TO & 8.54 & \textbf{5.16} & 22.7 & TO & 0.45 & \textbf{0.19} & -- & -- \\
$N_{12}$ & 0.05 & 0.05 & \textbf{0.02} & -- & \textbf{0.21} & -- & 0.25 & 2.17 & 0.80 & \textbf{0.24} & 3.38 & 4.05 & 0.20 & \textbf{0.11} & -- & -- \\
$N_{13}$ & 0.06 & 0.06 & \textbf{0.03} & -- & \textbf{1.53} & -- & 3.47 & 25.8 & 2.83 & \textbf{1.55} & 36.1 & 51.1 & 0.30 & \textbf{0.18} & -- & -- \\
$N_{14}$ & 0.07 & 0.07 & \textbf{0.05} & -- & \textbf{1.48} & -- & 3.33 & 24.1 & \textbf{2.91} & 5.56 & 30.5 & 34.0 & \textbf{0.34} & 0.41 & -- & -- \\
$N_{15}$ & 0.06 & 0.06 & \textbf{0.02} & -- & 1.43 & -- & \textbf{0.46} & 40.4 & 2.47 & \textbf{1.23} & 9.99 & 38.8 & 0.22 & \textbf{0.13} & -- & -- \\
$N_{16}$ & 0.13 & 0.09 & \textbf{0.04} & -- & 36.5 & -- & \textbf{15.8} & 76.4 & 37.5 & \textbf{36.5} & TO & 158 & 0.60 & \textbf{0.33} & -- & -- \\
$N_{17}$ & 0.27 & 0.26 & \textbf{0.07} & -- & \textbf{115} & -- & 162 & TO & 71.0 & \textbf{39.6} & TO & TO & 0.92 & \textbf{0.44} & -- & -- \\
$N_{18}$ & 0.25 & \textbf{0.19} & -- & -- & \textbf{15.3} & -- & -- & TO & \textbf{20.3} & 21.7 & TO & -- & 1.20 & \textbf{0.53} & -- & -- \\
$N_{19}$ & \textbf{0.08} & 0.67 & 0.63 & -- & \textbf{1.92} & -- & 3.41 & 20.2 & 2.57 & \textbf{1.23} & 11.7 & 24.1 & \textbf{0.34} & 0.74 & -- & -- \\
$N_{20}$ & \textbf{0.11} & 0.68 & 0.63 & -- & \textbf{2.66} & -- & 3.55 & 21.8 & 4.44 & \textbf{1.55} & 14.8 & 30.3 & \textbf{0.50} & 0.75 & -- & -- \\
$N_{21}$ & \textbf{0.19} & 0.69 & 0.64 & -- & 7.93 & -- & \textbf{5.07} & TO & 11.6 & \textbf{4.10} & 37.9 & TO & 0.88 & \textbf{0.78} & -- & -- \\
$N_{22}$ & \textbf{0.27} & 0.76 & 0.68 & -- & \textbf{29.3} & -- & 55.2 & TO & \textbf{47.2} & 71.7 & TO & TO & \textbf{0.91} & 1.05 & -- & -- \\
$N_{23}$ & \textbf{0.43} & 0.83 & 0.71 & -- & \textbf{27.3} & -- & 47.6 & TO & \textbf{52.5} & 89.3 & TO & TO & \textbf{1.24} & 1.28 & -- & -- \\
$N_{24}$ & \textbf{0.41} & 0.80 & 0.70 & -- & \textbf{25.1} & -- & 30.1 & TO & \textbf{43.1} & 68.3 & TO & TO & \textbf{1.11} & 1.15 & -- & -- \\
$N_{25}$ & \textbf{0.51} & 0.80 & 0.67 & -- & 38.0 & -- & \textbf{16.6} & TO & 49.4 & \textbf{32.7} & 168 & TO & 1.17 & \textbf{0.95} & -- & -- \\
$N_{26}$ & \textbf{0.06} & 0.68 & 0.63 & -- & \textbf{3.68} & -- & 11.5 & 37.5 & 3.73 & \textbf{2.85} & 140 & 97.9 & \textbf{0.33} & 0.81 & -- & -- \\
$N_{27}$ & \textbf{0.05} & 0.68 & 0.63 & -- & \textbf{3.68} & -- & 13.0 & 44.3 & 4.24 & \textbf{3.04} & 141 & 99.3 & \textbf{0.33} & 0.80 & -- & -- \\
$N_{28}$ & \textbf{0.07} & 0.73 & 0.65 & -- & \textbf{3.66} & -- & 13.3 & 33.5 & \textbf{4.97} & 5.00 & 139 & 61.8 & \textbf{0.38} & 0.96 & -- & -- \\
$N_{29}$ & \textbf{0.21} & 0.69 & 0.63 & -- & 25.7 & -- & \textbf{3.11} & 178 & 22.5 & \textbf{2.45} & 34.5 & 103 & \textbf{0.61} & 0.77 & -- & -- \\
$N_{30}$ & \textbf{0.07} & 0.69 & 0.63 & -- & \textbf{2.62} & -- & 2.79 & 153 & 2.58 & \textbf{2.37} & 33.4 & 93.9 & \textbf{0.24} & 0.76 & -- & -- \\
$N_{31}$ & \textbf{0.16} & 0.73 & 0.66 & -- & 24.8 & -- & \textbf{24.0} & TO & 22.3 & \textbf{16.2} & TO & TO & \textbf{0.55} & 0.91 & -- & -- \\
$N_{32}$ & \textbf{0.13} & 0.71 & 0.64 & -- & 53.4 & -- & \textbf{49.6} & TO & 52.3 & \textbf{37.1} & TO & TO & \textbf{0.63} & 0.89 & -- & -- \\
\midrule
GeoMean & \textbf{0.13} & 0.22 & 0.14 & -- & 4.70 & -- & \textbf{4.36} & 29.2 & 7.58 & \textbf{4.40} & 15.2 & 32.2 & 0.52 & \textbf{0.40} & -- & -- \\
\bottomrule
\end{tabular}
\caption{Per-DNN runtime (seconds) with the four baseline-supported certifiers placed side by side, so \compiler and the baselines can be compared across certifiers on each DNN. Tool codes: \textsc{n}~$=$~\compiler, aL~$=$~auto\_LiRPA, IV~$=$~IVAN, ER~$=$~ERAN. Each certifier lists all four tools; for each certifier, the fastest tool on each DNN is shown in \textbf{bold}. TO~$=$ exceeded the $200$\,s threshold; --~$=$ the baseline does not implement that certifier (or errored). The final row is the per-column geometric mean over runs completed within the threshold. Aggregate speedups are in Table~\ref{tab:rq3-summary}.}
\label{tab:rq3-full}
\end{table*}

\clearpage

\clearpage

\subsection{Ablation Runtime Details}
\label{app:rq4-tables}

This appendix supplements \S\ref{sec:ablation} (RQ4) with per-DNN runtimes for full \compiler and its ablated variants. Table~\ref{tab:rq4-summary} summarises, per certifier, the slowdown of each variant relative to full \compiler; Tables~\ref{tab:rq4-fullA} and~\ref{tab:rq4-fullB} give the underlying per-DNN wall-clock times, with the variants placed side by side so they can be compared across certifiers on each of the 32 DNNs of Table~\ref{tab:allnets}. Variant codes relative to full \compiler (\textsc{n}): $-$S removes the \sparse representation, $-$R removes the domain-specific rewrites, and $-$RS removes both. TO marks a run exceeding the $200$\,s reporting threshold, \ding{55} a crash or out-of-memory, and -- missing data.

\begin{table}[htbp]
\centering\small
\begin{tabular}{ll rrr rr}
\toprule
Certifier & Variant & Mean & GeoMean & Median & $n$ & fail \\
\midrule
IBP & N$-$S & 49 & 24 & 24 & 28 & 4 \\
 & N$-$R & 51 & 8.0 & 3.3 & 32 & 0 \\
 & N$-$R$-$S & 55 & 25 & 26 & 28 & 4 \\
\addlinespace[2pt]
CROWN-IBP & N$-$S & 10 & 5.7 & 6.2 & 28 & 4 \\
 & N$-$R & 25 & 6.9 & 4.1 & 29 & 3 \\
 & N$-$R$-$S & 20 & 8.7 & 8.8 & 28 & 4 \\
\addlinespace[2pt]
DeepZ & N$-$S & 7.0 & 3.5 & 2.1 & 28 & 4 \\
 & N$-$R & 2.0 & 1.7 & 1.4 & 31 & 1 \\
 & N$-$R$-$S & 7.1 & 3.6 & 2.0 & 28 & 4 \\
\addlinespace[2pt]
DeepPoly & N$-$S & 3.2 & 2.3 & 1.4 & 28 & 4 \\
 & N$-$R & 16 & 8.3 & 7.8 & 14 & 18 \\
 & N$-$R$-$S & 38 & 13 & 8.6 & 17 & 15 \\
\addlinespace[2pt]
ReuseCert & N$-$S & 3.8 & 2.7 & 1.9 & 28 & 4 \\
 & N$-$R & 48 & 20 & 24 & 20 & 12 \\
 & N$-$R$-$S & 47 & 21 & 24 & 20 & 12 \\
\addlinespace[2pt]
PolyZono & N$-$S & 2.7 & 1.9 & 1.2 & 28 & 4 \\
 & N$-$R & 11 & 5.8 & 4.7 & 14 & 18 \\
 & N$-$R$-$S & 26 & 9.3 & 6.0 & 17 & 15 \\
\addlinespace[2pt]
ZID & N$-$S & 9.2 & 4.9 & 3.3 & 28 & 3 \\
 & N$-$R & 14 & 5.2 & 2.9 & 29 & 2 \\
 & N$-$R$-$S & 15 & 6.8 & 4.9 & 28 & 3 \\
\addlinespace[2pt]
SkipPoly & N$-$S & 2.9 & 2.2 & 1.6 & 28 & 3 \\
 & N$-$R & 44 & 17 & 14 & 17 & 14 \\
 & N$-$R$-$S & 46 & 17 & 13 & 17 & 14 \\
\bottomrule
\end{tabular}
\caption{Ablation slowdown of each variant relative to full \compiler (slowdown $=$ variant$/$\compiler runtime; ${>}1$ means the disabled optimization was helping). Variants: N$-$S removes \sparse, N$-$R removes rewrites, N$-$R$-$S removes both. Each row aggregates over the $n$ DNNs the variant completed; \emph{fail} counts DNNs where the variant crashed, ran out of memory, or timed out but full \compiler succeeded. Table~\ref{tab:rq4-fullA} and Table~\ref{tab:rq4-fullB} give the per-DNN times.}
\label{tab:rq4-summary}
\end{table}

\begin{table*}[htbp]
\centering\footnotesize\setlength{\tabcolsep}{3.5pt}
\begin{tabular}{l|rrrr|rrrr|rrrr|rrrr|}
\toprule
 & \multicolumn{4}{c|}{\textbf{IBP}} & \multicolumn{4}{c|}{\textbf{CROWN-IBP}} & \multicolumn{4}{c|}{\textbf{DeepZ}} & \multicolumn{4}{c|}{\textbf{DeepPoly}} \\
\cmidrule(lr){2-5} \cmidrule(lr){6-9} \cmidrule(lr){10-13} \cmidrule(lr){14-17}
Network & \textsc{n} & $-$S & $-$R & $-$RS & \textsc{n} & $-$S & $-$R & $-$RS & \textsc{n} & $-$S & $-$R & $-$RS & \textsc{n} & $-$S & $-$R & $-$RS \\
\midrule
$N_{1}$ & \textbf{0.03} & 0.09 & 0.04 & 0.08 & \textbf{0.19} & 0.26 & 0.21 & 0.23 & \textbf{0.07} & 0.12 & 0.10 & 0.12 & \textbf{0.33} & 0.44 & 0.42 & 0.44 \\
$N_{2}$ & \textbf{0.03} & 0.10 & 0.04 & 0.10 & \textbf{0.19} & 0.27 & 0.24 & 0.27 & \textbf{0.07} & 0.13 & 0.10 & 0.14 & \textbf{0.32} & 0.46 & 0.80 & 0.79 \\
$N_{3}$ & \textbf{0.05} & 0.19 & 0.08 & 0.17 & \textbf{0.47} & 0.50 & 0.66 & 0.60 & \textbf{0.25} & 0.32 & 0.31 & 0.34 & \textbf{1.27} & 1.55 & 3.52 & 3.29 \\
$N_{4}$ & \textbf{0.06} & 0.16 & 0.08 & 0.17 & \textbf{0.58} & 0.60 & 0.65 & 0.60 & \textbf{0.24} & 0.31 & 0.33 & 0.34 & \textbf{1.17} & 1.54 & 3.52 & 3.28 \\
$N_{5}$ & \textbf{0.09} & 0.27 & 0.15 & 0.27 & 0.84 & \textbf{0.83} & 1.49 & 1.16 & \textbf{0.58} & 0.75 & 0.79 & 0.76 & \textbf{2.97} & 3.18 & 8.21 & 6.95 \\
$N_{6}$ & \textbf{0.06} & 0.30 & 0.09 & 0.36 & \textbf{0.39} & 0.73 & 0.76 & 0.88 & \textbf{0.28} & 0.50 & 0.37 & 0.56 & \textbf{1.41} & 1.78 & 12.5 & 10.4 \\
$N_{7}$ & \textbf{0.10} & 0.58 & 0.16 & 0.66 & \textbf{0.84} & 1.19 & 1.69 & 1.60 & \textbf{0.69} & 1.09 & 0.90 & 1.16 & 2.85 & \textbf{2.63} & 26.2 & 24.5 \\
$N_{8}$ & \textbf{0.10} & 1.62 & 0.21 & 1.69 & \textbf{0.59} & 2.21 & 1.81 & 3.09 & \textbf{1.13} & 2.12 & 1.18 & 2.03 & \textbf{3.15} & 4.29 & 106 & 103 \\
$N_{9}$ & \textbf{0.11} & 1.63 & 0.20 & 1.69 & \textbf{0.57} & 2.24 & 1.76 & 3.02 & 1.28 & 2.08 & \textbf{1.21} & 2.26 & \textbf{3.67} & 4.91 & 106 & 109 \\
$N_{10}$ & \textbf{0.11} & 1.40 & 0.25 & 1.43 & \textbf{0.59} & 2.06 & 1.73 & 3.02 & \textbf{1.37} & 2.13 & 1.44 & 2.15 & \textbf{4.19} & 5.49 & 114 & 111 \\
$N_{11}$ & \textbf{0.07} & 1.68 & 0.30 & 1.88 & \textbf{0.27} & 1.75 & 1.97 & 3.30 & \textbf{0.69} & 2.33 & 0.78 & 2.56 & \textbf{1.55} & 2.94 & 117 & 121 \\
$N_{12}$ & \textbf{0.03} & 1.96 & 0.40 & 2.44 & \textbf{0.25} & 2.41 & 3.12 & 4.72 & \textbf{0.17} & 3.33 & 0.54 & 3.49 & \textbf{0.70} & 3.16 & TO & 98.9 \\
$N_{13}$ & \textbf{0.04} & 2.17 & 0.39 & 2.43 & \textbf{0.21} & 2.41 & 3.00 & 4.68 & \textbf{0.19} & 3.58 & 0.54 & 3.34 & \textbf{0.67} & 3.13 & TO & 93.6 \\
$N_{14}$ & \textbf{0.04} & 2.17 & 0.44 & 2.41 & \textbf{0.21} & 2.53 & 2.84 & 4.83 & \textbf{0.19} & 3.46 & 0.50 & 3.56 & \textbf{0.67} & 3.51 & TO & 97.2 \\
$N_{15}$ & \textbf{0.05} & 5.00 & 0.98 & 5.08 & \textbf{0.33} & 4.84 & 6.64 & 9.55 & \textbf{1.48} & 7.49 & 2.63 & 7.66 & \textbf{2.57} & 7.98 & \ding{55} & TO \\
$N_{16}$ & \textbf{0.11} & \ding{55} & 30.4 & \ding{55} & \textbf{0.52} & \ding{55} & 173 & \ding{55} & \textbf{36.6} & \ding{55} & 110 & \ding{55} & \textbf{34.5} & \ding{55} & \ding{55} & \ding{55} \\
$N_{17}$ & \textbf{0.26} & \ding{55} & 103 & \ding{55} & \textbf{0.87} & \ding{55} & TO & \ding{55} & \textbf{117} & \ding{55} & TO & \ding{55} & \textbf{69.6} & \ding{55} & \ding{55} & \ding{55} \\
$N_{18}$ & \textbf{0.22} & \ding{55} & 36.8 & \ding{55} & \textbf{1.02} & \ding{55} & \ding{55} & \ding{55} & \textbf{14.9} & \ding{55} & 61.2 & \ding{55} & \textbf{19.1} & \ding{55} & \ding{55} & \ding{55} \\
$N_{19}$ & \textbf{0.05} & 0.81 & 0.07 & 0.69 & \textbf{0.31} & 1.09 & 0.54 & 1.20 & \textbf{0.23} & 0.86 & 0.27 & 0.99 & \textbf{0.97} & 1.36 & 6.17 & 6.70 \\
$N_{20}$ & \textbf{0.07} & 0.92 & 0.10 & 0.83 & \textbf{0.59} & 1.33 & 0.89 & 1.53 & \textbf{0.41} & 1.23 & 0.47 & 1.21 & \textbf{1.69} & 2.14 & 11.3 & 11.0 \\
$N_{21}$ & \textbf{0.12} & 1.87 & 0.18 & 1.93 & \textbf{0.88} & 2.53 & 1.97 & 3.13 & \textbf{1.15} & 2.59 & 1.30 & 2.41 & \textbf{4.30} & 4.45 & 61.0 & 60.1 \\
$N_{22}$ & \textbf{0.11} & 2.78 & 0.25 & 3.26 & \textbf{0.58} & 3.57 & 2.27 & 4.98 & 4.10 & \textbf{3.68} & 3.82 & 3.82 & \textbf{7.91} & 10.1 & TO & TO \\
$N_{23}$ & \textbf{0.12} & 3.07 & 0.25 & 3.29 & \textbf{0.56} & 3.46 & 2.31 & 5.01 & 3.96 & \textbf{3.75} & 3.82 & 3.97 & \textbf{9.19} & 11.3 & TO & TO \\
$N_{24}$ & \textbf{0.12} & 3.13 & 0.24 & 3.13 & \textbf{0.46} & 3.26 & 2.17 & 4.92 & 3.20 & 3.86 & \textbf{2.89} & 3.80 & \textbf{6.73} & 9.81 & TO & TO \\
$N_{25}$ & \textbf{0.17} & 6.85 & 0.76 & 7.69 & \textbf{0.54} & 7.25 & 5.35 & 11.5 & \textbf{5.16} & 10.1 & 5.42 & 10.2 & \textbf{6.57} & 13.9 & TO & TO \\
$N_{26}$ & \textbf{0.04} & 5.39 & 1.26 & 6.14 & \textbf{0.24} & 5.65 & 9.49 & 12.7 & \textbf{0.34} & 9.98 & 1.62 & 9.34 & \textbf{0.79} & 8.35 & \ding{55} & \ding{55} \\
$N_{27}$ & \textbf{0.04} & 5.65 & 1.29 & 6.18 & \textbf{0.24} & 6.14 & 9.64 & 12.4 & \textbf{0.32} & 9.15 & 1.57 & 9.33 & \textbf{0.78} & 8.59 & \ding{55} & \ding{55} \\
$N_{28}$ & \textbf{0.04} & 5.44 & 1.39 & 6.41 & \textbf{0.24} & 5.45 & 9.61 & 12.8 & \textbf{0.32} & 8.55 & 1.57 & 8.89 & \textbf{0.84} & 8.90 & \ding{55} & \ding{55} \\
$N_{29}$ & \textbf{0.07} & 9.25 & 2.70 & 10.2 & \textbf{0.32} & 9.36 & 16.1 & 19.9 & \textbf{2.87} & 15.3 & 6.13 & 15.6 & \textbf{3.40} & 16.1 & \ding{55} & \ding{55} \\
$N_{30}$ & \textbf{0.06} & 9.23 & 2.73 & 9.99 & \textbf{0.27} & 9.40 & 15.9 & 19.8 & \textbf{2.74} & 15.2 & 6.05 & 15.2 & \textbf{2.96} & 16.0 & \ding{55} & \ding{55} \\
$N_{31}$ & \textbf{0.06} & 8.84 & 2.72 & 10.0 & \textbf{0.27} & 9.41 & 15.2 & 19.7 & \textbf{2.72} & 15.5 & 6.08 & 15.1 & \textbf{3.65} & 15.6 & \ding{55} & \ding{55} \\
$N_{32}$ & \textbf{0.12} & \ding{55} & 54.1 & \ding{55} & \textbf{0.57} & \ding{55} & TO & \ding{55} & \textbf{52.9} & \ding{55} & 181 & \ding{55} & \textbf{52.1} & \ding{55} & \ding{55} & \ding{55} \\
\midrule
GeoMean & \textbf{0.07} & 1.52 & 0.58 & 1.61 & \textbf{0.41} & 2.19 & 2.68 & 3.33 & \textbf{1.11} & 2.32 & 1.66 & 2.38 & \textbf{2.75} & 4.24 & 13.6 & 18.6 \\
\bottomrule
\end{tabular}
\caption{Per-DNN ablation runtime (seconds, batch size 10) for IBP, CROWN-IBP, DeepZ, DeepPoly, with \compiler and its ablated variants side by side. Variant codes relative to full \compiler (\textsc{n}): $-$S removes \sparse, $-$R removes rewrites, $-$RS removes both. For each certifier, the fastest variant on each DNN is shown in \textbf{bold}. TO~$=$ exceeded the $200$\,s threshold; \ding{55}~$=$ crashed or ran out of memory; --~$=$ no data. The final row is the per-column geometric mean over runs completed within the threshold. Remaining certifiers are in Table~\ref{tab:rq4-fullB}; aggregate slowdowns in Table~\ref{tab:rq4-summary}.}
\label{tab:rq4-fullA}
\end{table*}

\begin{table*}[htbp]
\centering\footnotesize\setlength{\tabcolsep}{3.5pt}
\begin{tabular}{l|rrrr|rrrr|rrrr|rrrr|}
\toprule
 & \multicolumn{4}{c|}{\textbf{ReuseCert}} & \multicolumn{4}{c|}{\textbf{PolyZono}} & \multicolumn{4}{c|}{\textbf{ZID}} & \multicolumn{4}{c|}{\textbf{SkipPoly}} \\
\cmidrule(lr){2-5} \cmidrule(lr){6-9} \cmidrule(lr){10-13} \cmidrule(lr){14-17}
Network & \textsc{n} & $-$S & $-$R & $-$RS & \textsc{n} & $-$S & $-$R & $-$RS & \textsc{n} & $-$S & $-$R & $-$RS & \textsc{n} & $-$S & $-$R & $-$RS \\
\midrule
$N_{1}$ & \textbf{0.17} & 0.25 & 0.36 & 0.38 & 0.50 & \textbf{0.46} & 0.52 & 0.49 & \textbf{0.21} & 0.26 & 0.23 & 0.24 & \textbf{0.27} & 0.33 & 0.34 & 0.34 \\
$N_{2}$ & \textbf{0.18} & 0.26 & 0.65 & 1.01 & \textbf{0.50} & 0.59 & 0.91 & 0.86 & \textbf{0.20} & 0.28 & 0.27 & 0.29 & \textbf{0.28} & 0.36 & 0.67 & 0.75 \\
$N_{3}$ & \textbf{0.45} & 0.58 & 2.35 & 2.35 & \textbf{1.82} & 1.86 & 3.76 & 3.42 & \textbf{0.50} & 0.58 & 0.67 & 0.60 & \textbf{0.93} & 1.07 & 3.00 & 2.87 \\
$N_{4}$ & \textbf{0.45} & 0.57 & 2.38 & 2.36 & 1.93 & \textbf{1.43} & 3.84 & 3.49 & \textbf{0.49} & 0.53 & 0.70 & 0.61 & \textbf{0.79} & 1.02 & 2.95 & 2.88 \\
$N_{5}$ & \textbf{0.93} & 1.43 & 4.10 & 4.05 & \textbf{3.68} & 4.33 & 9.12 & 7.79 & \textbf{0.85} & 1.03 & 1.49 & 1.15 & 1.87 & \textbf{1.84} & 6.60 & 6.06 \\
$N_{6}$ & \textbf{0.52} & 0.76 & 6.74 & 6.84 & 2.15 & \textbf{2.05} & 11.2 & 11.1 & \textbf{0.51} & 0.68 & 0.82 & 0.84 & \textbf{0.89} & 1.17 & 10.7 & 10.6 \\
$N_{7}$ & \textbf{1.01} & 1.57 & 11.4 & 11.9 & \textbf{4.10} & 4.84 & 25.9 & 24.8 & \textbf{0.83} & 1.29 & 1.65 & 1.68 & \textbf{1.76} & 2.55 & 24.2 & 23.2 \\
$N_{8}$ & \textbf{1.50} & 2.76 & 44.7 & 45.4 & \textbf{4.79} & 5.49 & 107 & 106 & \textbf{0.63} & 2.06 & 1.80 & 3.16 & \textbf{2.69} & 4.20 & 109 & 109 \\
$N_{9}$ & \textbf{1.56} & 3.32 & 46.5 & 45.0 & \textbf{4.95} & 5.80 & 110 & 107 & \textbf{0.62} & 1.83 & 1.86 & 2.98 & \textbf{3.09} & 4.52 & 108 & 108 \\
$N_{10}$ & \textbf{2.05} & 3.73 & 45.8 & 47.2 & \textbf{5.65} & 6.67 & 114 & 111 & \textbf{0.63} & 2.08 & 1.82 & 2.77 & \textbf{3.63} & 5.19 & 111 & 113 \\
$N_{11}$ & \textbf{0.94} & 2.49 & 70.1 & 69.7 & \textbf{2.29} & 4.03 & 118 & 119 & \textbf{0.58} & 2.28 & 2.09 & 3.82 & \textbf{1.17} & 2.85 & 116 & 121 \\
$N_{12}$ & \textbf{0.36} & 2.81 & 73.6 & 75.6 & \textbf{1.00} & 4.43 & TO & 95.3 & \textbf{0.27} & 3.36 & 2.83 & 5.71 & \textbf{0.58} & 3.31 & 92.4 & 98.8 \\
$N_{13}$ & \textbf{0.38} & 2.92 & 77.0 & 74.7 & \textbf{0.96} & 5.48 & TO & 96.1 & \textbf{0.31} & 3.58 & 2.91 & 5.54 & \textbf{0.64} & 3.34 & 98.1 & 105 \\
$N_{14}$ & \textbf{0.37} & 3.22 & 83.9 & 74.6 & \textbf{0.99} & 4.75 & TO & 94.6 & \textbf{0.27} & 3.57 & 2.93 & 5.61 & \textbf{0.65} & 3.65 & 94.8 & 100 \\
$N_{15}$ & \textbf{1.54} & 6.50 & TO & TO & \textbf{4.04} & 11.0 & \ding{55} & TO & \textbf{0.37} & 7.33 & 6.50 & 11.3 & \textbf{3.61} & 8.07 & TO & TO \\
$N_{16}$ & \textbf{6.68} & \ding{55} & \ding{55} & \ding{55} & \textbf{73.4} & \ding{55} & \ding{55} & \ding{55} & \textbf{1.05} & \ding{55} & 169 & \ding{55} & \textbf{61.5} & \ding{55} & \ding{55} & \ding{55} \\
$N_{17}$ & \textbf{19.7} & \ding{55} & \ding{55} & \ding{55} & \textbf{184} & \ding{55} & \ding{55} & \ding{55} & \textbf{2.21} & \ding{55} & TO & \ding{55} & TO & \ding{55} & \ding{55} & \ding{55} \\
$N_{18}$ & \textbf{8.50} & \ding{55} & \ding{55} & \ding{55} & \textbf{32.1} & \ding{55} & \ding{55} & \ding{55} & \ding{55} & \ding{55} & \ding{55} & \ding{55} & \textbf{33.2} & \ding{55} & \ding{55} & \ding{55} \\
$N_{19}$ & \textbf{0.43} & 1.20 & 5.26 & 6.02 & \textbf{1.47} & 1.75 & 6.28 & 6.84 & \textbf{0.38} & 1.18 & 0.58 & 1.18 & \textbf{0.71} & 1.43 & 5.88 & 6.44 \\
$N_{20}$ & \textbf{0.88} & 1.52 & 8.58 & 8.99 & \textbf{2.81} & 2.89 & 11.3 & 11.6 & \textbf{0.65} & 1.45 & 0.95 & 1.62 & \textbf{1.27} & 2.05 & 10.2 & 10.7 \\
$N_{21}$ & \textbf{1.66} & 3.11 & 45.4 & 46.2 & 6.50 & \textbf{6.19} & 61.4 & 61.4 & \textbf{1.05} & 2.36 & 2.09 & 3.26 & \textbf{2.83} & 3.88 & 66.1 & 62.2 \\
$N_{22}$ & \textbf{6.89} & 9.40 & 172 & 177 & 12.0 & \textbf{11.4} & TO & TO & \textbf{1.20} & 3.67 & 2.99 & 4.86 & \textbf{7.66} & 10.6 & TO & TO \\
$N_{23}$ & \textbf{6.84} & 9.73 & 172 & 172 & \textbf{12.3} & 12.8 & TO & TO & \textbf{1.31} & 3.68 & 3.05 & 4.88 & \textbf{8.20} & 11.7 & TO & TO \\
$N_{24}$ & \textbf{5.27} & 8.23 & 166 & 165 & \textbf{10.2} & 10.5 & TO & TO & \textbf{0.88} & 3.46 & 2.58 & 5.15 & \textbf{6.84} & 9.79 & TO & TO \\
$N_{25}$ & \textbf{6.40} & 12.9 & TO & TO & \textbf{11.8} & 16.2 & TO & TO & \textbf{1.46} & 8.01 & 6.03 & 11.9 & \textbf{5.90} & 13.8 & TO & TO \\
$N_{26}$ & \textbf{0.73} & 9.65 & \ding{55} & \ding{55} & \textbf{1.24} & 12.1 & \ding{55} & \ding{55} & \textbf{0.44} & 8.86 & 9.59 & 15.7 & \textbf{1.08} & 8.78 & \ding{55} & \ding{55} \\
$N_{27}$ & \textbf{0.73} & 9.60 & \ding{55} & \ding{55} & \textbf{1.37} & 11.7 & \ding{55} & \ding{55} & \textbf{0.38} & 9.38 & 9.50 & 16.1 & \textbf{1.01} & 8.51 & \ding{55} & \ding{55} \\
$N_{28}$ & \textbf{1.03} & 9.35 & \ding{55} & \ding{55} & \textbf{1.26} & 11.3 & \ding{55} & \ding{55} & \textbf{0.41} & 9.94 & 8.73 & 16.4 & \textbf{1.06} & 8.23 & \ding{55} & \ding{55} \\
$N_{29}$ & \textbf{4.88} & 16.5 & \ding{55} & \ding{55} & \textbf{5.67} & 21.2 & \ding{55} & \ding{55} & \textbf{0.55} & 15.0 & 15.2 & 23.9 & \textbf{5.95} & 16.0 & \ding{55} & \ding{55} \\
$N_{30}$ & \textbf{4.76} & 16.2 & \ding{55} & \ding{55} & \textbf{5.77} & 21.7 & \ding{55} & \ding{55} & \textbf{0.49} & 14.4 & 16.4 & 25.2 & \textbf{4.17} & 16.7 & \ding{55} & \ding{55} \\
$N_{31}$ & \textbf{4.72} & 17.0 & \ding{55} & \ding{55} & \textbf{5.78} & 22.9 & \ding{55} & \ding{55} & \textbf{0.49} & 14.5 & 16.4 & 24.4 & \textbf{5.55} & 16.1 & \ding{55} & \ding{55} \\
$N_{32}$ & \textbf{25.2} & \ding{55} & \ding{55} & \ding{55} & \textbf{105} & \ding{55} & \ding{55} & \ding{55} & \textbf{1.55} & \ding{55} & TO & \ding{55} & \textbf{92.1} & \ding{55} & \ding{55} & \ding{55} \\
\midrule
GeoMean & \textbf{1.57} & 3.18 & 18.0 & 18.6 & \textbf{4.36} & 5.46 & 14.1 & 19.2 & \textbf{0.59} & 2.61 & 2.81 & 3.63 & \textbf{2.44} & 3.89 & 17.8 & 18.0 \\
\bottomrule
\end{tabular}
\caption{Per-DNN ablation runtime (seconds, batch size 10) for ReuseCert, PolyZono, ZID, SkipPoly, with \compiler and its ablated variants side by side. Variant codes relative to full \compiler (\textsc{n}): $-$S removes \sparse, $-$R removes rewrites, $-$RS removes both. For each certifier, the fastest variant on each DNN is shown in \textbf{bold}. TO~$=$ exceeded the $200$\,s threshold; \ding{55}~$=$ crashed or ran out of memory; --~$=$ no data. The final row is the per-column geometric mean over runs completed within the threshold. Remaining certifiers are in Table~\ref{tab:rq4-fullA}; aggregate slowdowns in Table~\ref{tab:rq4-summary}.}
\label{tab:rq4-fullB}
\end{table*}

\clearpage

\clearpage


\subsection{Peak Memory}
\label{app:memory}
\begin{figure}[h]
\centering
\includegraphics[width=\columnwidth,trim={0in 3mm 0in 0in},clip]{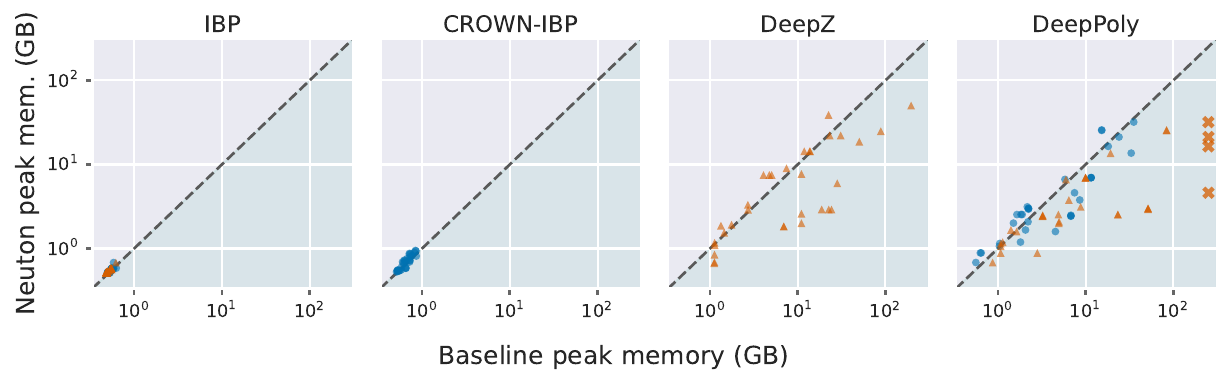}
\caption{Per-DNN peak memory of \compiler vs.\ existing libraries on all 32 DNNs
  (\protect\mkAuto~$=$ auto\_LiRPA, \protect\mkIvan~$=$ IVAN; log--log axes). A point \emph{below} the dashed diagonal means
  \compiler uses less memory on that DNN. \textsf{x} markers denote DNNs on which the baseline exceeded the $200$\,s
  timeout while \compiler finished.}
\label{fig:mem-scatter}
\end{figure}

Fig.~\ref{fig:mem-scatter} and Table~\ref{tab:mem-baselines} compare the peak memory of \compiler with the hand-optimized libraries on all 32 DNNs, for the four certifiers they implement, using the batch sizes of Table~\ref{tab:allnets}. Table~\ref{tab:mem-ablation} reports the peak memory of the ablated variants on the six representative DNNs (batch size $10$, as in \S\ref{sec:ablation}). Peak memory is the maximum resident set size of the process, measured with all tools pinned to $4$ threads.

\begin{table}[h]
\centering\footnotesize
\caption{Peak memory of hand-optimized libraries relative to \compiler on all 32 DNNs (geometric mean of
  baseline\,/\,\compiler over the DNNs both complete; $>1$ means \compiler uses less memory). ERAN is measured at batch
  size $1$ on the six representative DNNs. TO $=$ DNNs on which the baseline timed out.}
\label{tab:mem-baselines}
\begin{tabular}{@{}lcccc@{}}
\toprule
Baseline & IBP & CROWN-IBP & DeepZ & DeepPoly \\
\midrule
auto\_LiRPA & $0.98\times$ & $0.95\times$ & -- & $1.14\times$ \\
IVAN & $0.94\times$ & -- & $1.64\times$ & $2.26\times$\,{\scriptsize(4 TO)} \\
ERAN ($b{=}1$) & -- & -- & $2.05\times$ & $2.13\times$ \\
\bottomrule
\end{tabular}
\end{table}

\begin{table}[h]
\centering\footnotesize
\caption{Peak memory of each ablated variant relative to \compiler on the six representative DNNs
  (geometric mean over the DNNs on which the variant completes). Parentheses count the DNNs on which the variant ran
  out of memory while \compiler completed.}
\label{tab:mem-ablation}
\setlength{\tabcolsep}{3.5pt}
\begin{tabular}{@{}lccccccccc@{}}
\toprule
Variant & IBP & CROWN-IBP & DeepZ & DeepPoly & ReuseCert & PolyZono & ZID & SkipPoly & Overall \\
\midrule
N--S   & $18.1\times$\,(1) & $17.3\times$\,(1) & $6.3\times$\,(1) & $6.7\times$\,(1) & $6.8\times$\,(1) & $4.8\times$\,(1) & $11.5\times$\,(1) & $5.8\times$\,(1) & $8.6\times$\,(8) \\
N--R   & $3.4\times$ & $9.5\times$ & $1.3\times$ & $46.7\times$\,(3) & $129.8\times$\,(2) & $31.5\times$\,(3) & $6.3\times$ & $71.3\times$\,(2) & $11.1\times$\,(10) \\
N--R--S & $18.1\times$\,(1) & $17.8\times$\,(1) & $6.3\times$\,(1) & $75.5\times$\,(2) & $134.8\times$\,(2) & $53.5\times$\,(2) & $11.9\times$\,(1) & $75.9\times$\,(2) & $28.5\times$\,(12) \\
\bottomrule
\end{tabular}
\end{table}

\clearpage
\subsection{Certifier Specifications}
\label{app:specifications}

We give the complete \cf specifications of the eight certifiers evaluated in
\S\ref{sec:eval}. Each is compiled by \compiler into an efficient tensor program;
the specification is the only thing the user writes.

\textbf{IBP}

\begin{lstlisting}
def Shape as (Float l, Float u){[(curr[l]<=curr),(curr[u]>=curr)]};

func simplify_lower(Neuron n, Float coeff) = (coeff >= 0.0) ? (coeff * n[l]) : (coeff * n[u]);
func simplify_upper(Neuron n, Float coeff) = (coeff >= 0.0) ? (coeff * n[u]) : (coeff * n[l]);

func priority(Neuron n) = n[layer];
func stop(Neuron n) = true;

transformer ibp{
    Affine -> ((prev.dot(curr[weight]) + curr[bias]).map(simplify_lower), (prev.dot(curr[weight]) + curr[bias]).map(simplify_upper));
    
    Relu -> ((prev[l] >= 0 ? prev[l] : 0), (prev[u] >= 0 ? prev[u] : 0));

}

flow(forward, priority, stop, ibp);
\end{lstlisting}

\textbf{CROWN-IBP}
\begin{lstlisting}[escapeinside={(*@}{@*)}]
def Shape as (Float l, Float u, PolyExp L, PolyExp U){[(curr[l]<=curr),(curr[u]>=curr),(curr[L]<=curr),(curr[U]>=curr)]};

func simplify_lower(Neuron n, Float coeff) = (coeff >= 0.0) ? (n[l] * coeff) : (coeff * n[u]);
func simplify_upper(Neuron n, Float coeff) = (coeff >= 0.0) ? (coeff * n[u]) : (coeff * n[l]);

func replace_lower(Neuron n, Float coeff) = (coeff >= 0.0) ? (coeff * n[L]) : (coeff * n[U]);
func replace_upper(Neuron n, Float coeff) = (coeff >= 0.0) ? (coeff * n[U]) : (coeff * n[L]);

func priority(Neuron n) = n[layer];
func priority2(Neuron n, Float c) = -n[layer];
func stop(Neuron n) = false;
func stop_traverse(Neuron n, Float c) = false;

func backsubs_lower(PolyExp e, Neuron n) = (e.traverse(backward, priority2, stop_traverse, replace_lower){e <= n}).map(simplify_lower);
func backsubs_upper(PolyExp e, Neuron n) = (e.traverse(backward, priority2, stop_traverse, replace_upper){e >= n}).map(simplify_upper);

func relu(Float x) = x >= 0.0 ? x : 0.0;
func relu_n(Neuron n) = n[l] + n[u] >= 0.0 ? n : 0.0;
func compute_upper(Neuron n) = n[l] >= 0.0 ? n : (n[u] <= 0.0 ? 0.0 : (((n[u]) / ((n[u]) - (n[l]))) * ((n))) - (((n[u]) * (n[l])) / ((n[u]) - (n[l]))));

transformer crownibp{
    Affine -> curr[last_layer] == 1 ? (backsubs_lower(prev.dot(curr[weight]) + curr[bias], curr), backsubs_upper(prev.dot(curr[weight]) + curr[bias], curr), prev.dot(curr[weight]) + curr[bias], prev.dot(curr[weight]) + curr[bias]) : ((prev.dot(curr[weight]) + curr[bias]).map(simplify_lower), (prev.dot(curr[weight]) + curr[bias]).map(simplify_upper), prev.dot(curr[weight]) + curr[bias], prev.dot(curr[weight]) + curr[bias]);
    
    Relu -> (relu(prev[l]), relu(prev[u]), relu_n(prev), compute_upper(prev));

}

flow(forward, priority, stop, crownibp);
\end{lstlisting}

\textbf{DeepZ}
\begin{lstlisting}[escapeinside={(*@}{@*)}]
def Shape as (Float l, Float u, SymExp Z){[(curr[l]<=curr),(curr[u]>=curr),(curr In curr[Z])]};

func abs(Float x) = x > 0.0 ? x : 0.0-x;

func deepz_lower(Sym n, Float c) = (c > 0.0) ? c*(0.0 - 1.0) : c*(1.0);
func deepz_upper(Sym n, Float c) = (c > 0.0) ? c*(1.0) : c*(0.0 - 1.0);

func priority(Neuron n) = n[layer];
func stop(Neuron n) = true;

func x(Float l, Float u, SymExp z) = ((u * z) / (u - l)) + (((u * l) * (sym - 1)) / (2 * (u - l)));
func y(Float l, Float u, SymExp z) = (u / 2) * (1 + sym);

func compute_l(Float u, Float l) = (u * l) / (u - (l + 0.0001)) ;

transformer deepz{
    Affine -> ((prev[Z].dot(curr[weight]) + (curr[bias])).map(deepz_lower), (prev[Z].dot(curr[weight]) + (curr[bias])).map(deepz_upper), prev[Z].dot(curr[weight]) + (curr[bias]));
    Relu -> (0, 0, (prev[l] >= 0.0 ? prev[Z] : (prev[u] <= 0.0 ? 0.0 : x(prev[l], prev[u], prev[Z]))));
}

flow(forward, priority, stop, deepz);
\end{lstlisting}

\textbf{DeepPoly}

\begin{lstlisting}[escapeinside={(*@}{@*)}]
def Shape as (Float l, Float u, PolyExp L, PolyExp U){[(curr[l]<=curr),(curr[u]>=curr),(curr[L]<=curr),(curr[U]>=curr)]};

func simplify_lower(Neuron n, Float coeff) = (coeff >= 0.0) ? (n[l] * coeff) : (coeff * n[u]);
func simplify_upper(Neuron n, Float coeff) = (coeff >= 0.0) ? (coeff * n[u]) : (coeff * n[l]);

func replace_lower(Neuron n, Float coeff) = (coeff >= 0.0) ? (coeff * n[L]) : (coeff * n[U]);
func replace_upper(Neuron n, Float coeff) = (coeff >= 0.0) ? (coeff * n[U]) : (coeff * n[L]);

func priority(Neuron n) = n[layer];
func priority2(Neuron n, Float c) = -n[layer];
func stop(Neuron n) = false;
func stop_traverse(Neuron n, Float c) = false;

func backsubs_lower(PolyExp e, Neuron n) = (e.traverse(backward, priority2, stop_traverse, replace_lower){e <= n}).map(simplify_lower);
func backsubs_upper(PolyExp e, Neuron n) = (e.traverse(backward, priority2, stop_traverse, replace_upper){e >= n}).map(simplify_upper);

func relu(Float x) = x >= 0.0 ? x : 0.0;
func relu_n(Neuron n) = n[l] + n[u] >= 0.0 ? n : 0.0;
func compute_upper(Neuron n) = n[l] >= 0.0 ? n : (n[u] <= 0.0 ? 0.0 : (((n[u]) / ((n[u]) - (n[l]))) * ((n))) - (((n[u]) * (n[l])) / ((n[u]) - (n[l]))));


transformer deeppoly{
    Affine -> (backsubs_lower(prev.dot(curr[weight]) + curr[bias], curr), backsubs_upper(prev.dot(curr[weight]) + curr[bias], curr), prev.dot(curr[weight]) + curr[bias], prev.dot(curr[weight]) + curr[bias]);
    
    Relu -> (relu(prev[l]), relu(prev[u]), relu_n(prev), compute_upper(prev));
}

flow(forward, priority, stop, deeppoly);
\end{lstlisting}

\textbf{ReuseCert}

\begin{lstlisting}[escapeinside={(*@}{@*)}]

func simplify_lower(Neuron n, Float coeff) = (coeff >= 0.0) ? (n[l] * coeff) : (coeff * n[u]);
func simplify_upper(Neuron n, Float coeff) = (coeff >= 0.0) ? (coeff * n[u]) : (coeff * n[l]);

func replace_lower(Neuron n, Float coeff) = (coeff >= 0.0) ? (coeff * n[L]) : (coeff * n[U]);
func replace_upper(Neuron n, Float coeff) = (coeff >= 0.0) ? (coeff * n[U]) : (coeff * n[L]);

func priority(Neuron n) = n[layer];
func priority2(Neuron n, Float c) = -n[layer];
func stop(Neuron n) = false;
func stop_traverse(Neuron n, Float c) = n[layer] <= 1;

func backsubs_lower(PolyExp e, Neuron n) = (e.traverse(backward, priority2, stop_traverse, replace_lower){e <= n}).map(simplify_lower);
func backsubs_upper(PolyExp e, Neuron n) = (e.traverse(backward, priority2, stop_traverse, replace_upper){e >= n}).map(simplify_upper);

func backsubs_lower_L(PolyExp e, Neuron n) = (e.traverse(backward, priority2, stop_traverse, replace_lower){e <= n});
func backsubs_upper_U(PolyExp e, Neuron n) = (e.traverse(backward, priority2, stop_traverse, replace_upper){e >= n});

func relu(Float x) = x >= 0.0 ? x : 0.0;
func relu_n(Float l, Float u, Neuron n) = l + u >= 0.0 ? n[L] : 0.0;
func compute_upper(Float l, Float u, Neuron n) = l >= 0.0 ? n[U] : (u <= 0.0 ? 0.0 : (((u) / ((u) - (l))) * ((n[U]))) - (((u) * (l)) / ((u) - (l))));


transformer reusecert{
    Affine -> ((curr[last_layer] == 1) or (curr[layer] == 1)) ? 
                    (backsubs_lower(prev.dot(curr[weight]) + curr[bias], curr), backsubs_upper(prev.dot(curr[weight]) + curr[bias], curr), prev.dot(curr[weight]) + curr[bias], prev.dot(curr[weight]) + curr[bias]) : 
                    (0, 0, backsubs_lower_L(prev.dot(curr[weight]) + curr[bias], curr), backsubs_upper_U(prev.dot(curr[weight]) + curr[bias], curr));
    
    Relu -> (0, 0, relu_n(prev[L].map(simplify_lower), prev[U].map(simplify_upper), prev), compute_upper(prev[L].map(simplify_lower), prev[U].map(simplify_upper), prev));


}

flow(forward, priority, stop, reusecert);
\end{lstlisting}

\textbf{PolyZono}

\begin{lstlisting}[escapeinside={(*@}{@*)}]

def Shape as (Float l, Float u, PolyExp L, PolyExp U, SymExp Z){[curr[l]<=curr,curr[u]>=curr,curr[L]<=curr,curr[U]>=curr,curr In curr[Z]]};

func simplify_lower(Neuron n, Float coeff) = (coeff >= 0.0) ? (coeff * n[l]) : (coeff * n[u]);
func simplify_upper(Neuron n, Float coeff) = (coeff >= 0.0) ? (coeff * n[u]) : (coeff * n[l]);

func replace_lower(Neuron n, Float coeff) = (coeff >= 0.0) ? (coeff * n[L]) : (coeff * n[U]);
func replace_upper(Neuron n, Float coeff) = (coeff >= 0.0) ? (coeff * n[U]) : (coeff * n[L]);

func deepz_lower(Sym n, Float c) = (c > 0.0) ? c*(0.0 - 1.0) : c*(1.0);
func deepz_upper(Sym n, Float c) = (c > 0.0) ? c*(1.0) : c*(0.0 - 1.0);

func priority(Neuron n) = n[layer];
func priority2(Neuron n, Float c) = -n[layer];
func stop(Neuron n) = false;
func stop_traverse(Neuron n, Float c) = false;

func backsubs_lower(PolyExp e, Neuron n) = (e.traverse(backward, priority2, stop_traverse, replace_lower){e <= n}).map(simplify_lower);
func backsubs_upper(PolyExp e, Neuron n) = (e.traverse(backward, priority2, stop_traverse, replace_upper){e >= n}).map(simplify_upper);


func relu(Float x) = x > 0.0 ? x : 0.0;
func relu_n(Neuron n) = n[l] + n[u] >= 0.0 ? n : 0.0;
func compute_upper(Neuron n) = n[l] > 0.0 ? n : (n[u] < 0.0 ? 0.0 : (((n[u]) / ((n[u]) - (n[l]))) * ((n))) - (((n[u]) * (n[l])) / ((n[u]) - (n[l]))));

transformer polyzono{
    Affine -> (
        max((prev[Z].dot(curr[weight]) + (curr[bias])).map(deepz_lower),backsubs_lower(prev.dot(curr[weight]) + curr[bias], curr)), 
        min((prev[Z].dot(curr[weight]) + (curr[bias])).map(deepz_upper),backsubs_upper(prev.dot(curr[weight]) + curr[bias], curr)), 
        prev.dot(curr[weight]) + curr[bias], 
        prev.dot(curr[weight]) + curr[bias], 
        prev[Z].dot(curr[weight]) + curr[bias]
    );

    Relu -> (
        relu(prev[l]), 
        relu(prev[u]), 
        relu_n(prev), 
        compute_upper(prev), 
        (prev[l] >= 0.0 ? prev[Z] : (prev[u] <= 0.0 ? 0.0 : x(prev[l], prev[u], prev[Z])))
    );
}

flow(forward, priority, stop, polyzono);
\end{lstlisting}

\textbf{ZID}

\begin{lstlisting}[escapeinside={(*@}{@*)}]
def Shape as (Float l, Float u, PolyExp L, PolyExp U, SymExp Z){[curr[l]<=curr,curr[u]>=curr,curr[L]<=curr,curr[U]>=curr,curr In curr[Z]]};

func simplify_lower(Neuron n, Float coeff) = (coeff >= 0.0) ? (coeff * n[l]) : (coeff * n[u]);
func simplify_upper(Neuron n, Float coeff) = (coeff >= 0.0) ? (coeff * n[u]) : (coeff * n[l]);

func replace_lower(Neuron n, Float coeff) = (coeff >= 0.0) ? (coeff * n[L]) : (coeff * n[U]);
func replace_upper(Neuron n, Float coeff) = (coeff >= 0.0) ? (coeff * n[U]) : (coeff * n[L]);

func deepz_lower(Sym n, Float c) = (c > 0.0) ? c*(0.0 - 1.0) : c*(1.0);
func deepz_upper(Sym n, Float c) = (c > 0.0) ? c*(1.0) : c*(0.0 - 1.0);

func priority(Neuron n) = n[layer];
func priority2(Neuron n, Float c) = -n[layer];
func stop(Neuron n) = false;
func stop_traverse(Neuron n, Float c) = false;

func backsubs_lower(PolyExp e, Neuron n) = (e.traverse(backward, priority2, stop_traverse, replace_lower){e <= n}).map(simplify_lower);
func backsubs_upper(PolyExp e, Neuron n) = (e.traverse(backward, priority2, stop_traverse, replace_upper){e >= n}).map(simplify_upper);


func relu(Float x) = x > 0.0 ? x : 0.0;
func relu_n(Neuron n) = n[l] + n[u] > 0.0 ? n : 0.0;
func compute_upper(Neuron n) = n[l] > 0.0 ? n : (n[u] < 0.0 ? 0.0 : (((n[u]) / ((n[u]) - (n[l]))) * ((n))) - (((n[u]) * (n[l])) / ((n[u]) - (n[l]))));

transformer zid{
    Affine -> curr[last_layer] == 1 ? (
        backsubs_lower(prev.dot(curr[weight]) + curr[bias], curr), 
        backsubs_upper(prev.dot(curr[weight]) + curr[bias], curr), 
        prev.dot(curr[weight]) + curr[bias], 
        prev.dot(curr[weight]) + curr[bias], 
        0
    ) : 
    (curr[layer] <= 4 ? (
        (prev[Z].dot(curr[weight]) + (curr[bias])).map(deepz_lower), 
        (prev[Z].dot(curr[weight]) + (curr[bias])).map(deepz_upper), 
        prev.dot(curr[weight]) + curr[bias], 
        prev.dot(curr[weight]) + curr[bias], 
        prev[Z].dot(curr[weight]) + curr[bias]
    ) : 
    (
        (prev.dot(curr[weight]) + curr[bias]).map(simplify_lower), 
        (prev.dot(curr[weight]) + curr[bias]).map(simplify_upper), 
        prev.dot(curr[weight]) + curr[bias], 
        prev.dot(curr[weight]) + curr[bias], 
        0
    )
    )
    ;

    Relu -> curr[last_layer] == 1 ? (
        relu(prev[l]), 
        relu(prev[u]), 
        relu_n(prev), 
        compute_upper(prev),
        0
    ) : 
    (curr[layer] <= 4 ? (
        relu(prev[l]), 
        relu(prev[u]), 
        relu_n(prev), 
        compute_upper(prev),
        ((prev[l]) >= 0.0) ? (prev[Z]) : (((prev[u]) <= 0.0) ? 0.0 : x(prev[l], prev[u], prev[Z]))
    ) : 
    (
        relu(prev[l]), 
        relu(prev[u]), 
        relu_n(prev), 
        compute_upper(prev), 
        0
    )
    )
    ;
}


flow(forward, priority, stop, zid);
\end{lstlisting}

\textbf{SkipPoly}

\begin{lstlisting}[escapeinside={(*@}{@*)}]
def Shape as (Float l, Float u, PolyExp L, PolyExp U){[(curr[l]<=curr),(curr[u]>=curr),(curr[L]<=curr),(curr[U]>=curr)]};

func simplify_lower(Neuron n, Float coeff) = (coeff >= 0.0) ? (n[l] * coeff) : (coeff * n[u]);
func simplify_upper(Neuron n, Float coeff) = (coeff >= 0.0) ? (coeff * n[u]) : (coeff * n[l]);

func replace_lower(Neuron n, Float coeff) = (coeff >= 0.0) ? (coeff * n[L]) : (coeff * n[U]);
func replace_upper(Neuron n, Float coeff) = (coeff >= 0.0) ? (coeff * n[U]) : (coeff * n[L]);

func priority(Neuron n) = n[layer];
func priority2(Neuron n, Float c) = -n[layer];
func stop(Neuron n) = false;
func stop_traverse(Neuron n, Float c) = false;

func backsubs_lower(PolyExp e, Neuron n) = (e.traverse(backward, priority2, stop_traverse, replace_lower){e <= n}).map(simplify_lower);
func backsubs_upper(PolyExp e, Neuron n) = (e.traverse(backward, priority2, stop_traverse, replace_upper){e >= n}).map(simplify_upper);

func relu(Float x) = x >= 0.0 ? x : 0.0;
func relu_n(Neuron n) = n[l] + n[u] >= 0.0 ? n[L] : 0.0;
func compute_upper(Neuron n) = n[l] >= 0.0 ? n[U] : (n[u] <= 0.0 ? 0.0 : (((n[u]) / ((n[u]) - (n[l]))) * ((n[U]))) - (((n[u]) * (n[l])) / ((n[u]) - (n[l]))));

transformer skippoly{
    Affine -> (backsubs_lower(prev.dot(curr[weight]) + curr[bias], curr), backsubs_upper(prev.dot(curr[weight]) + curr[bias], curr), prev.dot(curr[weight]) + curr[bias], prev.dot(curr[weight]) + curr[bias]);
    
    Relu -> (relu(prev[l]), relu(prev[u]), relu_n(prev), compute_upper(prev));

}

flow(forward, priority, stop, skippoly);
\end{lstlisting}

\end{document}